\documentclass[lettersize,journal]{IEEEtran}
\usepackage{amsmath,amsfonts}
\usepackage{algorithmic}
\usepackage{array}
\usepackage[font=normalsize, labelfont=sf, textfont=sf]{subcaption}
\usepackage{textcomp}
\usepackage{stfloats}
\usepackage{url}
\usepackage{verbatim}
\usepackage[colorlinks,linkcolor=red,citecolor=blue]{hyperref}
\usepackage{graphicx}
\usepackage{makecell}
\usepackage{multicol}
\usepackage{multirow}
\usepackage{booktabs}
\usepackage{xcolor}
\def\BibTeX{{\rm B\kern-.05em{\sc i\kern-.025em b}\kern-.08em
    T\kern-.1667em\lower.7ex\hbox{E}\kern-.125emX}}
\usepackage{balance}

\newcommand{\method}{PhaseSurf}

\begin{document}
\title{Volume Rendering with Metric–Phase Fields for Thin Structures and Sign-Ambiguous Geometry}
\author{%
Junkai Deng, Jiayi Kong, Xuhui Chen, Guanli Hou, Fei Hou, Junhui Hou, Wenping Wang, Ying He%
\thanks{Junkai Deng, Jiayi Kong, Guanli Hou and Ying He are with College of Computing and Data Science, Nanyang Technological University, Singapore.}
\thanks{Xuhui Chen and Fei Hou are with the Key Laboratory of System Software (Chinese Academy of Sciences), Institute of Software, Chinese Academy of Sciences, and the University of Chinese Academy of Sciences. }
\thanks{Junhui Hou is with the Department of Computer Science, City University of Hong Kong, Hong Kong SAR, China.}%
\thanks{Wenping Wang is with Department of Computer Science \& Engineering, Texas A\&M University, Texas, USA.}
}

\markboth{Journal of \LaTeX\ Class Files,~Vol.~18, No.~9, September~2020}%
{How to Use the IEEEtran \LaTeX \ Templates}

\maketitle

\begin{abstract}
Neural implicit surface methods that couple volume rendering with signed distance functions (SDFs), such as VolSDF and NeuS, achieve high-quality multi-view reconstruction for watertight objects.
However, these approaches rely on a globally consistent inside--outside partition of space.
This assumption becomes fragile for thin structures (e.g., sheets, tubes, and thin walls) and can break down for sign-ambiguous geometry (e.g., surfaces with boundary, non-manifold junctions, and closely spaced surface components), often resulting in spurious closures or loss of fine structures.
Unsigned distance field (UDF) formulations relax the global sign requirement and better support non-watertight geometry, but are substantially harder to optimize: near the surface, unsigned distances exhibit cusp-like non-differentiability, which complicates stable, occlusion-aware rendering and surface extraction.
We propose \method{}, a neural volume rendering framework built on Metric--Phase Fields (MPFs).
MPFs decompose geometry into a nonnegative metric field capturing unsigned proximity and a continuous phase field providing a soft sign cue. By coupling the two through residual phase injection, MPFs yield a signed composite implicit function with stable near-surface gradients while retaining UDF-like robustness in sign-ambiguous regions. 
This design enables direct iso-surface extraction using standard Marching Cubes and avoids the failure modes of both pure SDF- and UDF-based formulations.
We further introduce MPF-specific training losses that enforce metric consistency, stabilize phase transitions near the interface, and suppress spurious geometry in free space. Experiments on multi-view datasets with thin structures and sign-ambiguous geometry show that \method{} produces more stable reconstructions and higher-quality geometry than UDF-based methods and recent 3DGS pipelines, while avoiding artifacts typical of SDF-based approaches.
\end{abstract}


\section{Introduction}
\label{sec:intro}

Recovering high-quality 3D geometry from posed multi-view images is a central problem in vision and graphics. Neural volume rendering has made this task practical by optimizing 3D representations directly from photometric supervision. However, standard volume rendering formulations that represent geometry implicitly through volumetric density fields often produce blurred geometry, and the extracted surfaces can be noisy or poorly defined.
To obtain sharper and more faithful reconstructions, a line of work bridges volume rendering with implicit surface representations by modeling scene geometry as the zero level set of a learned signed distance function (SDF).
Representative methods such as VolSDF~\cite{Yariv2021Volume} and NeuS~\cite{Wang2021NeuS}, along with many follow-up works, convert signed distances into rendering weights and achieve strong reconstruction and view synthesis performance.
\begin{figure}[!t]
\centering
\begin{minipage}{\linewidth}
\begin{minipage}[b][0.36\linewidth][t]{0.293\linewidth}
\begin{subcaptionblock}{\linewidth}
    \centering
    \includegraphics[width=\linewidth, page=1]{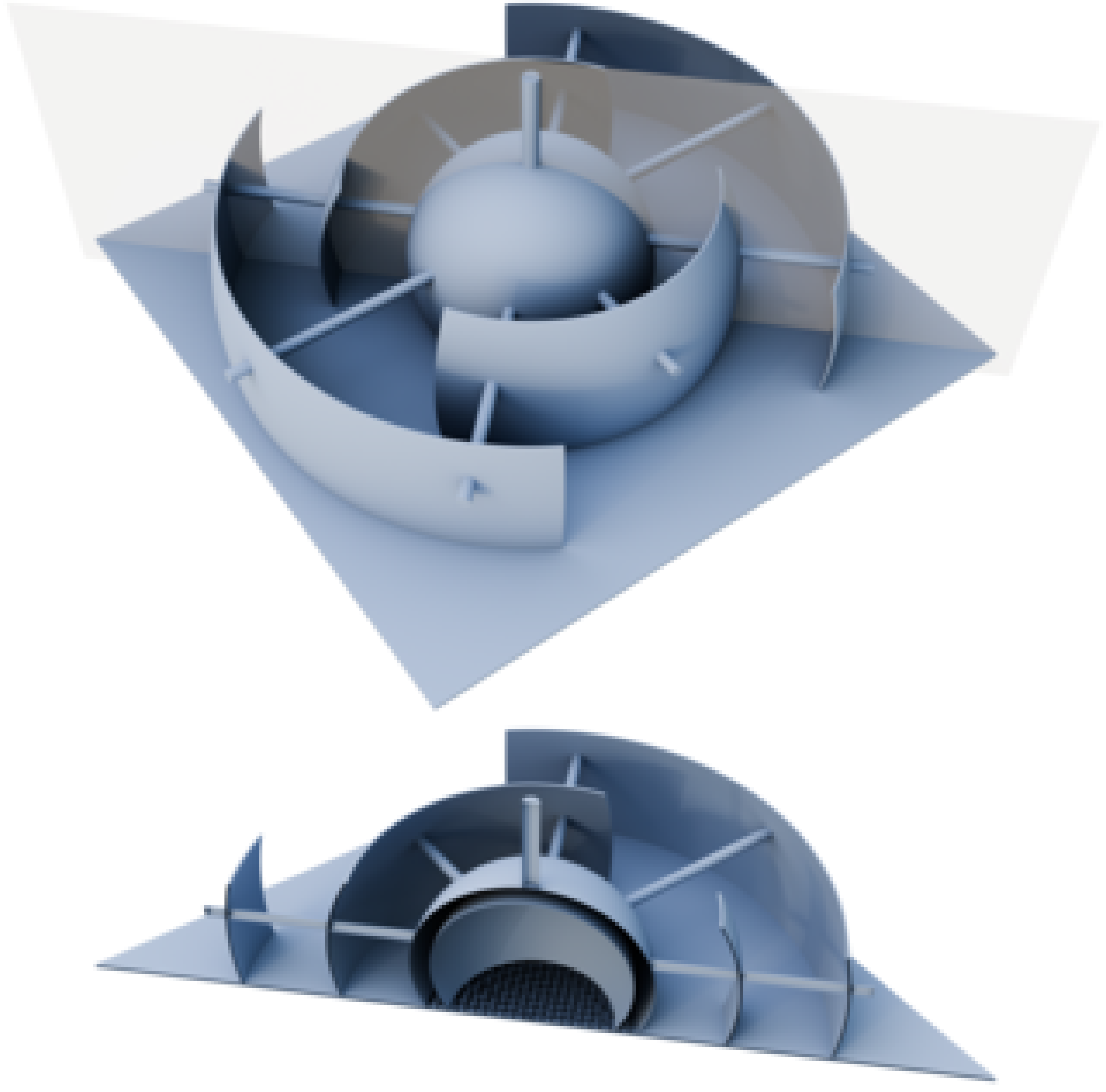}
    \caption{Reference}
    \label{fig:teaser model}
\end{subcaptionblock}
\end{minipage}%
\hfill%
\begin{minipage}[b][0.36\linewidth][s]{0.26\linewidth}
\begin{subcaptionblock}{\linewidth}
    \centering
    \includegraphics[width=\linewidth, page=2]{figs/teaser_toy.pdf}
    \caption{Metric}
    \label{fig:teaser r}
\end{subcaptionblock}\vspace{\fill}
\begin{subcaptionblock}{\linewidth}
    \centering
    \includegraphics[width=\linewidth, page=3]{figs/teaser_toy.pdf}
    \caption{Phase}
    \label{fig:teaser theta}
\end{subcaptionblock}
\end{minipage}%
\hfill%
\begin{minipage}[b][0.36\linewidth][s]{0.26\linewidth}
\begin{subcaptionblock}{\linewidth}
    \centering
    \includegraphics[width=\linewidth, page=4]{figs/teaser_toy.pdf}
    \caption{Indicator}
    \label{fig:teaser P}
\end{subcaptionblock}\vspace{\fill}
\begin{subcaptionblock}{\linewidth}
    \centering
    \includegraphics[width=\linewidth, page=5]{figs/teaser_toy.pdf}
    \caption{MPF}
    \label{fig:teaser phi}
\end{subcaptionblock}
\end{minipage}%
\hfill%
\begin{minipage}[b][0.36\linewidth][t]{0.0812\linewidth}
\begin{minipage}{\linewidth}
    \centering
    \includegraphics[width=\linewidth, page=6]{figs/teaser_toy.pdf}
\end{minipage}
\end{minipage}
\end{minipage}\vspace{5pt}
\rule{\linewidth}{1pt}\vspace{5pt}

\begin{subcaptionblock}{.32\linewidth}
    \centering
    \includegraphics[width=\linewidth]{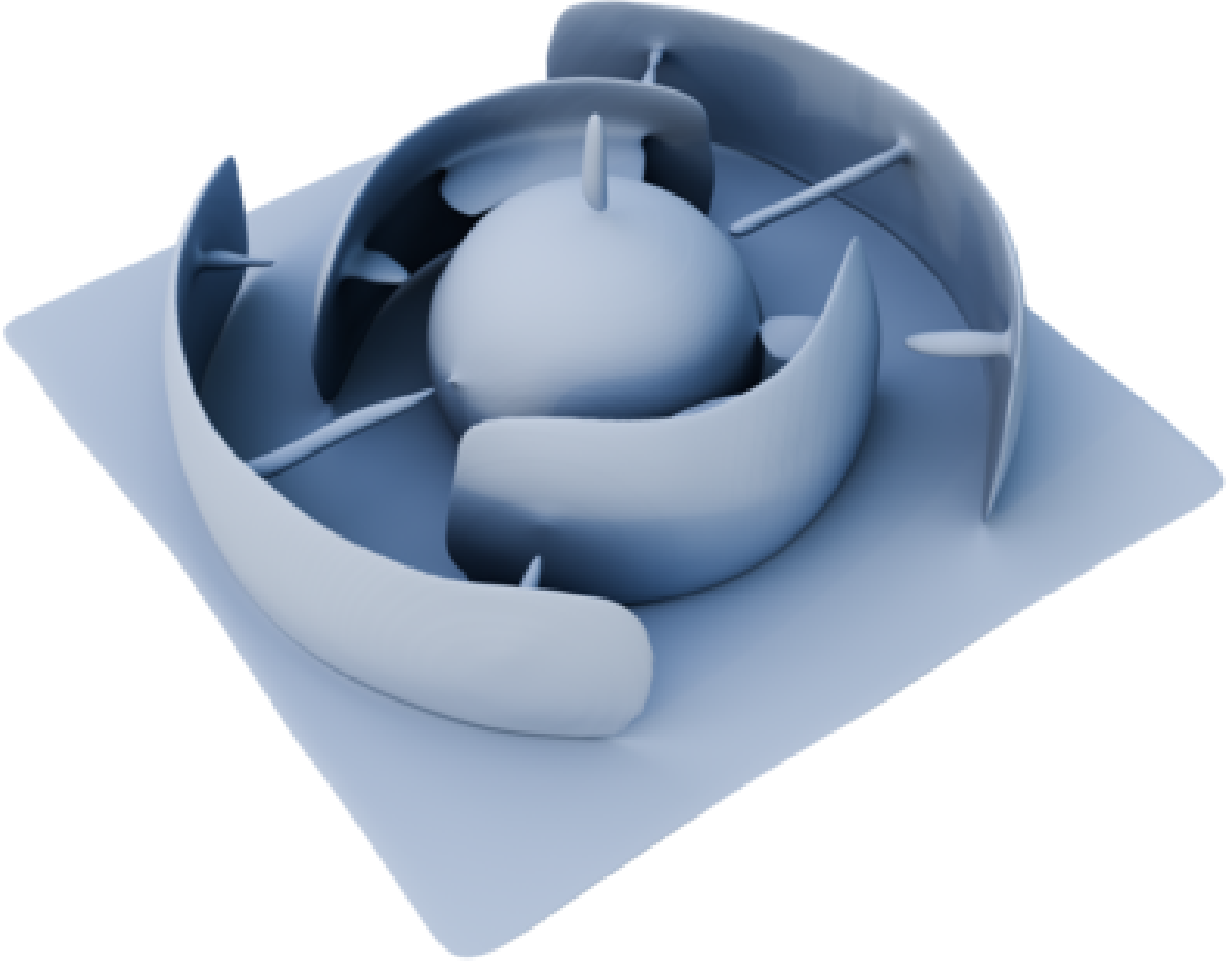}
    \caption{NSH}
    \label{fig:teaser nsh}
\end{subcaptionblock}\hfill%
\begin{subcaptionblock}{.32\linewidth}
    \centering
    \includegraphics[width=\linewidth]{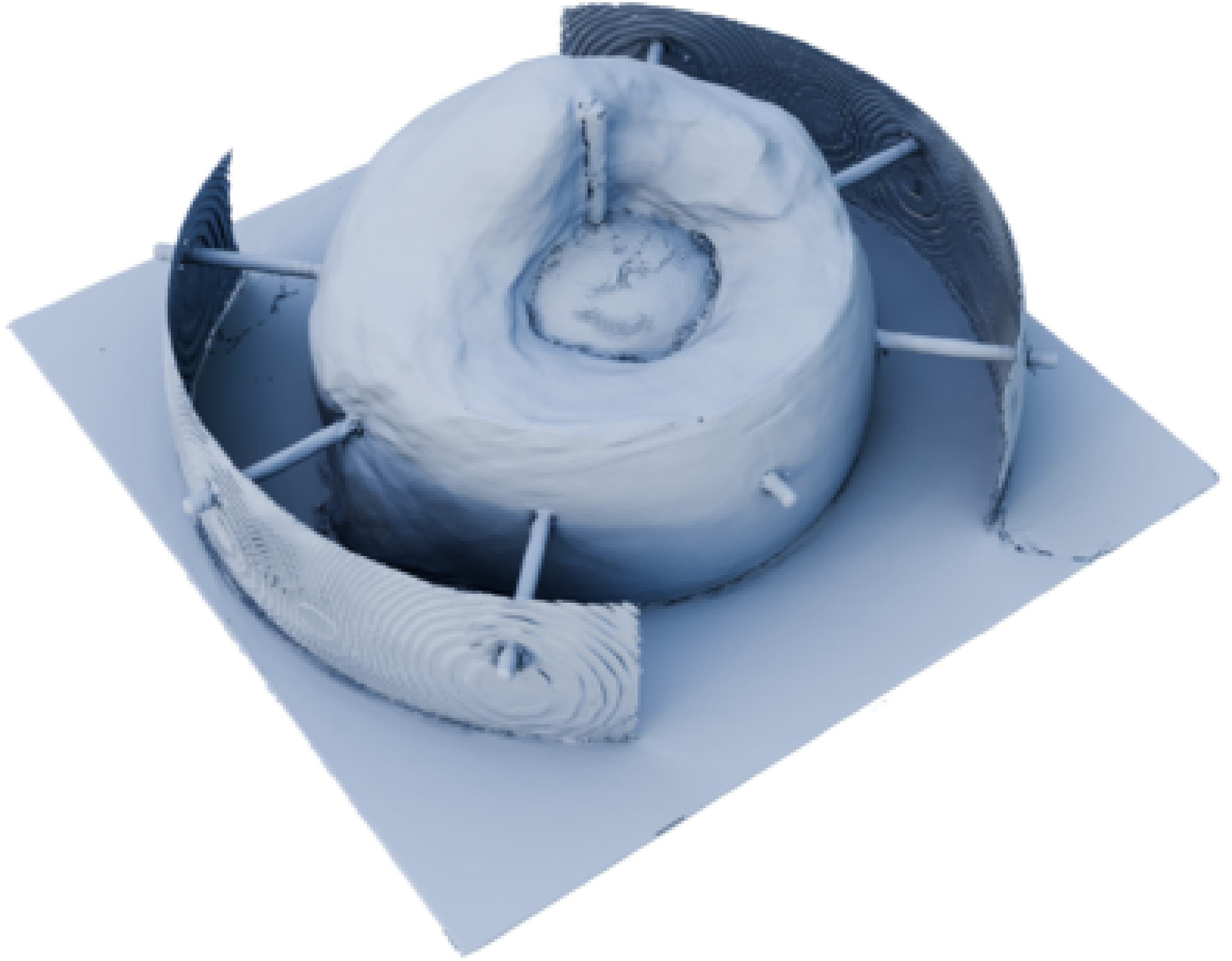}
    \caption{CAP-UDF}
    \label{fig:teaser capudf}
\end{subcaptionblock}\hfill
\begin{subcaptionblock}{.32\linewidth}
    \centering
    \includegraphics[width=\linewidth]{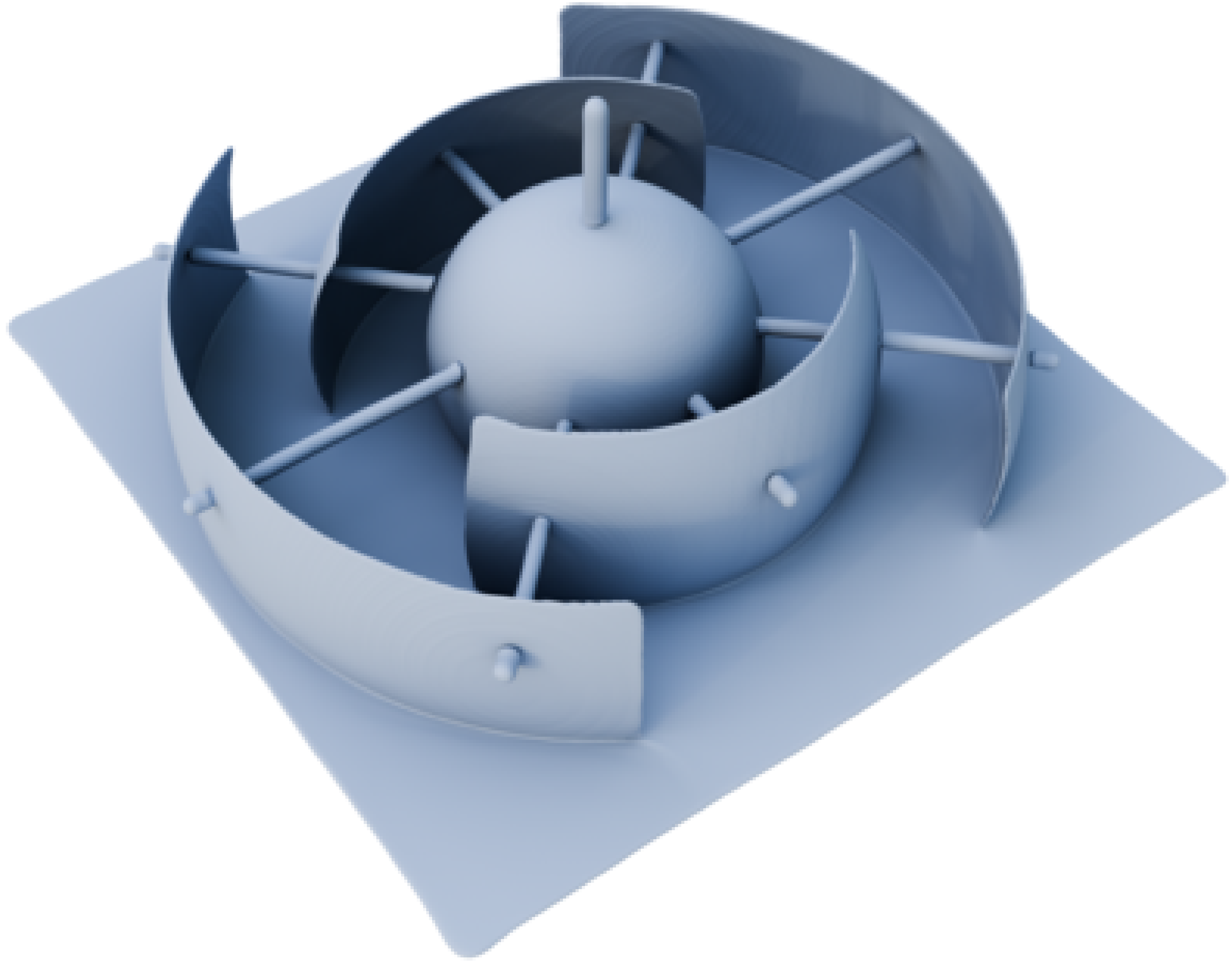}
    \caption{MPF}
    \label{fig:teaser mpf}
\end{subcaptionblock}
\caption{MPF on a challenging toy model. The model consists of two closely spaced open sheets, four slender tubular connections, and a closed spherical component. The sheets are single-layer surfaces with boundaries, whereas the sphere admits a well-defined inside-outside partition. The attachments form non-manifold junctions, and the tubes introduce narrow passages and nontrivial topology. Together, these features pose challenges to both SDF- and UDF-based reconstruction. 
(\subref{fig:teaser model})~Two views of the reference model. (\subref{fig:teaser r})--(\subref{fig:teaser phi})~Cross-sectional visualizations of the metric field $r$, phase field $\theta$, phase indicator $P$, and composite field $\phi$, respectively. (\subref{fig:teaser nsh})--(\subref{fig:teaser mpf})~Reconstructions produced by NSH~\cite{Wang2023NSH}, CAP-UDF~\cite{Zhou2024CAPUDF}, and MPF. Both baselines struggle to recover the thin structures, resulting in incomplete or distorted geometry. In contrast, MPF more faithfully preserves the thin sheets, tubular connections, and overall structure.} 
\label{fig:toymodel}
\end{figure}
Despite their success, SDF-based formulations implicitly assume a globally consistent inside--outside partition.
This assumption is natural for watertight solids, but becomes ambiguous for \emph{open} surfaces and sign-ill-posed configurations frequently encountered in practice, including clothing and cloth folds, thin sheets, layered structures, and non-manifold attachments.
On such shapes, enforcing a single globally signed field biases optimization toward watertight explanations, often producing spurious closures, thickness inflation, or missing sheets.

Unsigned distance fields (UDFs) offer an appealing alternative by removing the need for a global sign and naturally supporting open surfaces.
However, learning UDFs within a volume rendering framework is substantially more challenging.
Near the interface, a UDF behaves like an absolute value function and exhibits a cusp, complicating gradient-based optimization.
More importantly, volume rendering requires converting distances into unbiased and occlusion-aware rendering weights; without sign information, designing stable weight functions becomes delicate.
As a result, recent methods often rely on two-stage optimization or specialized training strategies to achieve robust convergence~\cite{Deng20242SUDF}.

In point-cloud reconstruction, S$^2$DF~\cite{Yang2025S2DF} addresses this non-differentiability by learning a scaled squared distance field that remains differentiable at the zero level set.
However, this idea does not directly translate to posed multi-view reconstruction:
S$^2$DF remains an unsigned, nonnegative field and therefore does not provide an inherent sign change for occlusion reasoning, while the squared-distance formulation yields vanishing gradients on the surface, weakening normal cues and near-surface regularization under purely photometric supervision.
Moreover, S$^2$DF relies on explicit surface anchoring and second-order (Hessian-based) regularization, which are expensive to evaluate densely along camera rays and difficult to constrain from images alone.

Extracting the zero level set from a learned UDF or S$^2$DF is considerably more difficult than from sign-based representations, where standard Marching Cubes can be applied directly. In the absence of sign information, surface extraction typically requires additional cues, such as using gradient directions to detect zero crossings~\cite{ren2023geoudf}, which can be unreliable due to unstable UDF gradients, or resorting to optimization-based extraction methods~\cite{Hou2023Robust,Chen2025DCUDF2}, which are computationally expensive and significantly slower.

In this work, we adopt Metric--Phase Fields (MPFs)~\cite{kong2026mpf} as a geometric backbone for neural volume rendering. MPFs decouple geometry into (i) a metric field $r(\mathbf{x}) \ge 0$ that captures unsigned proximity in a UDF-like manner, and (ii) a continuous phase field $\theta(\mathbf{x})$ that provides a soft sign/side cue via $P(\mathbf{x}) = \tanh(\beta\,\theta(\mathbf{x}))$. The metric and phase components are coupled through residual phase injection to form a signed composite implicit function
$\phi(\mathbf{x}) = r(\mathbf{x}) P(\mathbf{x}) + \theta(\mathbf{x})$ with stable near-surface gradients. The metric branch retains the robustness of unsigned representations for open and thin structures, while the phase branch supplies signed behavior where meaningful, enabling standard iso-surface extraction and SDF-style regularization near the interface.
In sign-ambiguous regions such as thin sheets or tubes, boundaries, and non-manifold junctions, MPFs tend to form a \emph{double cover}—a narrow two-sided shell—that supports a local sign change even when the enclosed volume is arbitrarily thin (Fig.~\ref{fig:toymodel}).

MPFs have demonstrated advantages in reconstructing thin and open geometry from unoriented point clouds~\cite{kong2026mpf}.
Our goal is to bring these advantages to posed multi-view reconstruction by integrating MPFs into a differentiable volume rendering pipeline.
Since the composite field $\varphi$ is signed, we can directly adopt the NeuS~\cite{Wang2021NeuS} signed-field-to-opacity mapping to obtain occlusion-aware rendering weights concentrated near $\{\varphi=0\}$.
Nevertheless, integrating MPFs into NeuS/VolSDF-style frameworks is not a simple drop-in replacement.
Unlike an SDF, $\varphi$ is not inherently distance-calibrated; under purely photometric supervision, its scale and near-surface gradients can become poorly conditioned.
Moreover, the MPF decomposition introduces additional degrees of freedom: many different pairs $(r,\theta)$ can induce similar rendered colors, so photometric supervision alone does not uniquely determine how the two branches should share responsibility.
Finally, sign-ambiguous regimes such as open boundaries and thin structures are especially challenging under multi-view supervision, where only 2D photometric cues mediated by visibility are available.
Preserving thin and open geometry without spurious closures or floaters therefore requires MPF-tailored regularization.

Building on these insights, we introduce \method{}, a neural volume rendering framework that addresses the above challenges by rendering with the signed composite field $\varphi$ while enforcing MPF-specific geometric constraints.
Residual phase injection ensures non-vanishing gradients near the interface, and additional coupling and regularization terms stabilize optimization in sign-ambiguous regions.
This design preserves the advantages of SDF-based surface rendering, including standard Marching Cubes extraction, while substantially improving modeling capacity for open surfaces and thin structures.
Evaluation on datasets with sign-ambiguous geometry and thin structures shows consistent improvements in reconstruction fidelity and training stability over strong SDF- and UDF-based baselines, as well as recent 3D Gaussian Splatting (3DGS) pipelines.

\section{Related Work}
\label{sec:relatedwork}

We briefly review the most relevant work on SDF-, UDF- and 3DGS-based 3D reconstruction from posed multi-view images. 

\paragraph{SDF Methods}
A main line of work couples differentiable rendering with signed distance functions to obtain sharp geometry from multi-view supervision.
Early approaches such as DVR~\cite{Niemeyer2020DVR} and IDR~\cite{yariv2020idr} learn implicit surfaces through differentiable surface rendering.
Building on NeRF-style volume rendering, later methods integrate SDFs into volumetric rendering to jointly optimize geometry and appearance, notably UNISURF~\cite{Oechsle2021unisurf}, VolSDF~\cite{Yariv2021Volume}, and NeuS~\cite{Wang2021NeuS}, which achieve high-quality reconstruction and view synthesis.
Follow-up work improves fidelity and/or efficiency, including HF-NeuS~\cite{Wang2022HFNeuS}, Neuralangelo~\cite{Li2023Neuralangelo}, and NeuS2~\cite{Wang2023NeuS2}.

\paragraph{UDF Methods}
To relax the watertight assumption implicit in SDF-based formulations, several methods learn unsigned distance functions from multi-view photometric supervision.
Representative approaches include NeUDF~\cite{Liu2023NeUDF} and NeuralUDF~\cite{Long2023NeuralUDF}, which introduce UDF-specific mappings from distances to rendering weights or densities, as well as 2S-UDF~\cite{Deng20242SUDF}, which improves robustness via a two-stage optimization strategy.
NeAT~\cite{Meng2023NeAT} also targets arbitrary-topology reconstruction by augmenting SDF modeling with a validity field.
Recent work explores learned or data-driven rendering priors to stabilize UDF learning and improve surface extraction.

\paragraph{3DGS Methods}
3D Gaussian Splatting represents scenes using explicit Gaussian primitives and enables efficient optimization and real-time rendering~\cite{Kerbl20233D}.
Because Gaussians do not directly define an explicit surface, extracting high-quality geometry typically requires additional constraints or dedicated meshing pipelines, such as SuGaR~\cite{guedon2024sugar}, 2DGS~\cite{Deng20242SUDF}, Gaussian surfels~\cite{Dai2024GaussianSurfels}, GOF~\cite{Yu2024GOF}, PGSR~\cite{chen2024pgsr}, and GS2Mesh~\cite{wolf2024gsmesh}.
Hybrid approaches combine Gaussians with implicit surfaces to improve geometric quality, including 3DGSR~\cite{Lyu20243DGSR}, NeuSG~\cite{chen2025neusgneuralimplicitsurface}, and GSDF~\cite{yu2024gsdf}, while additional regularization schemes aim to improve multi-view geometric consistency.

\paragraph{Challenges}
Despite significant progress, reconstructing high-quality surfaces for sign-ambiguous geometries (e.g., open boundaries, thin sheets, and non-manifold junctions) from posed images remains challenging. SDF-based renderers couple geometry and sign in a single field and assume a globally consistent inside--outside partition; when this assumption is violated, they often favor watertight explanations, leading to spurious closures, thickness inflation, or missing thin parts. UDF-based formulations remove the global sign requirement but introduce their own difficulties under image-only supervision: cusp-like behavior near the interface makes optimization fragile, and converting unsigned distances into stable, occlusion-aware rendering weights is non-trivial; reliable surface extraction often requires additional cues or costly post-processing. 3DGS methods excel at efficient view synthesis, but they represent geometry indirectly through opacity distributions over Gaussians, making meshing sensitive to heuristics and regularization; thin or open structures may therefore be over-smoothed, fragmented, or affected by floaters. These challenges are amplified in sign-ambiguous regions, where purely 2D supervision provides weak geometric constraints. Metric--Phase Fields have recently shown promising results for reconstructing thin sheets and other sign-ambiguous geometries from unoriented point clouds. However, extending MPFs from point-cloud reconstruction to posed multi-view images introduces additional challenges. In point-cloud settings, surface samples are directly available and provide strong geometric supervision for learning the metric and phase fields. In contrast, multi-view reconstruction relies only on indirect photometric cues mediated through visibility and rendering. As a result, the additional degrees of freedom introduced by the metric--phase decomposition must be carefully regularized to avoid degenerate solutions. This motivates integrating MPFs into a volume rendering framework with MPF-specific constraints that stabilize optimization while preserving flexibility for open and thin structures.

\section{Preliminaries}
\label{sec:prelim}

\subsection{Implicit Functions for Surface Representations}
\label{subsec:implicit}
Let $S \subset \mathbb{R}^3$ denote the (unknown) target surface and let
$X=\{\mathbf{p}_i\}_{i=1}^N \subset \mathbb{R}^3$ be an \emph{unoriented} point cloud sampled from $S$. For a query point $\mathbf{x}\in\mathbb{R}^3$, we write $\|\cdot\|_2$ for the Euclidean norm and
$\nabla$ for the spatial gradient. We denote the unsigned distance to $S$ by
$\mathrm{dist}(\mathbf{x},S)=\min_{\mathbf{y}\in S}\|\mathbf{x}-\mathbf{y}\|_2$.

A common way to represent $S$ is as the zero level set of a scalar field. A signed distance function $s:\mathbb{R}^3\to\mathbb{R}$ satisfies $
  s(\mathbf{x})=\pm \mathrm{dist}(\mathbf{x},S)$, where the sign is determined by a globally consistent inside--outside partition.
When differentiable, an SDF obeys the Eikonal condition $\|\nabla s(\mathbf{x})\|_2 = 1$
in a tubular neighborhood of $S$, away from singularities such as the medial axis. An unsigned distance function discards the sign, $
  u(\mathbf{x})=\mathrm{dist}(\mathbf{x},S)\ge 0$, and therefore naturally accommodates open surfaces and sign-ambiguous configurations. However,
$u$ is generally \emph{not differentiable} on the interface $\{u=0\}$ (cusp-like behavior), and it does not exhibit a sign change across $S$, making standard iso-surfacing difficult.

Metric--phase fields~\cite{kong2026mpf} decouple \emph{metric proximity} from \emph{phase/sign information}
by learning two continuous fields: an \textbf{unsigned metric field} $r:\mathbb{R}^3\to\mathbb{R}_{\ge 0}$, and a \textbf{phase field} $\theta:\mathbb{R}^3\to\mathbb{R}$. The phase field induces a bounded indicator
\begin{equation}
  P(\mathbf{x}) = \tanh\!\big(\beta\,\theta(\mathbf{x})\big)\in(-1,1),
  \label{eq:phase-indicator}
\end{equation}
where $\beta>0$ is a learnable scaling coefficient controlling the sharpness of the phase
transition (larger $\beta$ yields a steeper transition and faster saturation). MPFs combine metric and phase via a gated-metric formulation with residual phase injection:
\begin{equation}
  \varphi(\mathbf{x}) = r(\mathbf{x})\,P(\mathbf{x}) + \theta(\mathbf{x}).
  \label{eq:mpf}
\end{equation}
The reconstructed surface is extracted as the zero level set $
  \widehat{S}=\{\mathbf{x}\in\mathbb{R}^3:\; \varphi(\mathbf{x})=0\}$. Intuitively, $r(\mathbf{x})$ captures unsigned proximity (UDF-like behavior), while
$\theta(\mathbf{x})$ encodes a soft inside--outside \emph{preference} where such a notion is meaningful. The gated product $rP$ injects signed behavior away from the surface, and the
additive residual $\theta$ stabilizes optimization and gradients near $\{\varphi=0\}$.

MPFs unify desirable properties of SDFs and UDFs, which are particularly useful for
sign-ambiguous geometry such as thin structures, open boundaries, and non-manifold junctions.
\emph{(i) Sign consistency and coincident zero sets.}
Since $\tanh(\cdot)$ is odd and strictly increasing and $\beta>0$, one has
$\mathrm{sign}(P(\mathbf{x}))=\mathrm{sign}(\theta(\mathbf{x}))$ and $P(\mathbf{x})=0 \Leftrightarrow \theta(\mathbf{x})=0$.
Because $r(\mathbf{x})\ge 0$, the product $r(\mathbf{x})P(\mathbf{x})$ has the same sign as $\theta(\mathbf{x})$ (or is zero).
Hence, for any $\mathbf{x}$ with $\theta(\mathbf{x})\neq 0$,
\begin{equation}  \mathrm{sign}\big(\varphi(\mathbf{x})\big)=\mathrm{sign}\big(\theta(\mathbf{x})\big),
  \qquad
  \varphi(\mathbf{x})=0 \;\Leftrightarrow\; \theta(\mathbf{x})=0.
  \label{eq:sign-consistency}
\end{equation} \emph{(ii) Stable near-surface gradients via residual phase injection.}
Differentiating~\eqref{eq:mpf} yields
\begin{equation}
  \nabla\varphi(\mathbf{x})
  = P(\mathbf{x})\,\nabla r(\mathbf{x})
     + \Big(1 + r(\mathbf{x})\,\beta\,\mathrm{sech}^2(\beta\theta(\mathbf{x}))\Big)\nabla\theta(\mathbf{x}).
  \label{eq:mpf-gradient}
\end{equation}
Crucially, near the interface where $r(\mathbf{x})\approx 0$, the factor multiplying $\nabla\theta(\mathbf{x})$
is close to $1$, providing a non-vanishing gradient pathway to the phase branch and stabilizing
optimization around the zero level set. \emph{(iii) First-order agreement at constrained samples.}
In point-based learning, one typically enforces \emph{both} $r(\mathbf{p})=0$ and $\varphi(\mathbf{p})=0$ for
samples $\mathbf{p}\in X$. Since $\varphi(\mathbf{p})=0 \Rightarrow \theta(\mathbf{p})=0$ by~\eqref{eq:sign-consistency},
substituting $r(\mathbf{p})=0$ and $P(\mathbf{p})=0$ into~\eqref{eq:mpf-gradient} gives
\begin{equation}
  \nabla\varphi(\mathbf{p})=\nabla\theta(\mathbf{p}),
  \label{eq:gradient-agreement}
\end{equation}
so interface normals computed from $\nabla\varphi$ are stable and coincide with those of
$\theta$ at constrained samples. By enforcing an on-surface unit-gradient (Eikonal) constraint on $\theta$, \eqref{eq:gradient-agreement} implies that $\phi$ exhibits a sign change
across the interface, unlike a UDF which is nonnegative everywhere.  \emph{(iv) Double covering in sign-ambiguous regions.}
In regions where a globally consistent inside--outside labeling is ill-defined (e.g., thin
sheets, boundaries, and non-manifold junctions), MPFs tend to form a narrow two-sided shell
around the underlying surface. This induced double covering exhibits opposite signs on its
two sides and therefore supports a meaningful sign change, even though the enclosed volume
can be arbitrarily thin. \emph{(v) Standard iso-surface extraction.}
Unlike UDFs, the composite field $\varphi$ is signed and exhibits sign changes across the
surface, so $\widehat{S}=\{\varphi=0\}$ can be extracted directly using standard Marching
Cubes. Figure~\ref{fig:toymodel} illustrates MPFs on a 3D toy model with thin sheets,
boundaries, and non-manifold junctions.

\subsection{Differentiable Volume Rendering with Implicit Functions}
\label{subsec:preliminaries_rendering_with_implicit_functions}

We assume posed multi-view images with known camera intrinsics and extrinsics. For a pixel with ray origin $\mathbf{o}\in\mathbb{R}^3$ and unit direction $\mathbf{d}\in\mathbb{S}^2$, we consider the camera ray $\mathbf{r}(t)=\mathbf{o}+t\mathbf{d}$. Neural volume rendering models appearance using a color field $c(\mathbf{x},\mathbf{d})\in[0,1]^3$ and a nonnegative density (opacity) field $\sigma(\mathbf{x})\ge 0$, which determines the probability of ray termination. The rendered pixel color is given by the standard volume rendering integral
$C(\mathbf{r})=\int T(t)\sigma(\mathbf{r}(t))c(\mathbf{r}(t),\mathbf{d})\mathrm{d}t$, where $T(t)=\exp(-\int_{0}^{t}\sigma(\mathbf{r}(u))\mathrm{d}u)$ is the accumulated transmittance along the ray. In practice, the integral is approximated
by sampling points along the ray and performing alpha compositing.

Surface-based neural rendering methods parameterize geometry by a \emph{signed implicit field}
$f:\mathbb{R}^3\to\mathbb{R}$ whose zero level set $\{f=0\}$ represents the surface. The key idea is to convert
signed values $f(\mathbf{x})$ into densities or opacities so that rendering weights concentrate near the interface
and remain occlusion-aware. A common strategy defines a smooth occupancy via a monotone cumulative density function (CDF),
$\Phi(\mathbf{x})=\mathrm{CDF}\!\big(f(\mathbf{x})/\gamma\big)$, where $\gamma>0$ controls the thickness of the
transition band. Different choices of $\mathrm{CDF}$ lead to different renderers, e.g., the Laplace CDF in
VolSDF~\cite{Yariv2021Volume} and the logistic-CDF-based formulation in NeuS~\cite{Wang2021NeuS}, and a more general setting in UNIS~\cite{Deng2025UNIS}. 

Given posed images, model parameters are optimized using photometric supervision by minimizing the discrepancy
between the rendered color $\widehat{C}(\mathbf{r})$ and the ground-truth pixel color $C^{\star}(\mathbf{r})$,
aggregated over rays from all training views. This photometric loss is typically combined with geometric
regularizers on $f$ (e.g., Eikonal-style constraints) to stabilize optimization.

\section{Proposed Method}
\label{sec:method}

\begin{figure}[t]
    \centering
    \includegraphics[width=\linewidth]{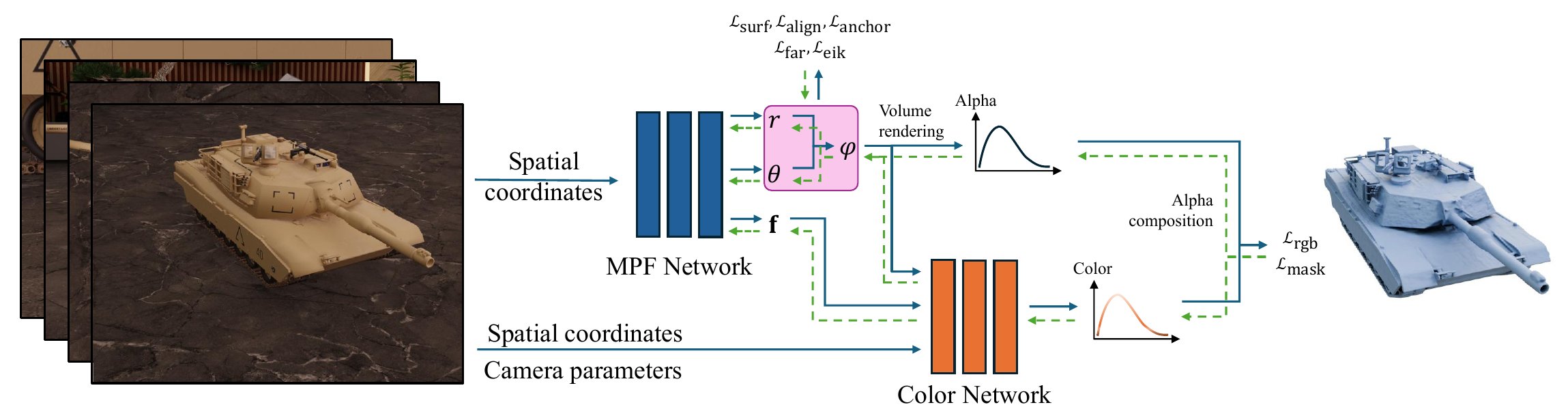}
    \caption{Pipeline. For each 3D sample along camera rays, the MPF network predicts an unsigned metric $r(\mathbf{x})\ge0$, a phase $\theta(\mathbf{x})$, and a feature $\mathbf{f}(\mathbf{x})$. We form the signed composite field $\varphi(\mathbf{x})=r(\mathbf{x})\,\tanh(\beta\theta(\mathbf{x}))+\theta(\mathbf{x})$ and use NeuS-style signed-field volume rendering to compute occlusion-aware weights. The radiance MLP predicts view-dependent color, and the volume rendering network produces RGB (supervised by input images) and accumulated opacity $\alpha$ (optionally supervised by masks). %
    }
    \label{fig:pipeline}
\end{figure}

Given posed multi-view images, our goal is to reconstruct a surface that may contain
sign-ambiguous structures such as thin sheets, open boundaries, and non-manifold junctions.
We build on the NeuS pipeline~\cite{Wang2021NeuS} and replace the SDF geometry backbone by an MPF. Concretely, we represent geometry using two coupled fields:
an unsigned metric field $r(\mathbf{x})\ge 0$ and a phase field $\theta(\mathbf{x})$,
and form the signed composite implicit function $  \varphi(\mathbf{x}) = r(\mathbf{x})\,P(\mathbf{x}) + \theta(\mathbf{x})$,
whose zero level set $\{\varphi=0\}$ defines the reconstructed surface.
Since $\varphi$ is signed, we can directly adopt NeuS-style rendering to obtain
occlusion-aware weights concentrated around the interface. The key difference from NeuS is
our MPF-specific regularization: under purely photometric supervision, the metric and phase
branches can otherwise drift or collapse. We therefore augment the RGB loss with a set of
losses tailored to MPFs to stabilize optimization near the interface and preserve 
UDF-like robustness in sign-ambiguous regions.

\subsection{MPF Geometry and Appearance Networks}
\label{subsec:method_networks}

As illustrated in Fig.~\ref{fig:pipeline}, our model contains two neural components:
(i) an MPF geometry network that predicts the metric--phase decomposition and a
geometry feature, and (ii) a radiance network that predicts view-dependent color.
The composite field $\varphi$ is used for NeuS-style signed surface rendering,
while MPF-specific regularizers are applied to the predicted fields to stabilize
image-only training.

\paragraph{Geometry network (MPF)}
We parameterize MPF geometry with an MLP that maps a 3D query point $\mathbf{x}$ to an unsigned metric field $r(\mathbf{x})\ge 0$, a phase field $\theta(\mathbf{x})\in\mathbb{R}$, and a learned feature vector $\mathbf{f}(\mathbf{x})$ for appearance.
Following NeuS~\cite{Wang2021NeuS}, we adopt a shared trunk with lightweight heads.
Given $\mathbf{x}$, we apply positional encoding $\gamma(\mathbf{x})$ and compute a shared representation $\mathbf{h}(\mathbf{x})=F_{\psi}(\gamma(\mathbf{x}))$.
Two scalar heads predict $\tilde r(\mathbf{x})=g_r(\mathbf{h}(\mathbf{x}))$ and $\theta(\mathbf{x})=g_{\theta}(\mathbf{h}(\mathbf{x}))$, while an additional head (or intermediate activation) outputs the appearance feature $\mathbf{f}(\mathbf{x})$.
Non-negativity of the metric branch is enforced by construction via $r(\mathbf{x})=\mathrm{softplus}(\tilde r(\mathbf{x}))$.
From the phase field we derive a bounded indicator $P(\mathbf{x})=\tanh(\beta\,\theta(\mathbf{x}))$, where $\beta>0$ controls the sharpness of the phase transition and is learned with a positive parameterization (e.g., $\beta=\mathrm{softplus}(\tilde\beta)$).
The signed composite implicit function is then formed as
$\varphi(\mathbf{x}) = r(\mathbf{x})\,P(\mathbf{x}) + \theta(\mathbf{x})$, and geometry is extracted from the zero level set $\{\varphi=0\}$. Surface normals are computed from the composite field via automatic differentiation as
$\mathbf{n}(\mathbf{x})=\nabla_{\mathbf{x}}\varphi(\mathbf{x})/(\|\nabla_{\mathbf{x}}\varphi(\mathbf{x})\|_2+\varepsilon)$.
A key practical benefit of residual phase injection is improved conditioning near the interface: around $\{\varphi=0\}$, the metric term becomes small, while the additive phase term provides a direct, non-vanishing gradient pathway.
This stabilizes normal estimation and supports robust optimization with ray-sampled regularizers, including anchoring, alignment, and Eikonal-style constraints.

\paragraph{Appearance network (radiance)}
We model view-dependent radiance using a rendering MLP $c_{\omega}$, following NeuS.
Given a query point $\mathbf{x}$ and a viewing direction $\mathbf{d}$, the radiance network takes as input the encoded position and direction, the normal from the geometry network, and the learned feature $\mathbf{f}(\mathbf{x})$, and predicts
$
c(\mathbf{x},\mathbf{d})
=
\mathrm{sigmoid}\big(
c_{\omega}\big(
\gamma(\mathbf{x}),\,\gamma(\mathbf{d}),\,\mathbf{n}(\mathbf{x}),\,\mathbf{f}(\mathbf{x})
\big)
\big)\in[0,1]^3. 
$
Importantly, the radiance network predicts color only; opacity is derived from the signed composite field $\varphi$ via the NeuS-style mapping described in Section~\ref{subsec:method_rendering}.
This design mirrors the standard NeuS pipeline and isolates the effect of replacing the SDF geometry backbone with an MPF.

\subsection{Volume Rendering with MPFs}
\label{subsec:method_rendering}

For each pixel ray $\mathbf{r}(t)=\mathbf{o}+t\mathbf{d}$, we sample depths
$t_1<t_2<\cdots<t_n$ and evaluate the composite field $\varphi_i=\varphi(\mathbf{r}(t_i))$ and
color $c_i=c(\mathbf{r}(t_i),\mathbf{d})$.
We adopt the same signed-field-to-opacity mapping as NeuS~\cite{Wang2021NeuS}. Specifically,
we define an occupancy function via a monotone CDF, 
  $\Phi_i = \mathrm{CDF}\left(\varphi_i/s\right)$, where $s>0$ controls the width of the transition band (learned as in NeuS). 
  
The per-interval
opacity is estimated from the change of occupancy along the ray, $\alpha_i
=
\operatorname{clip}_{[0,1]}((\Phi_i - \Phi_{i+1})/(\Phi_i + \varepsilon))$,
with a small $\varepsilon$ for numerical stability. Rendering weights are then computed as
$w_i = \alpha_i \prod_{j<i}(1-\alpha_j)$ and the rendered color is $\widehat{C}(\mathbf{r}) = \sum_{i=1}^{n} w_ic_i$. Since $\varphi$ is signed, this formulation produces
well-conditioned zero crossings and occlusion-aware weights, while MPFs retain flexibility
in sign-ambiguous regions through the metric/phase decomposition.

\subsection{Training Objective}
\label{subsec:method_objective}

We optimize all network parameters by combining standard photometric supervision with MPF-specific
geometric regularization.

\paragraph{Photometric supervision} Let $\mathcal{R}$ denote the set of camera rays corresponding to
pixels sampled from the training images. The RGB loss is
\begin{equation}
\mathcal{L}_{\mathrm{rgb}}
= \frac{1}{|\mathcal{R}|}\sum_{\mathbf{r}\in\mathcal{R}}
\|\widehat{C}(\mathbf{r})-C^{\star}(\mathbf{r})\|_1.
\label{eq:method_rgb}
\end{equation}

\paragraph{Near-surface regularization}
In MPF learning, several constraints are naturally imposed in a narrow band around the surface.
In point-cloud reconstruction, such constraints can be  enforced using explicit surface samples. In multi-view reconstruction, however, we do not have direct 3D supervision; instead, we exploit
the fact that NeuS-style rendering weights concentrate near the zero level set of the signed composite field.
We therefore use the rendering weights $\{w_i\}$ as soft indicators of near-surface locations
along each ray and apply the following regularizers at high-weight samples. First, we enforce an Eikonal-style constraint on both the composite field $\varphi$ and the phase field $\theta$ to encourage well-conditioned
signed behavior near the interface:
\begin{multline}
\mathcal{L}_{\mathrm{surf}}
= \frac{1}{|\mathcal{R}|n}\sum_{\mathbf{r}\in\mathcal{R}}
\sum_{i=1}^{n} w_i \bigg( \left( \|\nabla\varphi(\mathbf{r}(t_i))\|_2 - 1 \right)^2 +\\ \left( \|\nabla\theta(\mathbf{r}(t_i))\|_2 - 1 \right)^2\bigg).
\label{eq:method_surf}
\end{multline}
Although $\nabla\varphi \approx \nabla\theta$ near the interface, we intentionally include both terms to
improve stability early in training when the predicted interface (and thus the near-surface set) can be noisy. Second, to prevent degenerate decompositions under purely photometric supervision (e.g., metric collapse or phase
dominance), we align the gradients of the metric and phase fields near the surface:
\begin{equation}
\mathcal{L}_{\mathrm{align}}
= \frac{1}{|\mathcal{R}|n}\sum_{\mathbf{r}\in\mathcal{R}}
\sum_{i=1}^{n} w_i
\left(
1 - \left\langle \widehat{\nabla r}_i,\ \widehat{\nabla\theta}_i \right\rangle^2
\right),
\label{eq:method_align}
\end{equation}
where $\widehat{\nabla r}_i=\nabla r(\mathbf{r}(t_i))/(\|\nabla r(\mathbf{r}(t_i))\|_2+\varepsilon)$
(and analogously for $\widehat{\nabla\theta}_i$), and the squared inner product makes the constraint
orientation-invariant, which is desirable for sign-ambiguous regions where a consistent global orientation may not exist. Finally, we anchor the MPF interface to the rendered surface by encouraging both the composite field and the metric field
to be small at near-surface samples:
\begin{equation}
\mathcal{L}_{\mathrm{anchor}}
= \frac{1}{|\mathcal{R}|n}\sum_{\mathbf{r}\in\mathcal{R}}
\sum_{i=1}^{n} w_i
\Big( |\varphi(\mathbf{r}(t_i))| + r(\mathbf{r}(t_i)) \Big).
\label{eq:method_anchor}
\end{equation}
Together, these near-surface terms make $\varphi$ behave SDF-like where rendering requires stable signed gradients,
while keeping the metric--phase decomposition well-conditioned.

\paragraph{Free-space regularization}
Away from the surface, MPFs should exhibit stable and unambiguous behavior: the metric branch should behave like an unsigned
distance field, while the phase branch should saturate to a clear sign.
We apply the following regularizers to samples in free space (i.e., low rendering weights). We define $\overline{w}_{i}^{\mathrm{free}}=1-\exp(-100|r|)$ to account for occluded surfaces. First, we encourage the metric field $r$ to approximate an unsigned distance field by enforcing an Eikonal-style constraint:
\begin{equation}
\mathcal{L}_{\mathrm{eik}}
= \frac{1}{|\mathcal{R}|n}\sum_{\mathbf{r}\in\mathcal{R}}
\sum_{i=1}^{n}\overline{w}_{i}^{\mathrm{free}}
\left(\|\nabla r(\mathbf{r}(t_i))\|_2 - 1 \right)^2 .
\label{eq:method_eik}
\end{equation}
Second, we adopt a common regularizer found in UDF reconstruction literatures~\cite{Long2023NeuralUDF,Deng20242SUDF} that prevents large regions with near-zero UDF values by encouraging the UDF to be as large as possible.
In our case, this is achieved by regularizing $r$ which should resemble the UDF of the geometry. We additionally constraint the absolute value of the composite field ($\lvert\varphi\rvert$) as well.
\begin{equation}
\mathcal{L}_{\mathrm{far}}
= \frac{1}{|\mathcal{R}|}\sum_{\mathbf{r}\in\mathcal{R}}
\sum_{i=1}^{n} \overline{w}_{i}^{\mathrm{free}}
\exp\!\left(-\alpha_f\left(r(\mathbf{r}(t_i)) + |\varphi(\mathbf{r}(t_i))|\right)\right).
\label{eq:method_far}
\end{equation}
Here $\alpha_f=50$ controls the sharpness of the barrier. %
This loss is active where the sample point is at least a little distance away from the surface. We don't apply this loss exactly on the surface to prevent this loss from messing up with the actual surface.
Together, these terms encourage $r$ to remain a stable unsigned proximity measure away from the surface, while $\theta$ yields a clear,
well-separated phase, reducing spurious structures in unobserved regions.

\paragraph{Optional silhouette supervision}
When foreground masks are available, we additionally supervise the accumulated opacity
$\widehat{A}(\mathbf{r})=\sum_{i=1}^{n} w_i$ using an $\ell_1$ loss:
\begin{equation}
\mathcal{L}_{\mathrm{mask}}
= \frac{1}{|\mathcal{R}|}\sum_{\mathbf{r}\in\mathcal{R}}
\Big\|\widehat{A}(\mathbf{r}) - A^{\star}(\mathbf{r})\Big\|_1 .
\label{eq:method_mask}
\end{equation}
This provides a strong geometric cue that reduces the ambiguity of purely photometric training, especially for thin/open structures.

The total training loss is a weighted sum of four components:
\begin{equation}
\begin{aligned}
\mathcal{L}
= {} &
\underbrace{\mathcal{L}_{\mathrm{rgb}}}_{\text{photometric supervision}}
 \\[2pt]
& +\; \underbrace{
\lambda_{\mathrm{anchor}}\mathcal{L}_{\mathrm{anchor}}
+\lambda_{\mathrm{surf}}\mathcal{L}_{\mathrm{surf}}
+\lambda_{\mathrm{align}}\mathcal{L}_{\mathrm{align}}
}_{\text{near-surface regularization}}
\\[2pt]
& +\;
\underbrace{
\lambda_{\mathrm{eik}}\mathcal{L}_{\mathrm{eik}}
+\lambda_{\mathrm{far}}\mathcal{L}_{\mathrm{far}}
}_{\text{free-space regularization}}
\quad + \underbrace{
\lambda_{\mathrm{mask}}\mathcal{L}_{\mathrm{mask}}
}_{\text{optional silhouette supervision}}.
\end{aligned}
\label{eq:method_full_loss}
\end{equation}
We follow the standard NeuS training pipeline for optimization and ray sampling, and normalize each scene to a fixed bounding volume.
For initialization, we adopt NeuS-style sphere initialization to obtain a well-conditioned signed field at the start of optimization:
we initialize the phase branch $\theta$ (and thus the composite field $\varphi$) to approximate the signed distance of a coarse sphere,
and initialize the metric branch $r$ as a nonnegative, distance-like field (e.g., roughly proportional to $|\theta|$) so that training begins
with stable zero crossings and meaningful gradients.
When coarse geometric priors are available (e.g., a visual hull, sparse SfM points, or depth-based initialization), we can initialize $\theta$
from the corresponding coarse implicit surface instead of a sphere. During training, we jointly optimize the network parameters together with the NeuS sharpness parameter $s$ (controlling the transition width)
and the phase scale $\beta$.
To balance the loss terms and avoid the early collapse of the metric--phase decomposition, we use a warm-up schedule:
in the first $T_{\mathrm{warm}}=50,000$ iterations, we optimize primarily with $\mathcal{L}_{\mathrm{rgb}}$ and near-surface terms and $\mathcal{L}_{\mathrm{eik}}$ that directly
stabilize the rendered interface (e.g., $\mathcal{L}_{\mathrm{anchor}}$ and $\mathcal{L}_{\mathrm{surf}}$, and $\mathcal{L}_{\mathrm{mask}}$
when masks are available). We then progressively enable the coupling and free-space regularizers by ramping $\lambda_{\mathrm{align}}$ and
$\lambda_{\mathrm{far}}$ from $0$ to their target values. In practice, we choose $\lambda$ such that the four groups in Eq.~\eqref{eq:method_full_loss} contribute
comparable gradient magnitudes. Optionally, we treat rendering weights as soft selectors and stop gradients through $w_i$ when used inside
regularization losses, preventing trivial solutions that reduce the penalty by reshaping the weight distribution.

\paragraph{Remarks}
In our work, the regularization losses look similar to the one used by the original MPF literature~\cite{kong2026mpf}. However, MPF is designed for surface reconstruction from known point clouds, while our work focuses on reconstruction from multi-view images.

In the point cloud reconstruction task, a subset of ground truth location are known as priors in the form of input point clouds. This information not only makes the separation of near surface and far-away points easy, but also makes the definition of various losses ($\mathcal{L}_{\mathrm{anchor}}$, $\mathcal{L}_{\mathrm{surface}}$, $\mathcal{L}_{\mathrm{align}}$, $\mathcal{L}_{\mathrm{far}}$) straightforward. In this case, $\mathcal{L}_{\mathrm{anchor}}$ and $\mathcal{L}_{\mathrm{surface}}$ act as supervision losses.

In the multi-view images reconstruction task however, such ground truth is not available. The location of the surface is dynamic and can change anytime. This poses a fundamental distinction between our work and the original MPF. In our work, we use a simple hypothesis that the surface location in every iteration of training can be roughly determined from the rendering weights of each sample along the ray. In our case, the only supervision losses are $\mathcal{L}_{\mathrm{rgb}}$ and the optional $\mathcal{L}_{\mathrm{mask}}$, while the other loss terms $\mathcal{L}_{\mathrm{anchor}}$, $\mathcal{L}_{\mathrm{surface}}$, $\mathcal{L}_{\mathrm{align}}$, $\mathcal{L}_{\mathrm{far}}$ are regularization terms.

The two tasks -- reconstruction from point clouds and reconstruction from multi-view images -- are related. But in our case, they are fundamentally different and the adaptation is therefore nontrivial.

\section{Experiments}
\label{sec:experiments}

\subsection{Experiment Settings}
\paragraph{Datasets} The common benchmarks, such as DTU~\cite{Anaaes2016Large} and Tanks \& Temples (T\&T)~\cite{Knapitsch2017TnT}, are widely adopted for assessing multi-view reconstruction quality from posed images. However, these datasets primarily contain regular object- or scene-level geometry and are dominated by closed or nearly watertight surfaces, with few thin-sheet structures or non-manifold junctions. To evaluate behavior beyond the watertight assumption, we test on garment models selected from DeepFashion3D~\cite{Zhu2020DF3D}, which contains garment-like surfaces with boundaries and thin structures. We also introduce the \emph{Thin Structure and Sign-Ambiguous (TSSA)} benchmark, a curated collection of challenging geometries exhibiting thin structures and sign ambiguity. It comprises eight real and synthetic scenes, featuring open surfaces, thin tubes and sheets, closely spaced layers, and non-manifold junctions, precisely the regimes where SDF-based methods tend to produce spurious closures and UDF-based methods often struggle with stable rendering and surface extraction. For each scene, we use 120 images for training and evaluating novel view synthesis quality.

\paragraph{Baselines} Since our primary focus is reconstructing sign-ambiguous geometry, where the watertight assumptions underlying SDF-based methods are often violated, we mainly compare against UDF- and 3D Gaussian Splatting approaches. Specifically, we include GaussianUDF~\cite{li2025gaussianudf}, a state-of-the-art UDF-based method, as well as recent 3DGS pipelines including PGSR~\cite{chen2024pgsr},  GeoSVR~\cite{li2025geosvr} and GGGS~\cite{zhang2026geometrygroundedgaussiansplatting}. For completeness, we also report results for NeuS~\cite{Wang2021NeuS}, HF-NeuS~\cite{Wang2022HFNeuS} and Neuralangelo~\cite{Li2023Neuralangelo}, the classical SDF-based neural rendering methods. We assess geometric accuracy using Chamfer Distance (CD) and normal consistency (NC) whenever ground-truth 3D geometry is available. The normal consistency is measured as cosine similarity between the ground-truth normal and the predicted normal.

\subsection{Results and Comparisons}

\paragraph{Results on the DTU dataset} Table~\ref{tab:dtu cd} shows the Chamfer distance metrics among our method and representative implicit-field methods including VolSDF and NeuS. We particularly found that our method can resolve thin structures better than baseline methods, as shown in Figure~\ref{fig:dtu}, which can be attributed to our decomposition greatly increases the express capability of the neural network that encodes the implicit representation of the geometry, which as a result can better represent sign changes and ambiguity in a small region that require significant changes of the implicit representation value.

On the other hand, however, our expressiveness can have adverse effects on regions with high specular lights or shadows that change dramatically across viewing angles and lack enough consistent color or texture cues for the geometry, as shown in Figure~\ref{fig:dtu-dent}. A less expressive neural network usually favors a flat surface in this region, which is usually the desired behavior. But our enhanced expression capability results in a greater tendency to distort the surface to create dents or holes in these regions in order to compensate for the color changes. This results in the slightly inferior Chamfer distance metrics in some test cases in Table~\ref{tab:dtu cd}.

\begin{table*}[!ht]
    \setlength{\tabcolsep}{2pt}
    \centering
    \caption{Chamfer distance comparison on the DTU dataset~\cite{Anaaes2016Large}.}.
    \label{tab:dtu cd}
   \resizebox{0.7\linewidth}{!}{
    \begin{tabular}{c ccccc ccccc ccccc c}
    \toprule
        DTU & 24 & 37 & 40 & 55 & 63 & 65 & 69 & 83 & 97 & 105 & 106 & 110 & 114 & 118 & 122 & Mean\\
    \midrule
        VolSDF \cite{Yariv2021Volume} & 1.14 & 1.26 & 0.81 & 0.49 & 1.25 & 0.70 & 0.72 & 1.29 & 1.18 & 0.70 & 0.66 & 1.08 & 0.42 & 0.61 & 0.55 & 0.86 \\
        NeuS \cite{Wang2021NeuS} & 1.00 &	1.37 &	0.93 &	0.43 &	1.10 &	0.65 &	0.57 &	1.48 &	1.09 &	0.83 &	0.52 &	1.20 &	0.35 &	0.49 & 0.54 & 0.84 \\
        HF-NeuS \cite{Wang2022HFNeuS} & 0.76 & 1.32 & 0.70 & 0.39 & 1.06 & 0.63 & 0.63 & 1.15 & 1.12 & 0.80 & 0.52 & 1.22 & 0.33 & 0.49 & 0.50 & 0.77 \\
        Neuralangelo \cite{Li2023Neuralangelo} & 0.37 & 0.72 & 0.35 & 0.35 & 0.87 & 0.54 & 0.53 & 1.29 & 0.97 & 0.73 & 0.47 & 0.74 & 0.32 & 0.41 & 0.43 & 0.61 \\
        Ours & 0.94 &	0.87 &	0.75 &	0.39 &	1.05 &	0.63 &	0.58 &	1.34 &	1.01 &	0.96 &	0.50 &	1.19 &	0.36 &	0.46 &	0.49 & 0.77 \\
    \bottomrule
    \end{tabular}}
\end{table*}

\begin{figure*}[t]
\centering
\setlength{\tabcolsep}{2pt}
\begin{tabular}{ccccc}
    \includegraphics[width=0.180\linewidth, page=1]{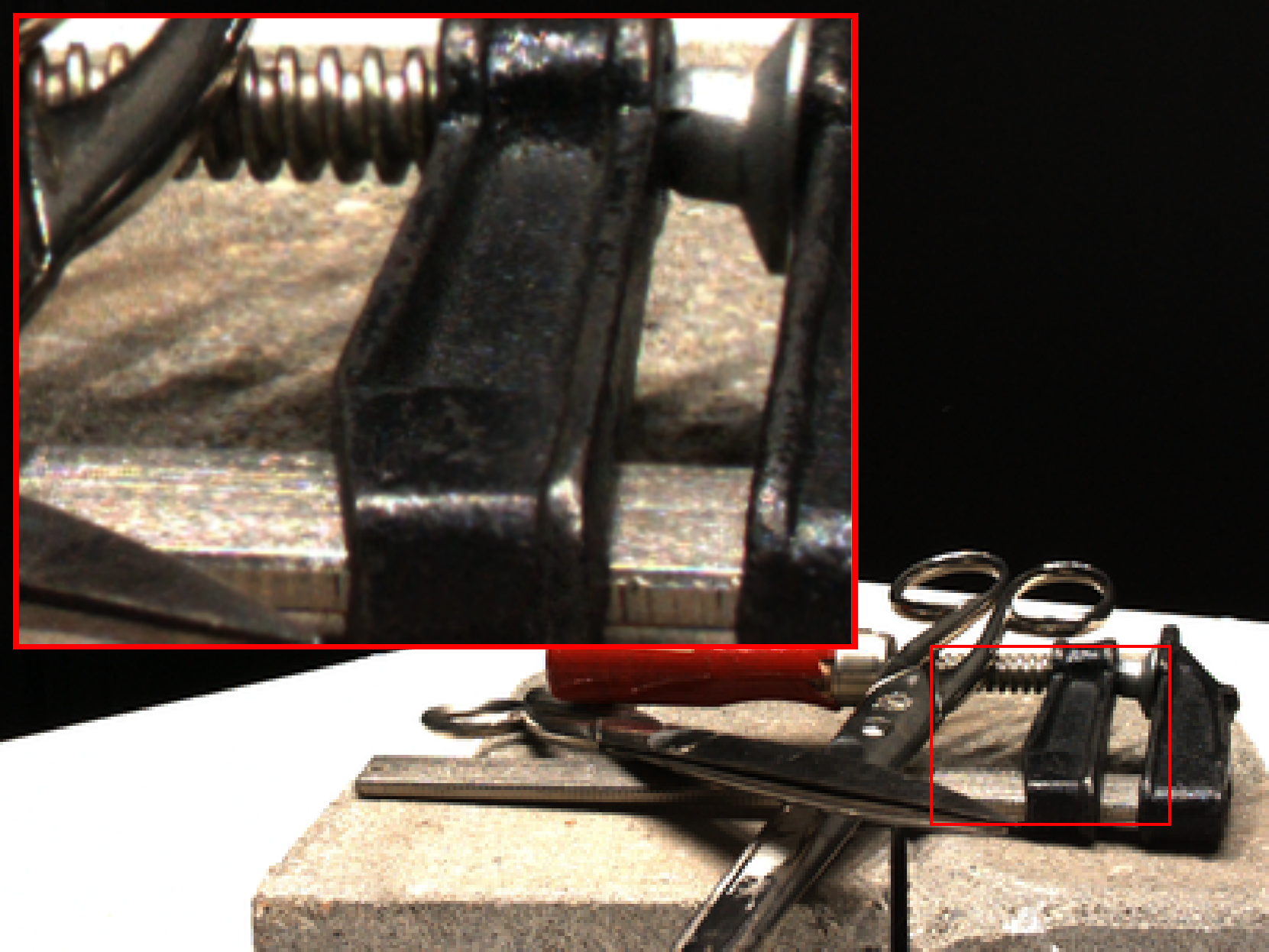} & \includegraphics[width=0.178\linewidth, page=3]{figs/exp/dtu/dtu_scan37/043/neus} & \includegraphics[width=0.178\linewidth, page=3]{figs/exp/dtu/dtu_scan37/043/volsdf} & \includegraphics[width=0.178\linewidth, page=3]{figs/exp/dtu/dtu_scan37/043/hfs} & \includegraphics[width=0.178\linewidth, page=3]{figs/exp/dtu/dtu_scan37/043/mpfs1} \\
    \begin{minipage}[b][0.0834\linewidth][c]{0.180\linewidth}\centering Viewed from back\end{minipage} & \includegraphics[width=0.180\linewidth, page=3]{figs/exp/dtu/dtu_scan37/extra00/neus} & \includegraphics[width=0.180\linewidth, page=3]{figs/exp/dtu/dtu_scan37/extra00/volsdf} & \includegraphics[width=0.180\linewidth, page=3]{figs/exp/dtu/dtu_scan37/extra00/hfs} & \includegraphics[width=0.180\linewidth, page=3]{figs/exp/dtu/dtu_scan37/extra00/mpfs1} \\[0.15in]
    \includegraphics[width=0.180\linewidth, page=1]{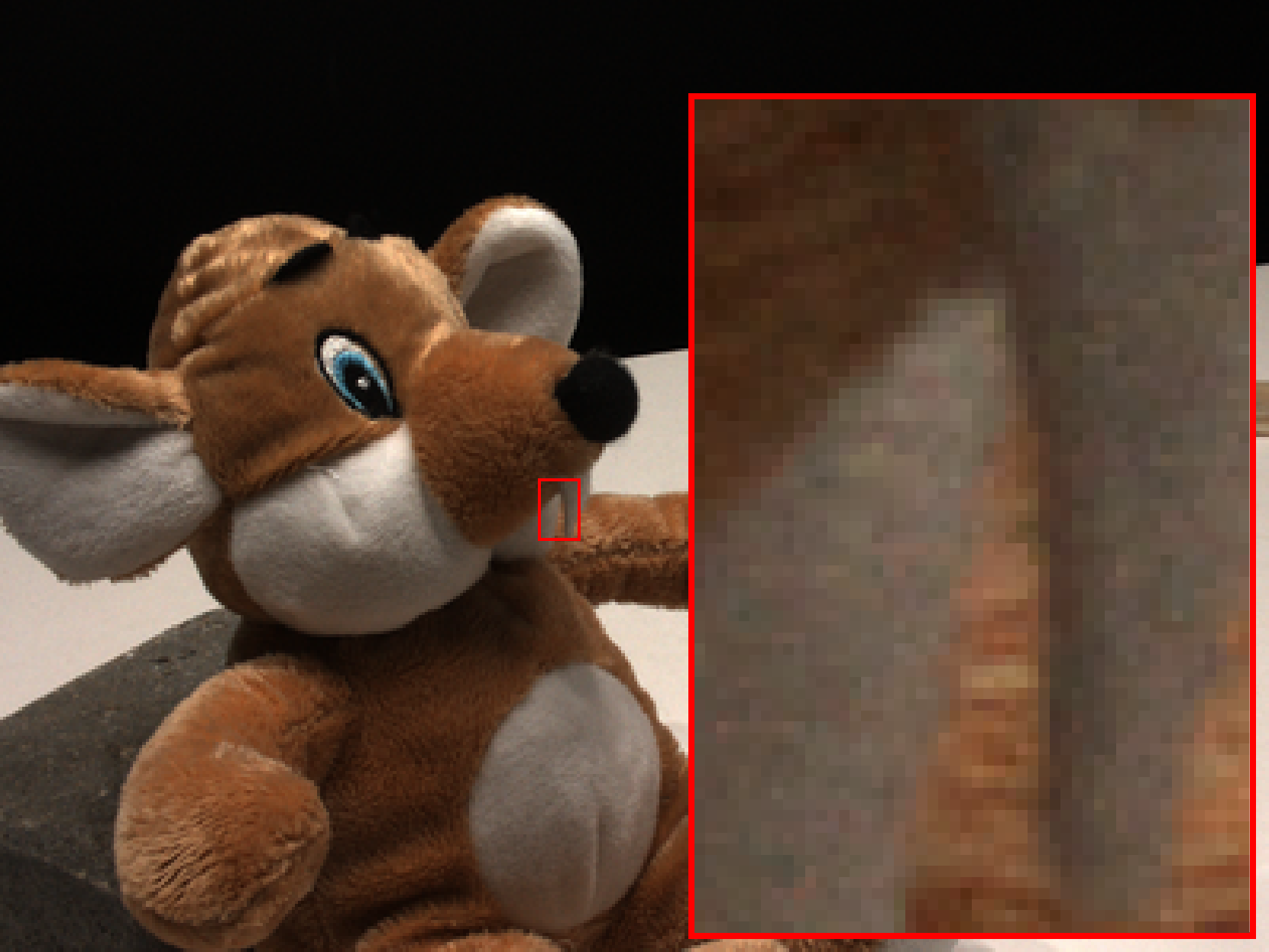} & \includegraphics[width=0.091\linewidth, page=3]{figs/exp/dtu/dtu_scan105/028/neus} & \includegraphics[width=0.091\linewidth, page=3]{figs/exp/dtu/dtu_scan105/028/volsdf} & \includegraphics[width=0.091\linewidth, page=3]{figs/exp/dtu/dtu_scan105/028/hfs} & \includegraphics[width=0.091\linewidth, page=3]{figs/exp/dtu/dtu_scan105/028/mpfs1} \\
    \begin{minipage}[b][0.120\linewidth][c]{0.180\linewidth}\centering Viewed from\\opposite side\end{minipage} & \includegraphics[width=0.120\linewidth, page=3]{figs/exp/dtu/dtu_scan105/extra00/neus} & \includegraphics[width=0.120\linewidth, page=3]{figs/exp/dtu/dtu_scan105/extra00/volsdf} & \includegraphics[width=0.120\linewidth, page=3]{figs/exp/dtu/dtu_scan105/extra00/hfs} & \includegraphics[width=0.120\linewidth, page=3]{figs/exp/dtu/dtu_scan105/extra00/mpfs1} \\
    & NeuS & VolSDF & HF-NeuS & Ours
\end{tabular}
\caption{Visualizations and local enlarged views of the reconstructed results on the DTU dataset~\cite{Anaaes2016Large}. For each showcased data, we present an enlarged local rendering that corresponds to the input camera view in the upper row. We render the same part of the mesh of each method in a novel angle that can better demonstrate the difference of each method in the lower row. In the top two rows, scissors and screwdrivers are stacked on top a set of bricks. All methods except our method reconstructs spurious structures that connect the tools to the surface below. In the bottom two rows, the teeth of the toy bear represents a layer of thin surface, and all methods except our method cannot faithfully reconstruct the thin plane structure and connects to the face behind the teeth. Our method is more flexible with representation of the implicit geometry and produces the best result at preventing complex structures from connecting together, as well as reconstructing thin surfaces.}
\label{fig:dtu}
\end{figure*}

\begin{figure}
\centering
\setlength{\tabcolsep}{2pt}
\begin{tabular}{ccc}
    \includegraphics[width=0.420\linewidth]{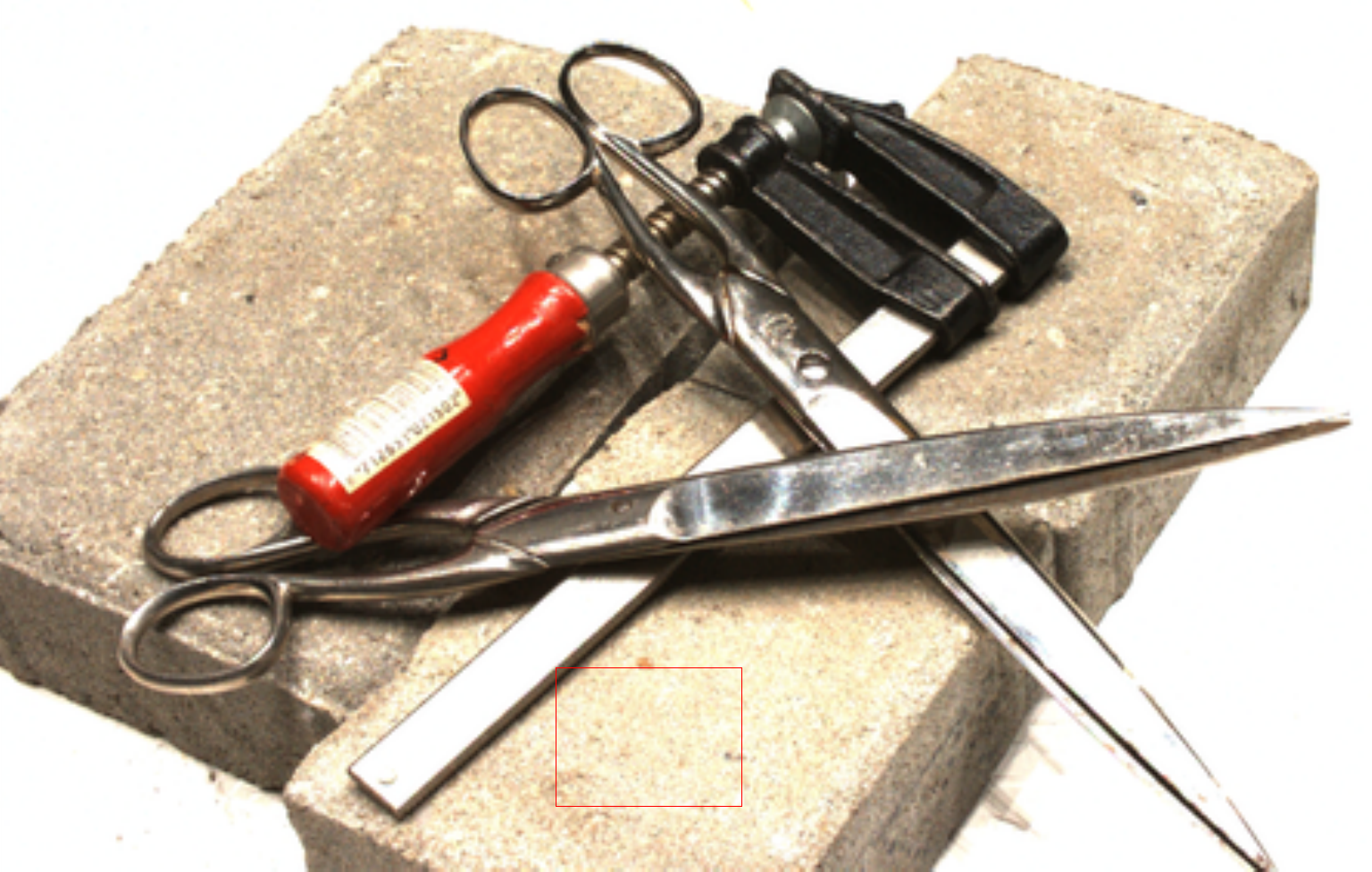} &
    \includegraphics[width=0.262\linewidth]{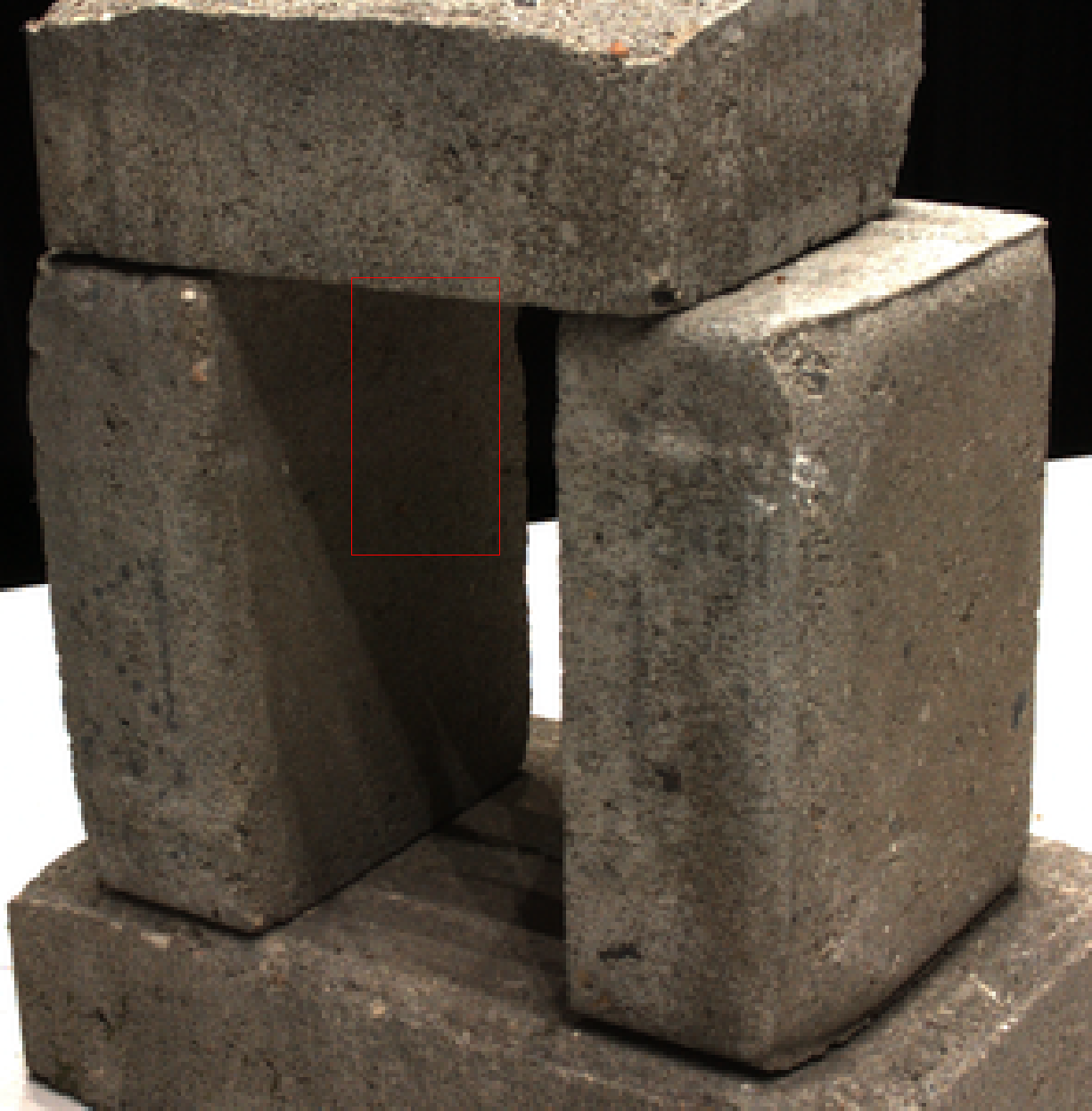} &
    \includegraphics[width=0.248\linewidth]{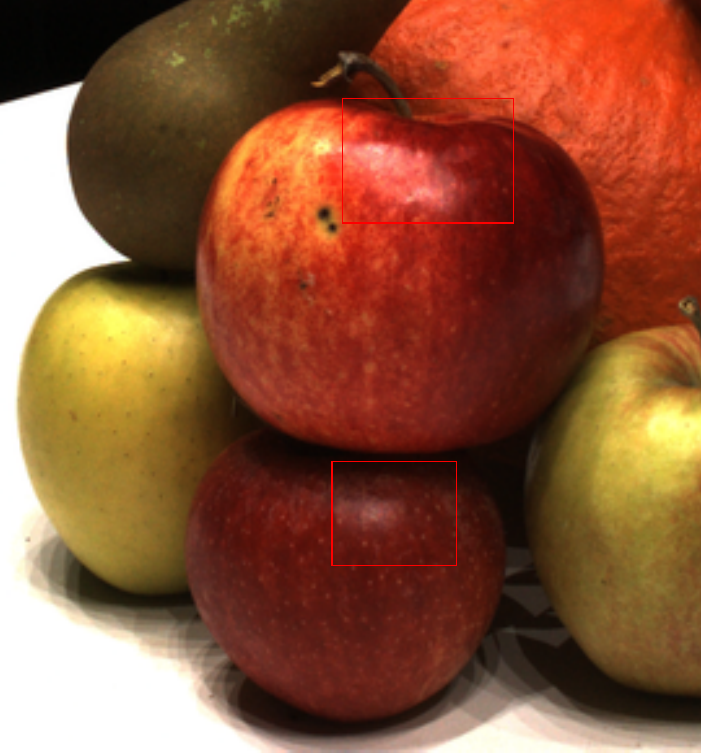} \\
    \includegraphics[width=0.420\linewidth]{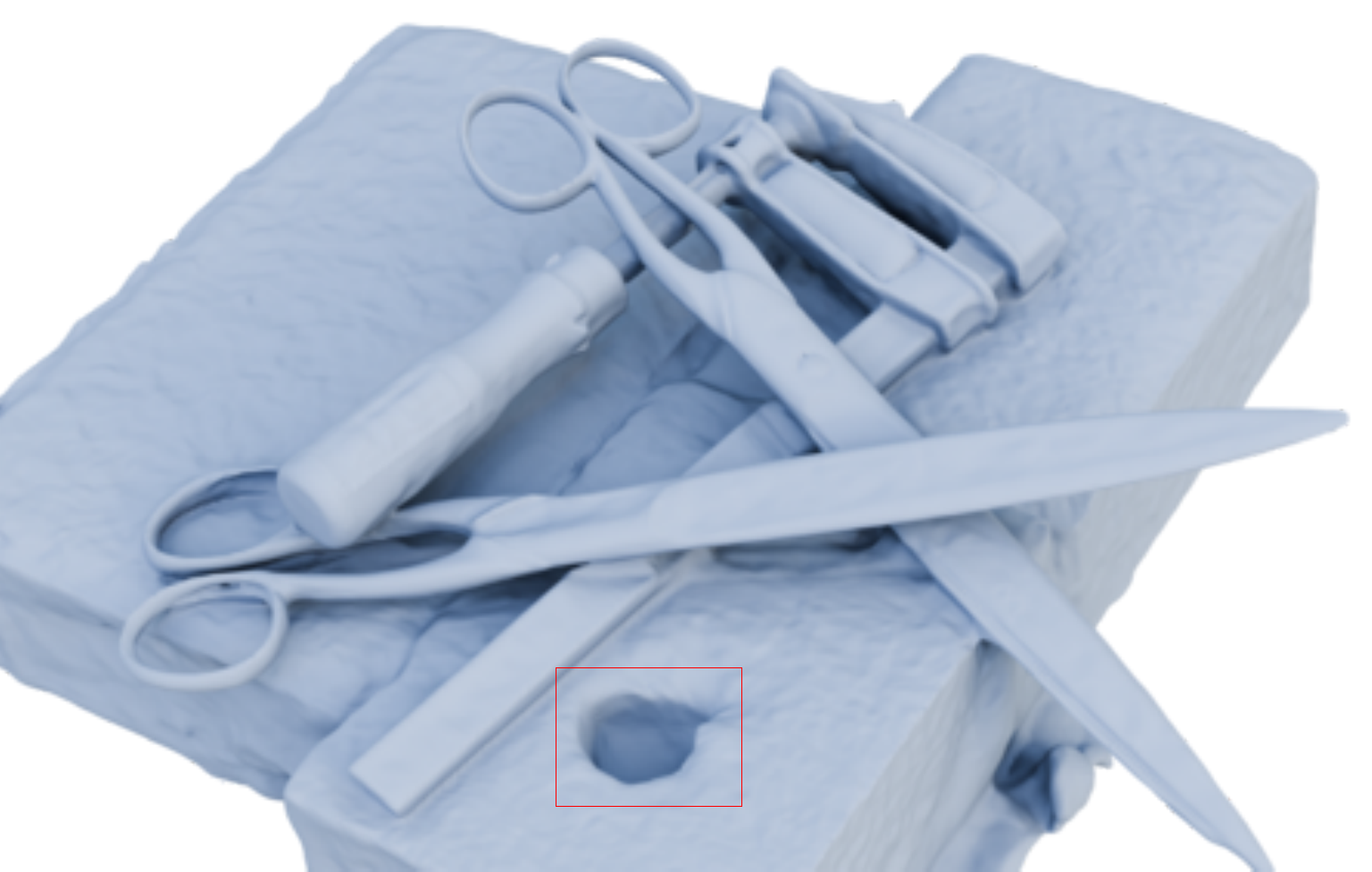} &
    \includegraphics[width=0.262\linewidth]{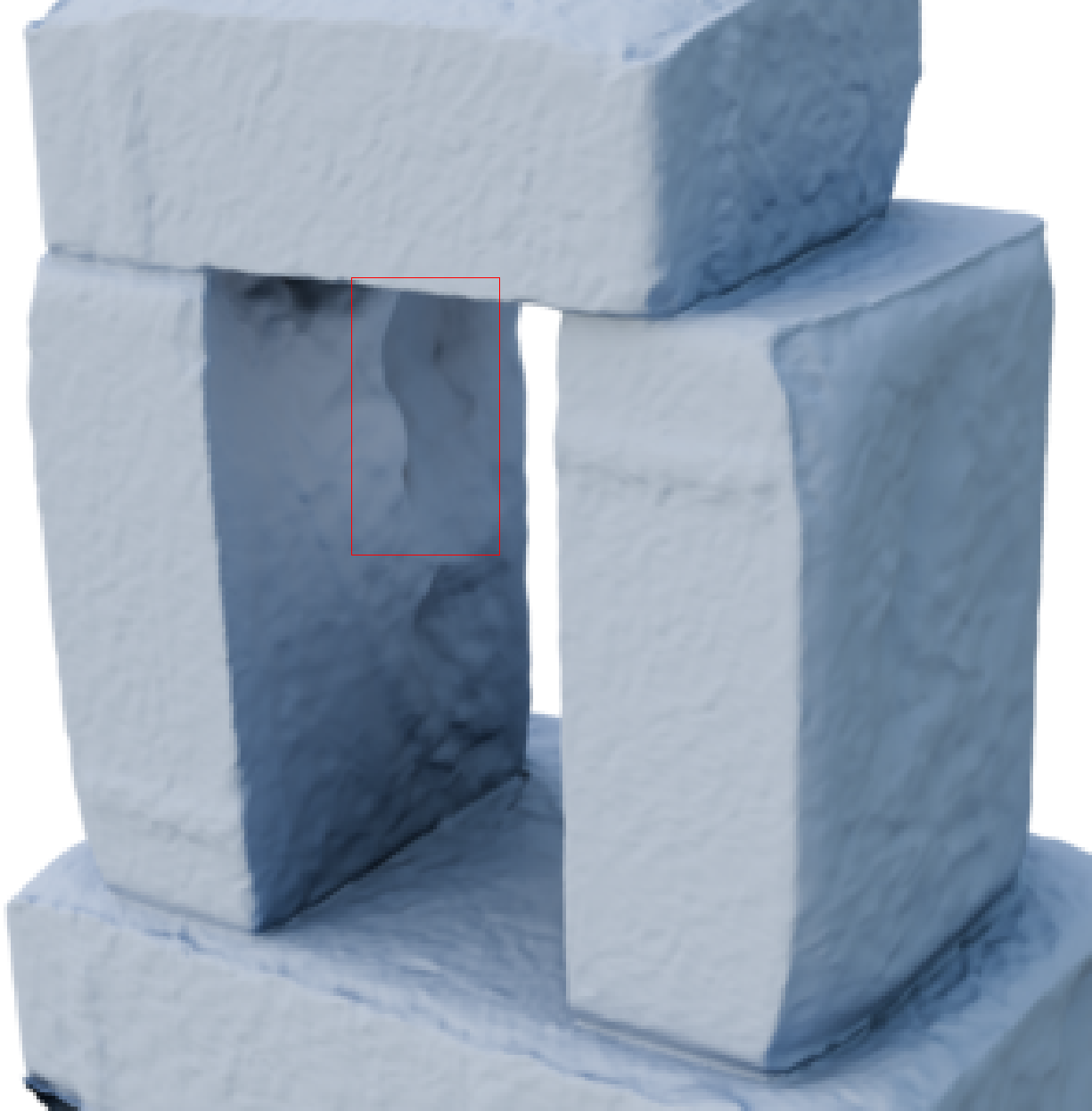} &
    \includegraphics[width=0.248\linewidth]{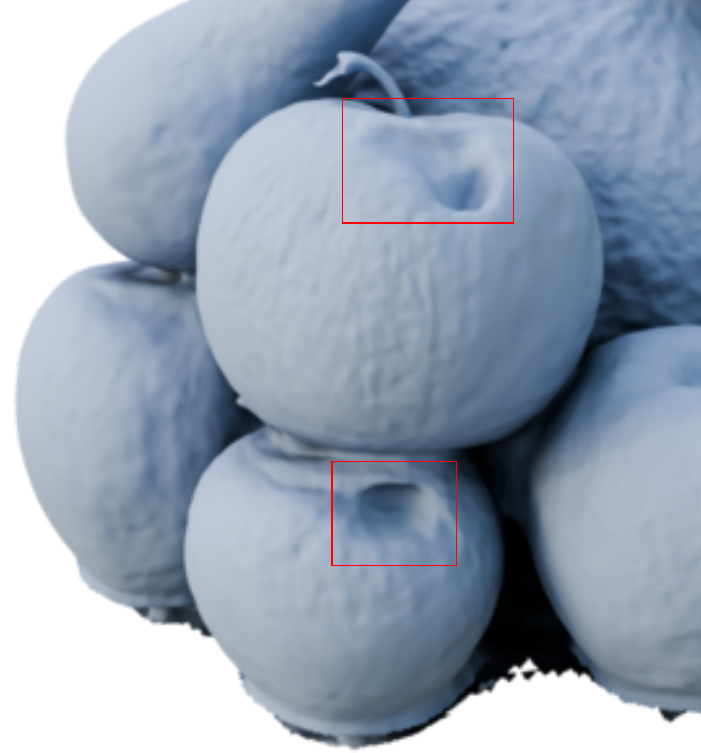} \\
\end{tabular}
\caption{Limitation: Our enhanced expressive capability may cause the reconstructed surface to have an increased tendency to dent under high specular lighting or shadow conditions, where color changes drastically at different viewing angles and lacking adequate color consistency cues.}
\label{fig:dtu-dent}
\end{figure}

\begin{figure*}[!htbp]
    \centering
\setlength\tabcolsep{2pt}
    \begin{tabular}{ccccccccc}
    Reference & NeuS & HF-NeuS & \scalebox{0.9}{Neuralangelo} & PGSR & GeoSVR & \scalebox{0.9}{GaussianUDF} & GGGS & Ours \\
    \includegraphics[width=.1020\linewidth, page=1]{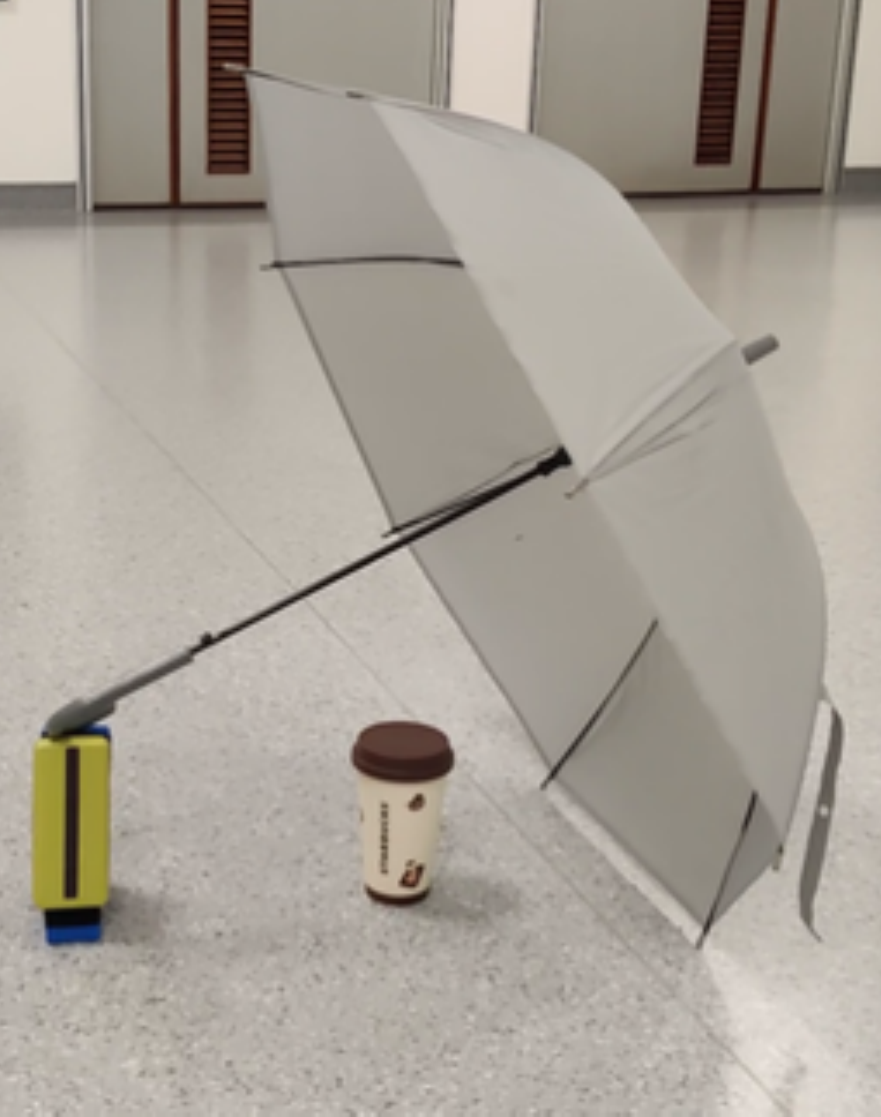} &
    \includegraphics[width=.1020\linewidth, page=2]{figs/exp/custom/umbrella/027/neus} &
    \includegraphics[width=.1020\linewidth, page=2]{figs/exp/custom/umbrella/027/hfs} &
    \includegraphics[width=.1020\linewidth, page=2]{figs/exp/custom/umbrella/027/neuralangelo} &
    \includegraphics[width=.1020\linewidth, page=2]{figs/exp/custom/umbrella/027/pgsr_post} &
    \includegraphics[width=.1020\linewidth, page=2]{figs/exp/custom/umbrella/027/geosvr_post} &
    \includegraphics[width=.1020\linewidth, page=2]{figs/exp/custom/umbrella/027/gaussianudf} &
    \includegraphics[width=.1020\linewidth, page=2]{figs/exp/custom/umbrella/027/gggs_post} &
    \includegraphics[width=.1020\linewidth, page=2]{figs/exp/custom/umbrella/027/mpfs2} \\
    \includegraphics[width=.0765\linewidth, page=2]{figs/exp/custom/umbrella/027/reference}\hfill\includegraphics[width=.0190\linewidth, page=3]{figs/exp/custom/umbrella/027/reference} &
    \includegraphics[width=.0765\linewidth, page=3]{figs/exp/custom/umbrella/027/neus}\hfill\includegraphics[width=.0205\linewidth, page=4]{figs/exp/custom/umbrella/027/neus} &
    \includegraphics[width=.0765\linewidth, page=3]{figs/exp/custom/umbrella/027/hfs}\hfill\includegraphics[width=.0205\linewidth, page=4]{figs/exp/custom/umbrella/027/hfs} &
    \includegraphics[width=.0765\linewidth, page=3]{figs/exp/custom/umbrella/027/neuralangelo}\hfill\includegraphics[width=.0205\linewidth, page=4]{figs/exp/custom/umbrella/027/neuralangelo} &
    \includegraphics[width=.0765\linewidth, page=3]{figs/exp/custom/umbrella/027/pgsr_post}\hfill\includegraphics[width=.0205\linewidth, page=4]{figs/exp/custom/umbrella/027/pgsr_post} &
    \includegraphics[width=.0765\linewidth, page=3]{figs/exp/custom/umbrella/027/geosvr_post}\hfill\includegraphics[width=.0205\linewidth, page=4]{figs/exp/custom/umbrella/027/geosvr_post} &
    \includegraphics[width=.0765\linewidth, page=3]{figs/exp/custom/umbrella/027/gaussianudf}\hfill\includegraphics[width=.0205\linewidth, page=4]{figs/exp/custom/umbrella/027/gaussianudf} &
    \includegraphics[width=.0765\linewidth, page=3]{figs/exp/custom/umbrella/027/gggs_post}\hfill\includegraphics[width=.0205\linewidth, page=4]{figs/exp/custom/umbrella/027/gggs_post} &
    \includegraphics[width=.0765\linewidth, page=3]{figs/exp/custom/umbrella/027/mpfs2}\hfill\includegraphics[width=.0205\linewidth, page=4]{figs/exp/custom/umbrella/027/mpfs2} \\
    \includegraphics[width=.1020\linewidth, page=1]{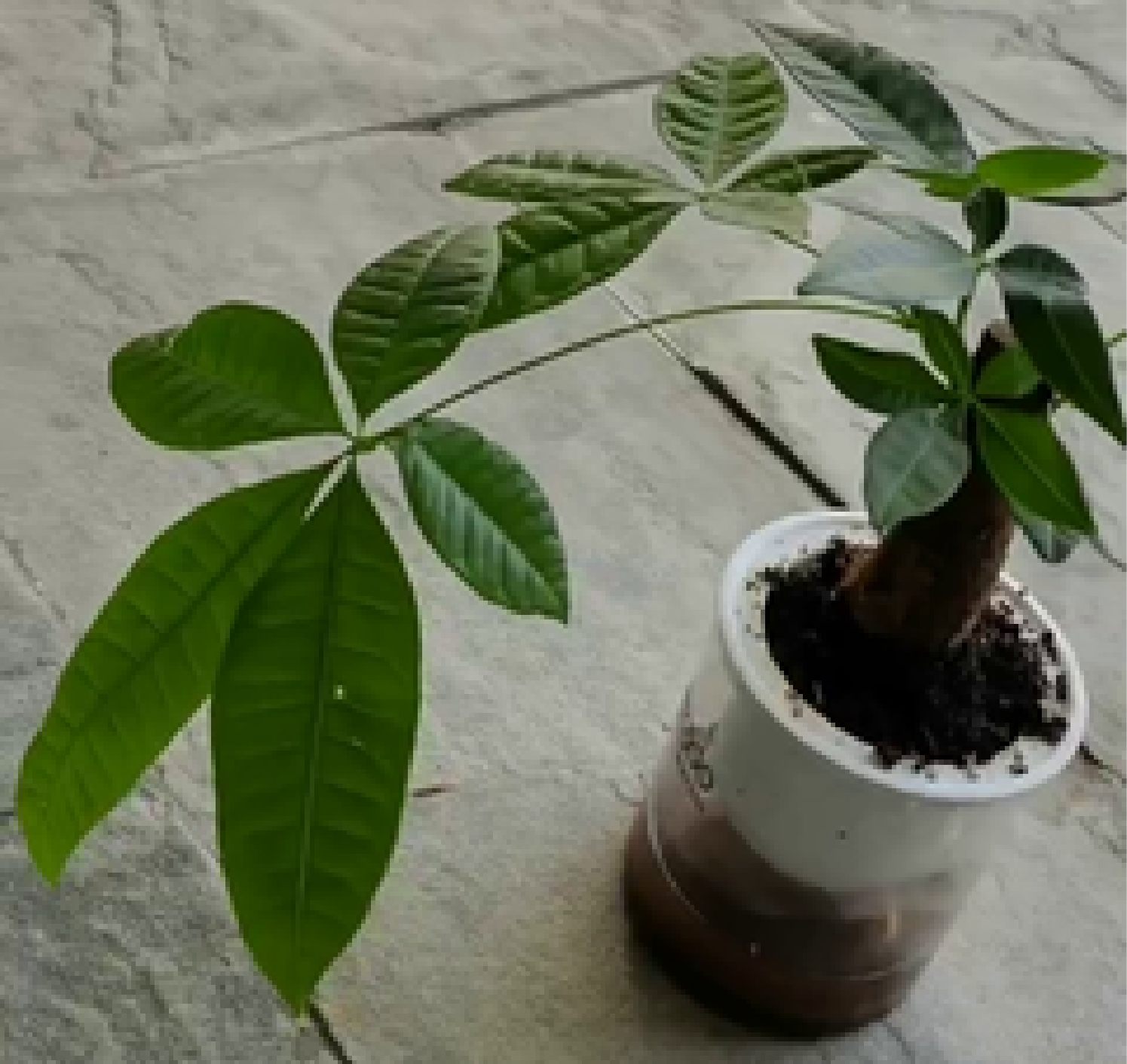} &
    \includegraphics[width=.1020\linewidth, page=2]{figs/exp/neudf/real_capture_plant/152/neus} &
    \includegraphics[width=.1020\linewidth, page=2]{figs/exp/neudf/real_capture_plant/152/hfs} &
    \includegraphics[width=.1020\linewidth, page=2]{figs/exp/neudf/real_capture_plant/152/neuralangelo} &
    \includegraphics[width=.1020\linewidth, page=2]{figs/exp/neudf/real_capture_plant/152/pgsr} &
    \includegraphics[width=.1020\linewidth, page=2]{figs/exp/neudf/real_capture_plant/152/geosvr} &
    \includegraphics[width=.1020\linewidth, page=2]{figs/exp/neudf/real_capture_plant/152/gaussianudf} &
    \includegraphics[width=.1020\linewidth, page=2]{figs/exp/neudf/real_capture_plant/152/gggs} &
    \includegraphics[width=.1020\linewidth, page=2]{figs/exp/neudf/real_capture_plant/152/mpfs2} \\
    \includegraphics[width=.1020\linewidth, page=2]{figs/exp/neudf/real_capture_plant/152/reference} &
    \includegraphics[width=.1020\linewidth, page=3]{figs/exp/neudf/real_capture_plant/152/neus} &
    \includegraphics[width=.1020\linewidth, page=3]{figs/exp/neudf/real_capture_plant/152/hfs} &
    \includegraphics[width=.1020\linewidth, page=3]{figs/exp/neudf/real_capture_plant/152/neuralangelo} &
    \includegraphics[width=.1020\linewidth, page=3]{figs/exp/neudf/real_capture_plant/152/pgsr} &
    \includegraphics[width=.1020\linewidth, page=3]{figs/exp/neudf/real_capture_plant/152/geosvr} &
    \includegraphics[width=.1020\linewidth, page=3]{figs/exp/neudf/real_capture_plant/152/gaussianudf} &
    \includegraphics[width=.1020\linewidth, page=3]{figs/exp/neudf/real_capture_plant/152/gggs} &
    \includegraphics[width=.1020\linewidth, page=3]{figs/exp/neudf/real_capture_plant/152/mpfs2} \\
    \includegraphics[width=.1020\linewidth]{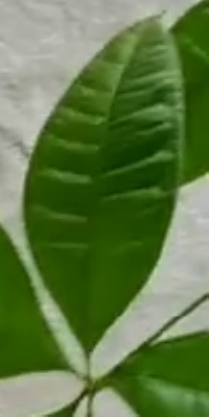} & 
    \includegraphics[width=.1020\linewidth]{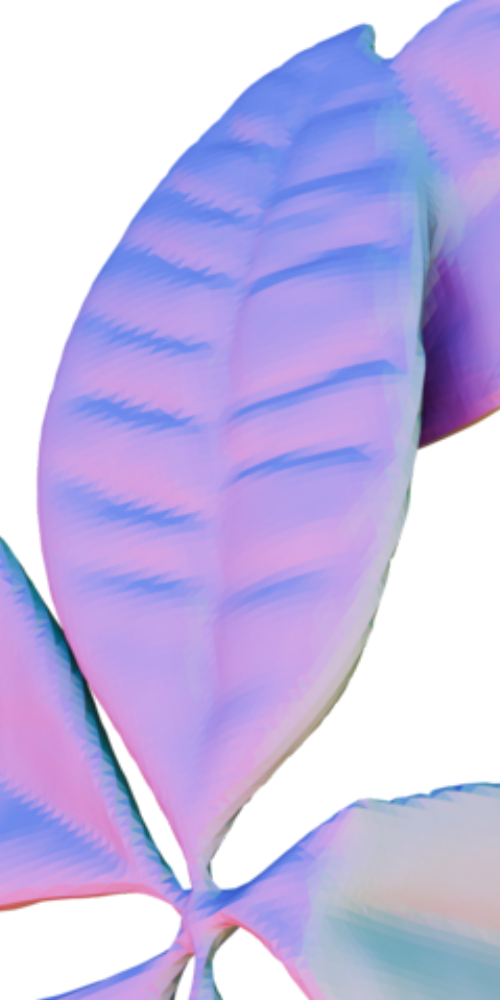} &
    \includegraphics[width=.1020\linewidth]{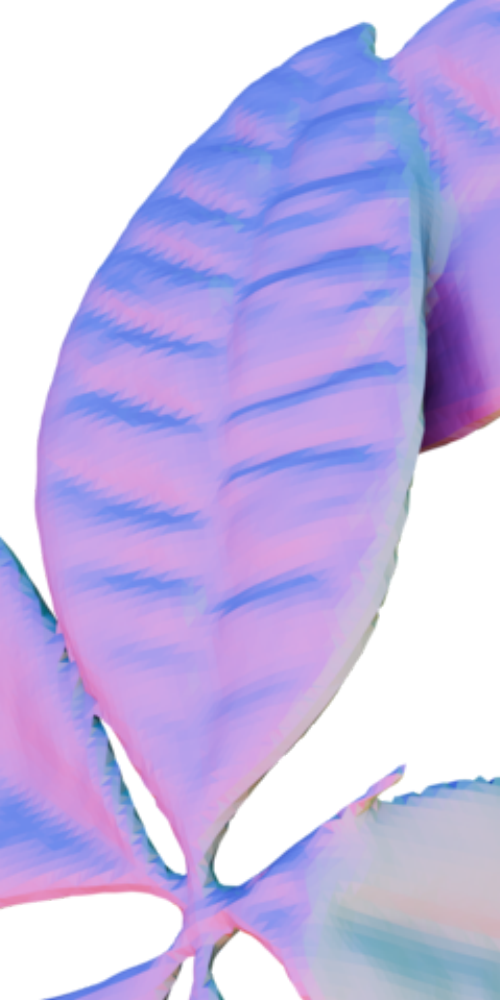} &
    \includegraphics[width=.1020\linewidth]{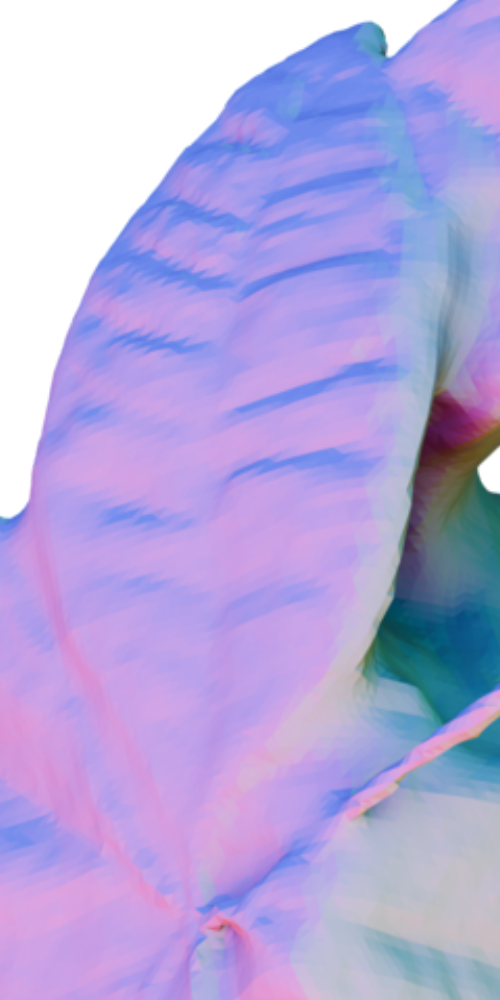} &
    \includegraphics[width=.1020\linewidth]{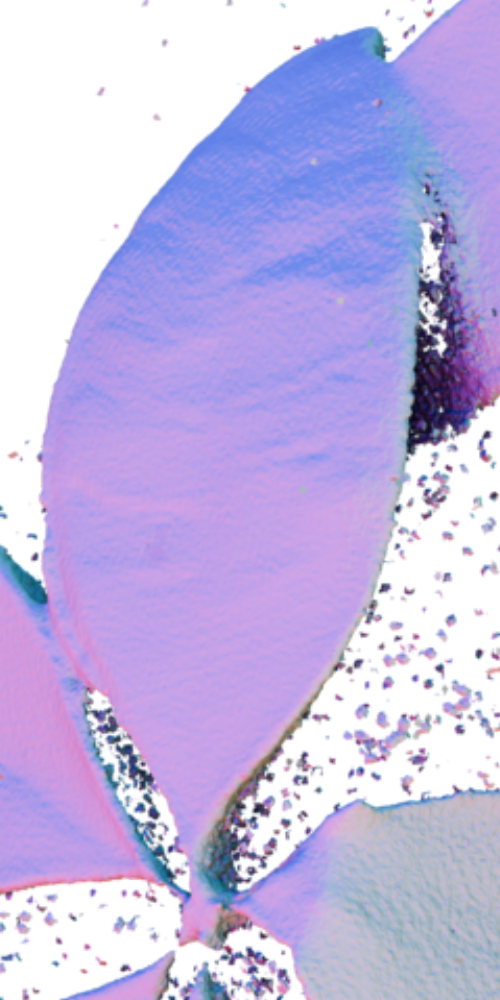} &
    \includegraphics[width=.1020\linewidth]{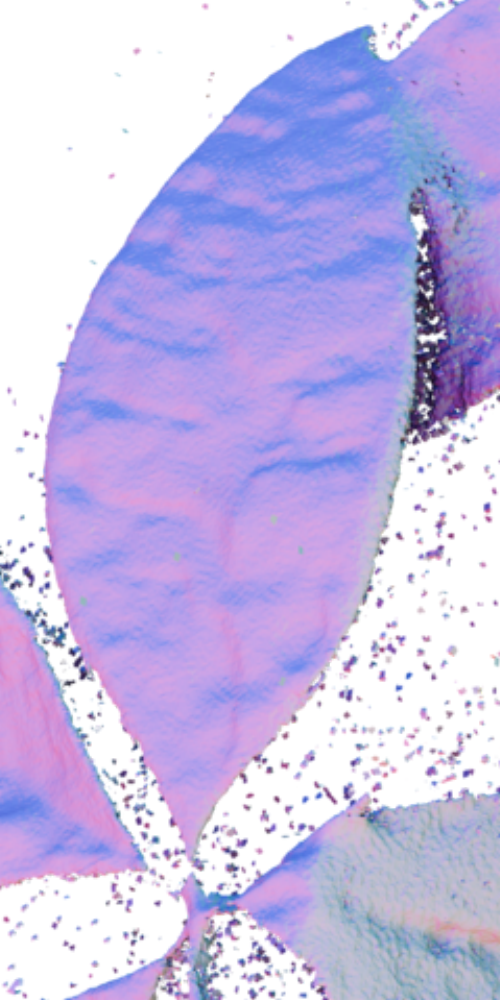} &
    \includegraphics[width=.1020\linewidth]{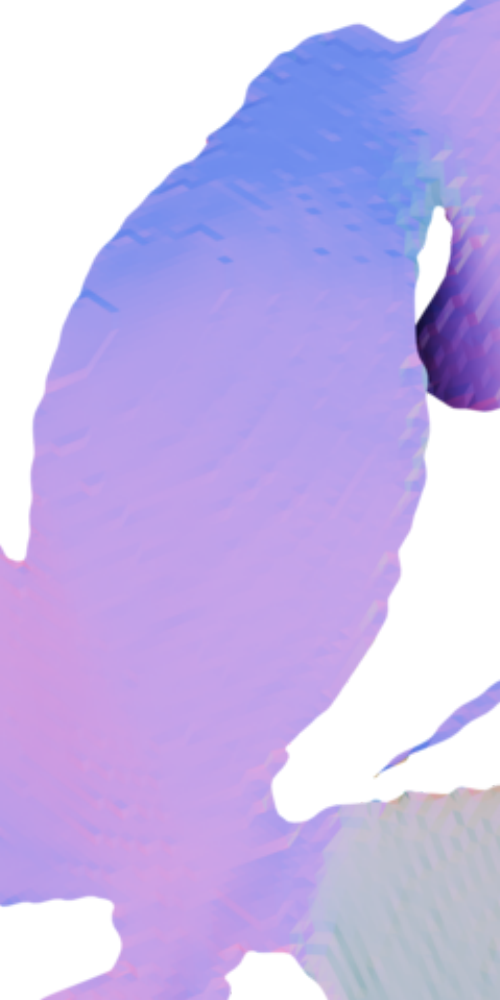} &
    \includegraphics[width=.1020\linewidth]{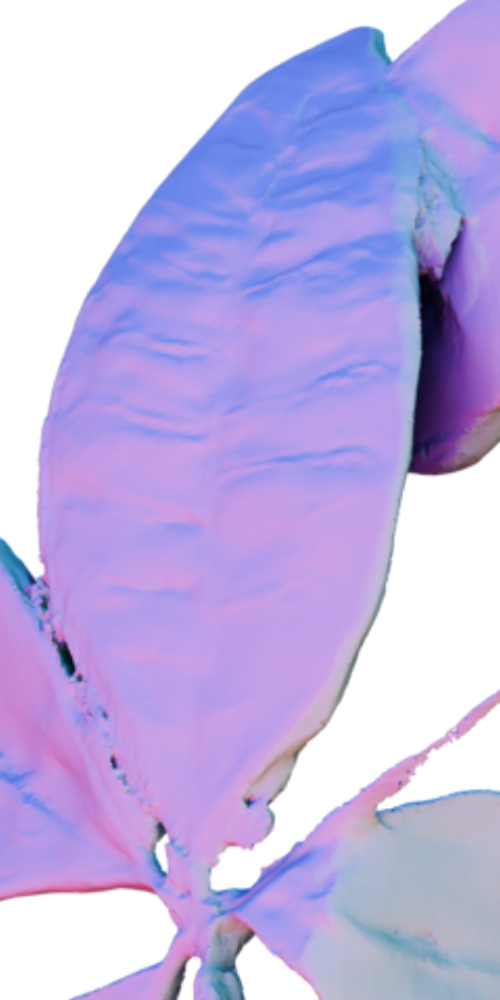} &
    \includegraphics[width=.1020\linewidth]{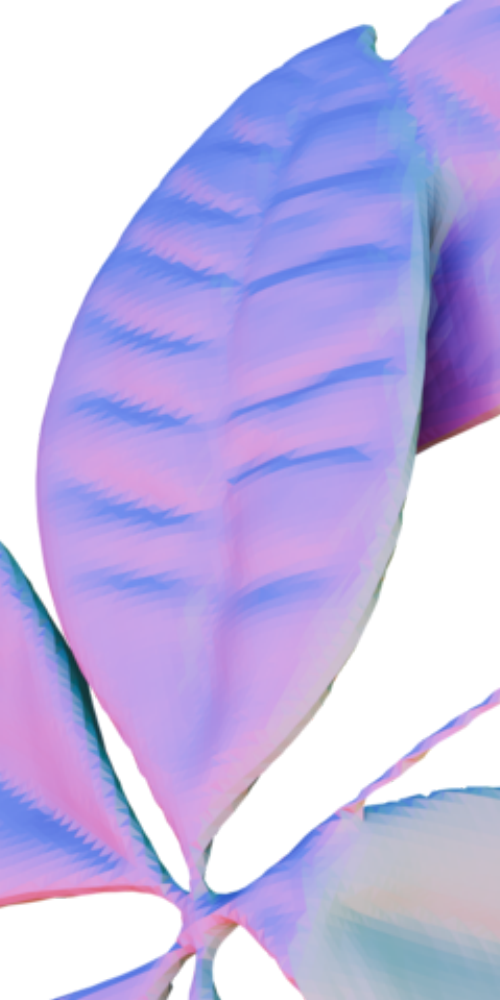} \\
    \includegraphics[width=.1020\linewidth, page=1]{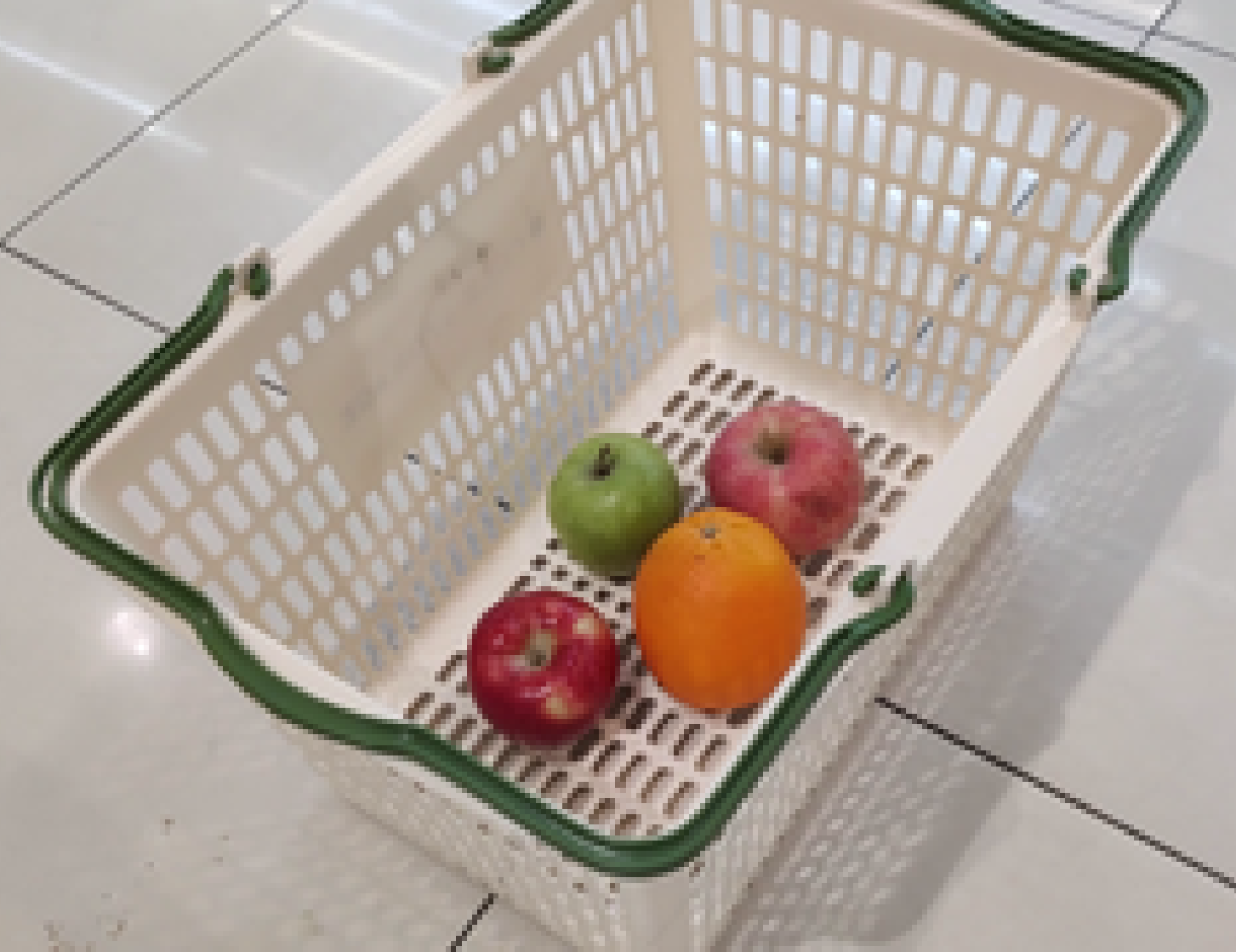} &
    \includegraphics[width=.1020\linewidth, page=2]{figs/exp/custom/basket/075/neus} &
    \includegraphics[width=.1020\linewidth, page=2]{figs/exp/custom/basket/075/hfs} &
    \includegraphics[width=.1020\linewidth, page=2]{figs/exp/custom/basket/075/neuralangelo} &
    \includegraphics[width=.1020\linewidth, page=2]{figs/exp/custom/basket/075/pgsr_post} &
    \includegraphics[width=.1020\linewidth, page=2]{figs/exp/custom/basket/075/geosvr_post} &
    \includegraphics[width=.1020\linewidth, page=2]{figs/exp/custom/basket/075/gaussianudf} &
    \includegraphics[width=.1020\linewidth, page=2]{figs/exp/custom/basket/075/gggs_post} &
    \includegraphics[width=.1020\linewidth, page=2]{figs/exp/custom/basket/075/mpfs1} \\
    \includegraphics[width=.1020\linewidth, page=2]{figs/exp/custom/basket/075/reference} &
    \includegraphics[width=.1020\linewidth, page=3]{figs/exp/custom/basket/075/neus} &
    \includegraphics[width=.1020\linewidth, page=3]{figs/exp/custom/basket/075/hfs} &
    \includegraphics[width=.1020\linewidth, page=3]{figs/exp/custom/basket/075/neuralangelo} &
    \includegraphics[width=.1020\linewidth, page=3]{figs/exp/custom/basket/075/pgsr_post} &
    \includegraphics[width=.1020\linewidth, page=3]{figs/exp/custom/basket/075/geosvr_post} &
    \includegraphics[width=.1020\linewidth, page=3]{figs/exp/custom/basket/075/gaussianudf} &
    \includegraphics[width=.1020\linewidth, page=3]{figs/exp/custom/basket/075/gggs_post} &
    \includegraphics[width=.1020\linewidth, page=3]{figs/exp/custom/basket/075/mpfs1} \\
    \includegraphics[width=.1020\linewidth, page=1]{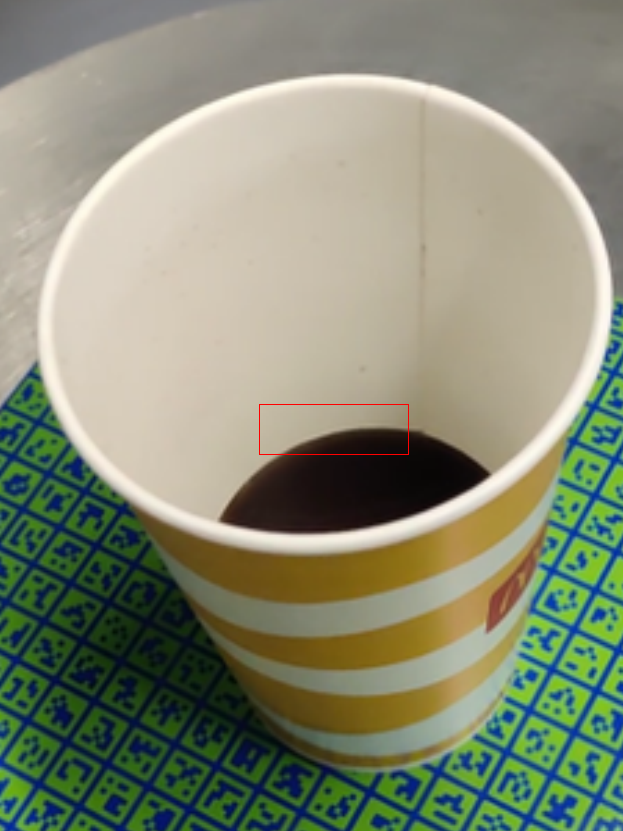} & \includegraphics[width=.1020\linewidth, page=2]{figs/exp/custom/cup/050/neus} &  \includegraphics[width=.1020\linewidth, page=2]{figs/exp/custom/cup/050/hfs} &  \includegraphics[width=.1020\linewidth, page=2]{figs/exp/custom/cup/050/neuralangelo} & \includegraphics[width=.1020\linewidth, page=2]{figs/exp/custom/cup/050/pgsr_post} & 
    \includegraphics[width=.1020\linewidth, page=2]{figs/exp/custom/cup/050/geosvr_post} &
    \includegraphics[width=.1020\linewidth, page=2]{figs/exp/custom/cup/050/gaussianudf} & \includegraphics[width=.1020\linewidth, page=2]{figs/exp/custom/cup/050/gggs_post} & \includegraphics[width=.1020\linewidth, page=2]{figs/exp/custom/cup/050/mpfs1.pdf} \\
    \includegraphics[width=.1020\linewidth, page=2]{figs/exp/custom/cup/050/reference} & \includegraphics[width=.1020\linewidth, page=3]{figs/exp/custom/cup/050/neus} &  \includegraphics[width=.1020\linewidth, page=3]{figs/exp/custom/cup/050/hfs} &  \includegraphics[width=.1020\linewidth, page=3]{figs/exp/custom/cup/050/neuralangelo} & \includegraphics[width=.1020\linewidth, page=3]{figs/exp/custom/cup/050/pgsr_post} & 
    \includegraphics[width=.1020\linewidth, page=3]{figs/exp/custom/cup/050/geosvr_post} &
    \includegraphics[width=.1020\linewidth, page=3]{figs/exp/custom/cup/050/gaussianudf} & \includegraphics[width=.1020\linewidth, page=3]{figs/exp/custom/cup/050/gggs_post} & \includegraphics[width=.1020\linewidth, page=3]{figs/exp/custom/cup/050/mpfs1.pdf} \\
    Reference & NeuS & HF-NeuS & \scalebox{0.9}{Neuralangelo} & PGSR & GeoSVR & \scalebox{0.9}{GaussianUDF} & GGGS & Ours \\
    \end{tabular}   
   \caption{ Results on custom dataset for the real captured test cases. The umbrella scene contains thin structures, such as the umbrella rod and the retention strap, highlighted by colored boxes. 
The flower scene features open boundaries (e.g., petals), thin structures (e.g., the stem), and fine-scale geometric details (e.g., leaf veins). The basket scene contains dense thin lattice structures formed by narrow grid bars on the basket walls and base.
NeuS and 3DGS-based baselines exhibit missing or fragmented geometry in these thin regions and struggle to recover delicate venation patterns.  Finally, the cup scene is an open model and many baseline methods produce extravagant structures in such open scenes. 
In contrast, our method reconstructs these structures more completely and consistently, producing sharper and more faithful fine details. 
}
    \label{fig:results custom real}
\end{figure*}

\paragraph{Results on our dataset} Figure~\ref{fig:results custom real} shows comparison results on the real scenes from our TSSA dataset. SDF-based methods often suffer from spurious closures when reconstructing sign-ambiguous geometry, whereas UDF-based approaches generally recover the correct topology but may struggle with stable optimization and surface extraction. In comparison, \method{} consistently achieves higher reconstruction accuracy according to the quantitative metrics. 3DGS-based methods perform well on both closed and open surfaces in terms of rendering quality; however, they often struggle with thin or closely spaced structures, sometimes producing missing parts on TSSA scenes and introducing many floating artifacts that do not correspond to real geometry. In contrast, \method{} demonstrates strong robustness to thin and sign-ambiguous geometry, yielding stable and reliable reconstructions across the TSSA dataset.

Furthermore, Figure~\ref{fig:results custom syn} shows comparison results on the synthetic scenes from our TSSA dataset. Our method also produces significantly better geometric results. In these cases, SDF-based methods such as NeuS often produce spurious closures or thickness inflation in sign-ambiguous regions, while UDF-based methods may suffer from unstable optimization and fragmented surfaces. 3DGS baselines frequently generate incomplete or noisy meshes after surface extraction.

Take the LEGO scene as an example (Figure~\ref{fig:results custom syn} row 1). The conveyor belt contains many small circular openings, which introduce significant sign ambiguity due to the thin plate structure and closely spaced boundaries. NeuS and PGSR fail to recover most of these holes and instead produce overly smoothed surfaces that close the openings. In addition, PGSR introduces an unwanted planar surface on the inner side of the bucket (highlighted by the green box), which does not exist in the ground-truth geometry. GaussianUDF struggles to reconstruct the overall structure and produces fragmented geometry. GGGS yields better hole recovery than other baselines, but the reconstructed surface remains noisy and lacks smoothness. In contrast, our method accurately reconstructs the thin plate and preserves the hole openings with clean and well-defined boundaries.

We also report the quantitative geometric results (CD and NC) in Table~\ref{tab:geo_case_results_cd} and Table~\ref{tab:geo_case_results_nc} for synthetic cases, where the ground truth mesh is available. The quantitative metrics further confirm that our method consistently reconstructs complete and coherent geometry in these challenging scenarios. This improvement stems from the MPF representation, which decouples metric proximity from sign information and enables occlusion-aware rendering while preserving UDF-like flexibility in sign-ambiguous regions.

\begin{table*}[!htbp]
\centering
\caption{Geometry reconstruction results on the TAAS-Synthetic dataset measured by Chamfer Distance (CD, in $10^{-3}$ units). We report the directional scores from ground truth to prediction (gt→pred), from prediction to ground truth (pred→gt), and their bidirectional mean. Lower CD values indicate better geometric accuracy. Best results are shown in \textbf{bold}, and second-best results are \underline{underlined}. ``NA'' indicates reconstruction failure.}
\label{tab:geo_case_results_cd}
\setlength{\tabcolsep}{4pt}

\renewcommand{\arraystretch}{1.05}

\begin{tabular}{lcccccc}
\toprule
\multirow{2}{*}{Method}
& \multicolumn{3}{c}{High-Heeled Shoe}
& \multicolumn{3}{c}{Tank} \\
\cmidrule(lr){2-4}
\cmidrule(lr){5-7}
& mean $(\downarrow)$ & pred→gt $(\downarrow)$& gt→pred $(\downarrow)$
& mean $(\downarrow)$  & pred→gt $(\downarrow)$& gt→pred $(\downarrow)$\\
\midrule
GaussianUDF
& 10.319 & \underline{4.260} & 16.379
& 7.582 & 5.771 & 9.393 \\

GGGS
& \underline{6.299} & 6.021 & \underline{6.577}
& \underline{5.977} & \textbf{4.694} & \underline{7.260} \\

NeuS
& 9.707 & 8.936 & 10.477
& 6.442 & 4.925 & 7.959 \\

PGSR
& 7.607 & 6.591 & 8.622
& 8.294 & 5.266 & 11.322 \\

HF-NeuS
& 11.883 & 10.449 & 13.317
& 6.277 & 4.882 & 7.672 \\

Neuralangelo
& 11.224 & 4.959 & 17.489
& 10.087 & 5.691 & 14.483 \\

Ours
& \textbf{4.762} & \textbf{4.190} & \textbf{5.335}
& \textbf{4.985} & \underline{4.700} & \textbf{5.269} \\

\bottomrule
\end{tabular}

\vspace{6pt}

\begin{tabular}{lcccccc}
\toprule
\multirow{2}{*}{Method}
& \multicolumn{3}{c}{LEGO}
& \multicolumn{3}{c}{Figurine} \\
\cmidrule(lr){2-4}
\cmidrule(lr){5-7}
& mean $(\downarrow)$ & pred→gt $(\downarrow)$ & gt→pred $(\downarrow)$
& mean $(\downarrow)$ & pred→gt $(\downarrow)$ & gt→pred $(\downarrow)$ \\
\midrule
GaussianUDF 
& 7.693 & 5.490 & 9.895
& 9.038 & 5.590 & 12.486 \\

GGGS
& \underline{4.398} & \underline{4.455} & \underline{4.342}
& NA & NA & NA \\

NeuS 
& 5.826 & 5.626 & 6.026
& 10.280 & \underline{4.752} & 15.808 \\

PGSR 
& 4.771 & 4.900 & 4.642
& 7.516 & 8.373 & \textbf{6.659} \\

HF-NeuS
& 6.722 & 6.156 & 7.288
& 13.153 & 5.496 & 20.809 \\

Neuralangelo
& 5.862 & 5.973 & 5.750
& \underline{6.753} & 5.343 & 8.162 \\

Ours
& \textbf{4.079} & \textbf{4.280} & \textbf{3.877}
& \textbf{5.745} & \textbf{4.451} & \underline{7.039} \\

\bottomrule
\end{tabular}
\end{table*}

\begin{table*}
\centering
\caption{Geometry reconstruction results on the TAAS-Synthetic dataset measured by Normal Consistency (NC, \%). We report the directional scores from ground truth to prediction (gt→pred), from prediction to ground truth (pred→gt), and their bidirectional mean. Higher NC values indicate better geometric accuracy. Best results are shown in \textbf{bold}, and second-best results are \underline{underlined}. ``NA'' indicates reconstruction failure.}
\label{tab:geo_case_results_nc}

\begin{tabular}{lcccccc}
\toprule
\multirow{2}{*}{Method}
& \multicolumn{3}{c}{High-Heeled Shoe}
& \multicolumn{3}{c}{Tank} \\
\cmidrule(lr){2-4}
\cmidrule(lr){5-7}
& mean $(\uparrow)$ & pred→gt $(\uparrow)$ & gt→pred $(\uparrow)$
& mean $(\uparrow)$ & pred→gt $(\uparrow)$ & gt→pred $(\uparrow)$ \\
\midrule
GaussianUDF
& 89.49 & \underline{94.55} & 84.43
& 61.16 & 63.17 & 59.14 \\

GGGS
& 90.91 & 88.96 & \underline{92.87}
& 70.48 & 71.19 & 69.77 \\

NeuS
& 86.31 & 85.03 & 87.59
& 72.76 & 74.54 & 70.99 \\

PGSR
& \underline{91.46} & 92.03 & 90.88
& 65.09 & 69.17 & 61.02 \\

HF-NeuS
& 86.01 & 86.99 & 85.03
& \underline{73.80} & \underline{75.00} & \underline{72.60} \\

Neuralangelo
& 85.21 & 91.57 & 78.85
& 63.17 & 66.38 & 59.97 \\

Ours
& \textbf{96.49} & \textbf{96.46} & \textbf{96.53}
& \textbf{74.13} & \textbf{74.53} & \textbf{73.73} \\

\bottomrule
\end{tabular}

\vspace{6pt}

\begin{tabular}{lcccccc}
\toprule
\multirow{2}{*}{Method}
& \multicolumn{3}{c}{LEGO}
& \multicolumn{3}{c}{Figurine} \\
\cmidrule(lr){2-4}
\cmidrule(lr){5-7}
& mean $(\uparrow)$ & pred→gt $(\uparrow)$ & gt→pred $(\uparrow)$
& mean $(\uparrow)$ & pred→gt $(\uparrow)$ & gt→pred $(\uparrow)$\\
\midrule
GaussianUDF
& 50.21 & 48.75 & 51.67
& 60.77 & 49.47 & 72.08 \\

GGGS
& 63.81 & 64.78 & 62.83
& NA & NA & NA \\

NeuS
& 59.61 & 60.38 & 58.84
& 71.40 & \underline{88.24} & 54.55 \\

PGSR
& \underline{65.71} & \underline{65.72} & \underline{65.71}
& 79.73 & 81.89 & \textbf{77.57} \\

HF-NeuS
& 55.68 & 57.09 & 54.26
& 72.70 & \textbf{95.81} & 49.60 \\

Neuralangelo
& 61.41 & 60.62 & 62.20
& \underline{80.94} & 85.90 & 75.99 \\

Ours
& \textbf{71.20} & \textbf{71.37} & \textbf{71.02}
& \textbf{80.97} & 85.73 & \underline{76.22} \\

\bottomrule
\end{tabular}
\end{table*}

\begin{figure*}[!htbp]
\centering
\setlength\tabcolsep{2pt}
\begin{tabular}{cccccccc}
    Reference & NeuS & HF-NeuS & Neuralangelo & PGSR & GaussianUDF & GGGS & Ours \\
    \includegraphics[width=.1200\linewidth, page=1]{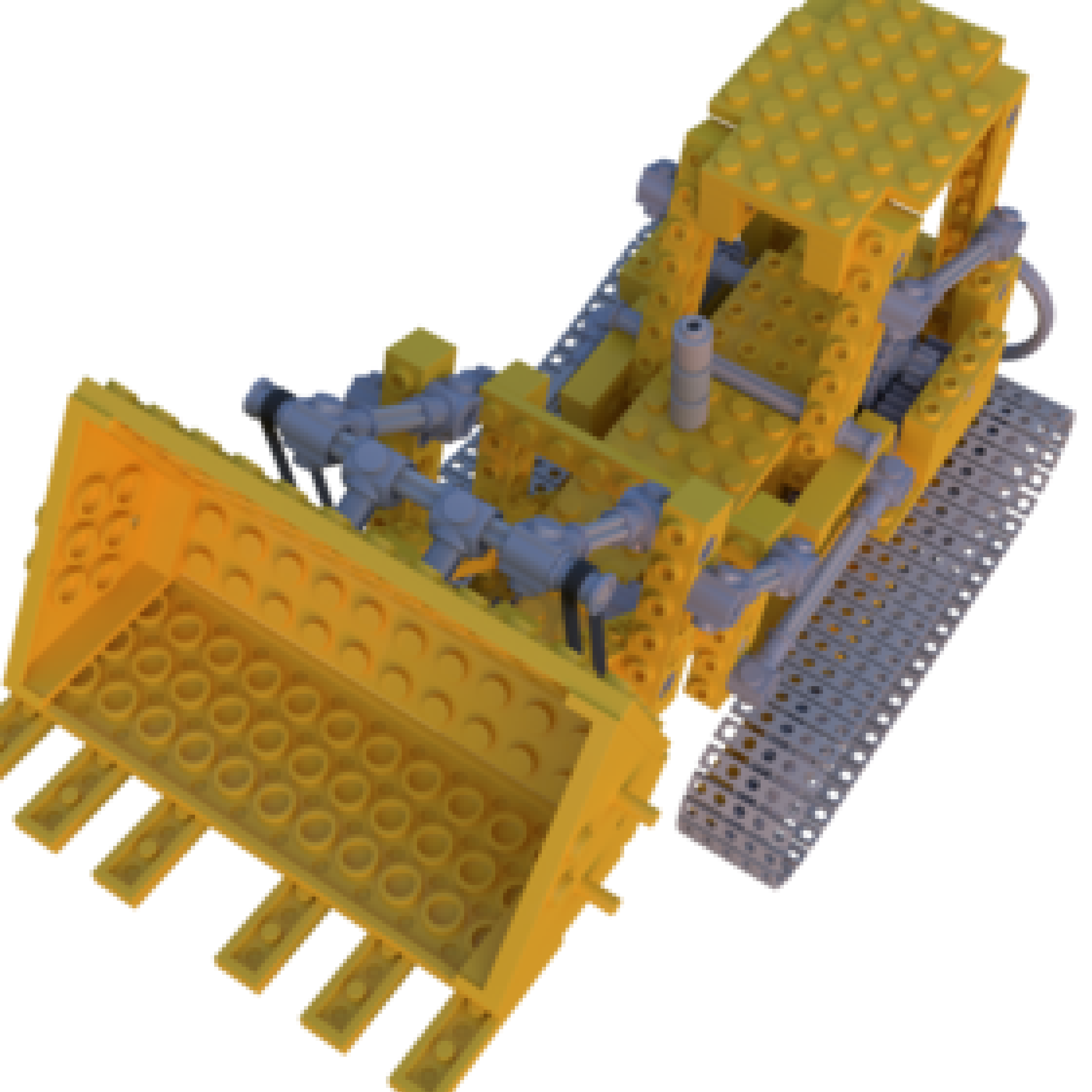} &
    \includegraphics[width=.1200\linewidth, page=2]{figs/exp/custom/lego/118/neus} &
    \includegraphics[width=.1200\linewidth, page=2]{figs/exp/custom/lego/118/hfs} &
    \includegraphics[width=.1200\linewidth, page=2]{figs/exp/custom/lego/118/neuralangelo} &
    \includegraphics[width=.1200\linewidth, page=2]{figs/exp/custom/lego/118/pgsr_post} &
    \includegraphics[width=.1200\linewidth, page=2]{figs/exp/custom/lego/118/gaussianudf} &
    \includegraphics[width=.1200\linewidth, page=2]{figs/exp/custom/lego/118/gggs_post} &
    \includegraphics[width=.1200\linewidth, page=2]{figs/exp/custom/lego/118/mpfs1} \\
    \includegraphics[width=.1200\linewidth, page=2]{figs/exp/custom/lego/118/reference} &
    \includegraphics[width=.1200\linewidth, page=3]{figs/exp/custom/lego/118/neus} &
    \includegraphics[width=.1200\linewidth, page=3]{figs/exp/custom/lego/118/hfs} &
    \includegraphics[width=.1200\linewidth, page=3]{figs/exp/custom/lego/118/neuralangelo} &
    \includegraphics[width=.1200\linewidth, page=3]{figs/exp/custom/lego/118/pgsr_post} &
    \includegraphics[width=.1200\linewidth, page=3]{figs/exp/custom/lego/118/gaussianudf} &
    \includegraphics[width=.1200\linewidth, page=3]{figs/exp/custom/lego/118/gggs_post} &
    \includegraphics[width=.1200\linewidth, page=3]{figs/exp/custom/lego/118/mpfs1} \\
    \includegraphics[width=.0800\linewidth, page=1]{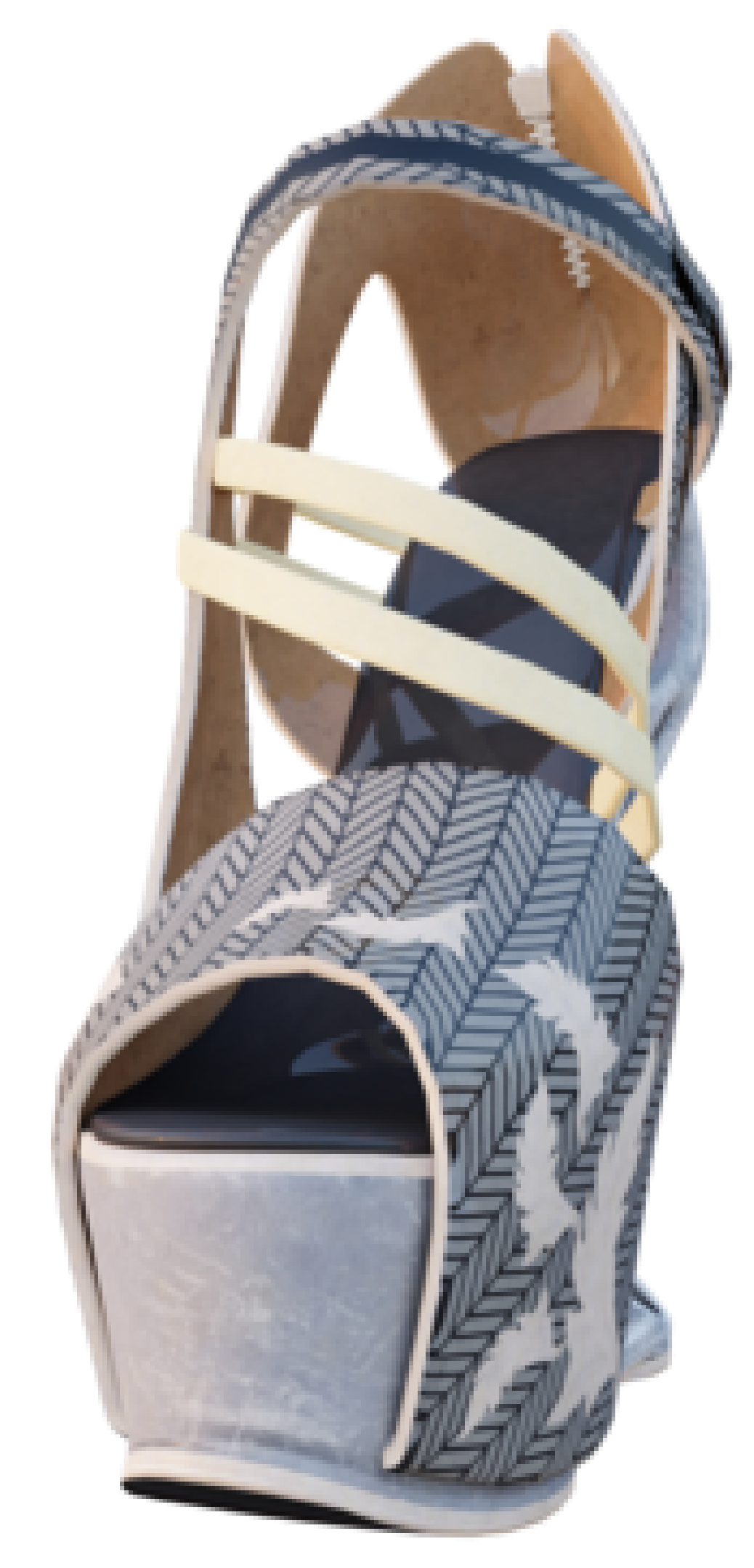} &
    \includegraphics[width=.0800\linewidth, page=2]{figs/exp/custom/shoe/076/neus} &
    \includegraphics[width=.0800\linewidth, page=2]{figs/exp/custom/shoe/076/hfs} &
    \includegraphics[width=.0800\linewidth, page=2]{figs/exp/custom/shoe/076/neuralangelo} &
    \includegraphics[width=.0800\linewidth, page=2]{figs/exp/custom/shoe/076/pgsr_post} &
    \includegraphics[width=.0800\linewidth, page=2]{figs/exp/custom/shoe/076/gaussianudf} &
    \includegraphics[width=.0800\linewidth, page=2]{figs/exp/custom/shoe/076/gggs_post} &
    \includegraphics[width=.0800\linewidth, page=2]{figs/exp/custom/shoe/076/mpfs1} \\
    \includegraphics[width=.1200\linewidth, page=2]{figs/exp/custom/shoe/076/reference} &
    \includegraphics[width=.1200\linewidth, page=3]{figs/exp/custom/shoe/076/neus} &
    \includegraphics[width=.1200\linewidth, page=3]{figs/exp/custom/shoe/076/hfs} &
    \includegraphics[width=.1200\linewidth, page=3]{figs/exp/custom/shoe/076/neuralangelo} &
    \includegraphics[width=.1200\linewidth, page=3]{figs/exp/custom/shoe/076/pgsr_post} &
    \includegraphics[width=.1200\linewidth, page=3]{figs/exp/custom/shoe/076/gaussianudf} &
    \includegraphics[width=.1200\linewidth, page=3]{figs/exp/custom/shoe/076/gggs_post} &
    \includegraphics[width=.1200\linewidth, page=3]{figs/exp/custom/shoe/076/mpfs1} \\
    \includegraphics[width=.1200\linewidth, page=1]{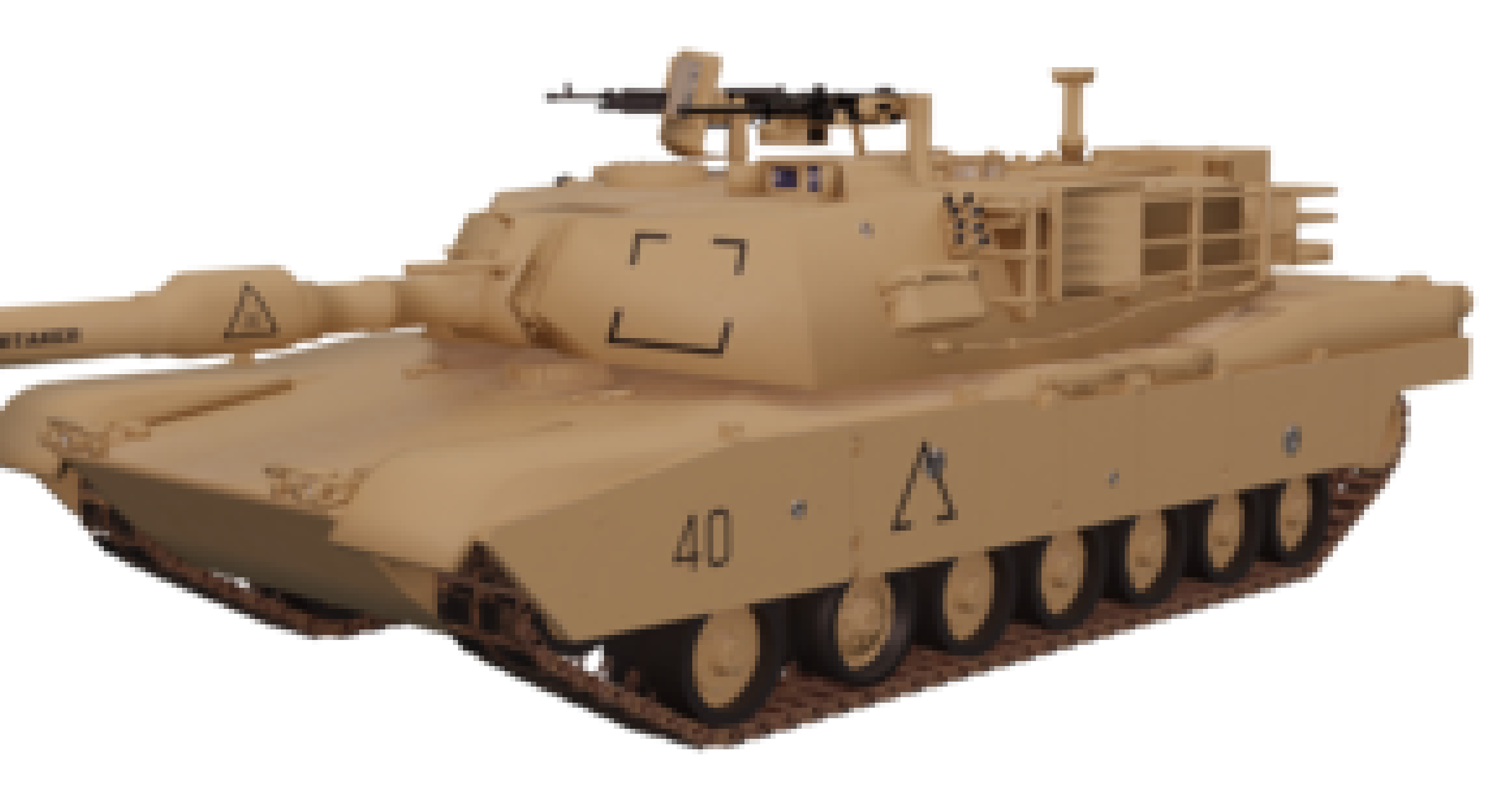} &
    \includegraphics[width=.1200\linewidth, page=2]{figs/exp/custom/tank/086/neus} &
    \includegraphics[width=.1200\linewidth, page=2]{figs/exp/custom/tank/086/hfs} &
    \includegraphics[width=.1200\linewidth, page=2]{figs/exp/custom/tank/086/neuralangelo} &
    \includegraphics[width=.1200\linewidth, page=2]{figs/exp/custom/tank/086/pgsr_post} &
    \includegraphics[width=.1200\linewidth, page=2]{figs/exp/custom/tank/086/gaussianudf} &
    \includegraphics[width=.1200\linewidth, page=2]{figs/exp/custom/tank/086/gggs_post} &
    \includegraphics[width=.1200\linewidth, page=2]{figs/exp/custom/tank/086/mpfs1} \\
    \includegraphics[width=.1200\linewidth, page=2]{figs/exp/custom/tank/086/reference} &
    \includegraphics[width=.1200\linewidth, page=3]{figs/exp/custom/tank/086/neus} &
    \includegraphics[width=.1200\linewidth, page=3]{figs/exp/custom/tank/086/hfs} &
    \includegraphics[width=.1200\linewidth, page=3]{figs/exp/custom/tank/086/neuralangelo} &
    \includegraphics[width=.1200\linewidth, page=3]{figs/exp/custom/tank/086/pgsr_post} &
    \includegraphics[width=.1200\linewidth, page=3]{figs/exp/custom/tank/086/gaussianudf} &
    \includegraphics[width=.1200\linewidth, page=3]{figs/exp/custom/tank/086/gggs_post} &
    \includegraphics[width=.1200\linewidth, page=3]{figs/exp/custom/tank/086/mpfs1} \\
    \includegraphics[width=.1000\linewidth, page=1]{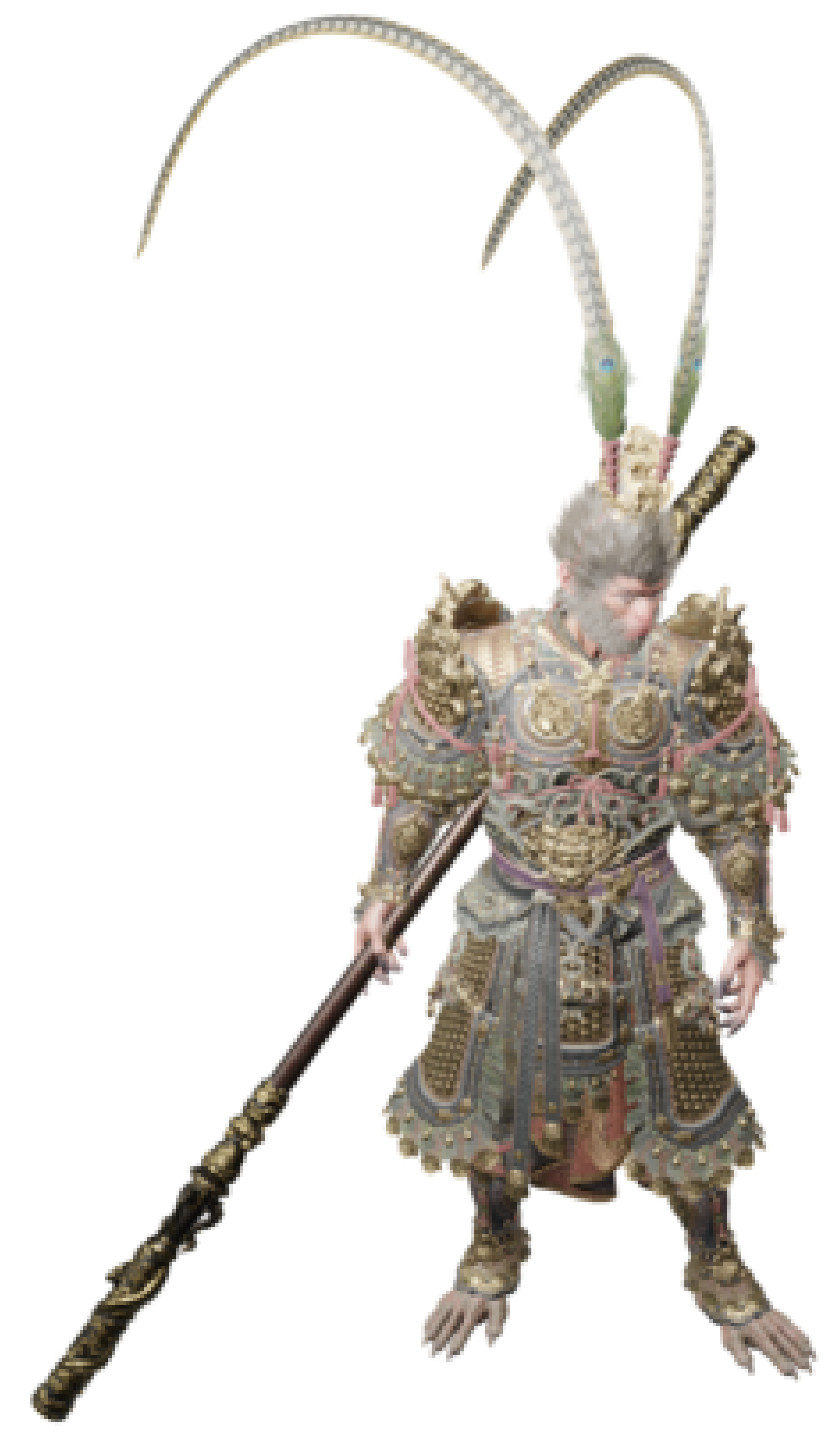} &
    \includegraphics[width=.1000\linewidth, page=2]{figs/exp/custom/figurine/088/neus} &
    \includegraphics[width=.1000\linewidth, page=2]{figs/exp/custom/figurine/088/hfs} &
    \includegraphics[width=.1000\linewidth, page=2]{figs/exp/custom/figurine/088/neuralangelo} &
    \includegraphics[width=.1000\linewidth, page=2]{figs/exp/custom/figurine/088/pgsr_post} &
    \includegraphics[width=.1000\linewidth, page=2]{figs/exp/custom/figurine/088/gaussianudf} &
    \includegraphics[width=.1000\linewidth, page=2]{figs/exp/custom/figurine/088/gggs_post} &
    \includegraphics[width=.1000\linewidth, page=2]{figs/exp/custom/figurine/088/mpfs1} \\
    \includegraphics[width=.1000\linewidth, page=2]{figs/exp/custom/figurine/088/reference} &
    \includegraphics[width=.0960\linewidth, page=3]{figs/exp/custom/figurine/088/neus} &
    \includegraphics[width=.0960\linewidth, page=3]{figs/exp/custom/figurine/088/hfs} &
    \includegraphics[width=.0960\linewidth, page=3]{figs/exp/custom/figurine/088/neuralangelo} &
    \includegraphics[width=.0960\linewidth, page=3]{figs/exp/custom/figurine/088/pgsr_post} &
    \includegraphics[width=.0960\linewidth, page=3]{figs/exp/custom/figurine/088/gaussianudf} &
    \includegraphics[width=.0960\linewidth, page=3]{figs/exp/custom/figurine/088/gggs_post} &
    \includegraphics[width=.0960\linewidth, page=3]{figs/exp/custom/figurine/088/mpfs1} \\
    Reference & NeuS & HF-NeuS & Neuralangelo & PGSR & GaussianUDF & GGGS & Ours \\
\end{tabular}
\caption{
Results on the TAAS-Synthetic dataset. Red boxes highlight challenging thin or sign-ambiguous structures. None of the baseline methods (NeuS, PGSR, GaussianUDF, and GGGS) is able to recover the complete geometry: they often produce missing parts, fragmented structures, or floating artifacts.
In contrast, our method reconstructs the full geometry more faithfully and preserves fine structures. GeoSVR fails to produce any mesh for these cases and is therefore excluded from the comparison.
}
\label{fig:results custom syn}
\end{figure*}

\paragraph{Additional results}
In this section, we conduct experiments on UDF reconstruction methods, present representative qualitative results on the Deep Fashion3D dataset~\cite{Zhu2020DF3D} commonly used in UDF reconstruction evaluation, as well as the Tanks \& Temples dataset~\cite{Knapitsch2017TnT} that focuses on scene-level reconstruction.

We compare against NeUDF~\cite{Liu2023NeUDF}, a widely-accepted method of UDF reconstruction, to our dataset. Theoretically, UDFs are capable of reconstructing geometries of arbitrary topology, including the thin structures and sign-ambiguous structures present in our data. However, in practice, optimizing UDFs with neural networks is very challenging.

\begin{figure}[!htb]
\centering
\begin{minipage}{\linewidth}
\begin{minipage}[t]{.2071\linewidth}
    \centering
    \begin{minipage}[t][6pt]{\linewidth}
    \centering
        Umbrella
    \end{minipage}\\[5pt]
    \includegraphics[width=\linewidth, page=2]{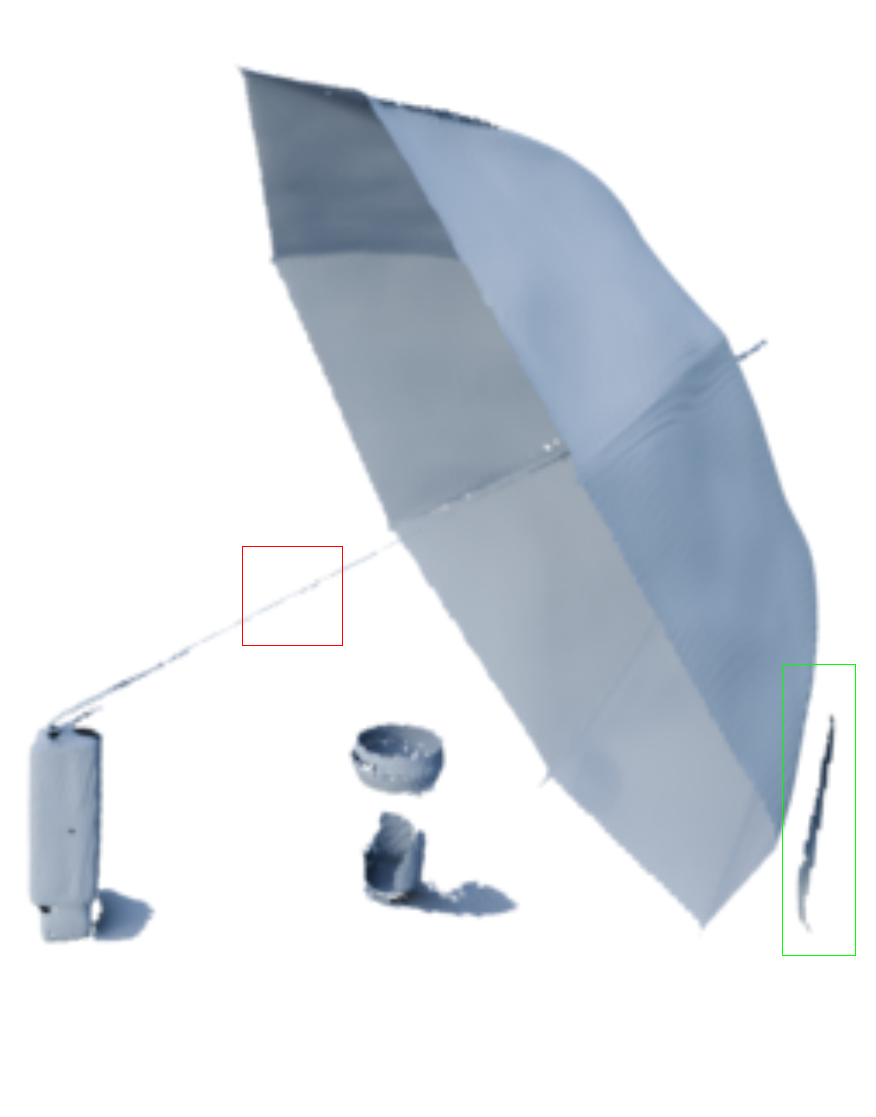} \\[5pt]
    \includegraphics[width=.75\linewidth, page=3]{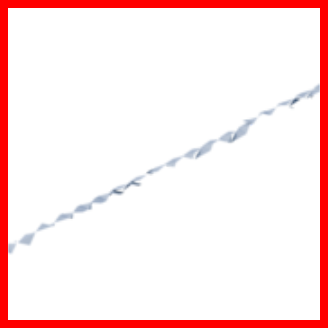}\hfill\includegraphics[width=.20\linewidth, page=4]{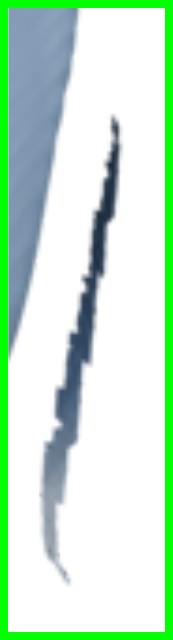}
\end{minipage}\hfill%
\begin{minipage}[t]{.2860\linewidth}
    \centering
    \begin{minipage}[t][6pt]{\linewidth}
    \centering
        Flower
    \end{minipage}\\[5pt]
    \includegraphics[width=\linewidth, page=2]{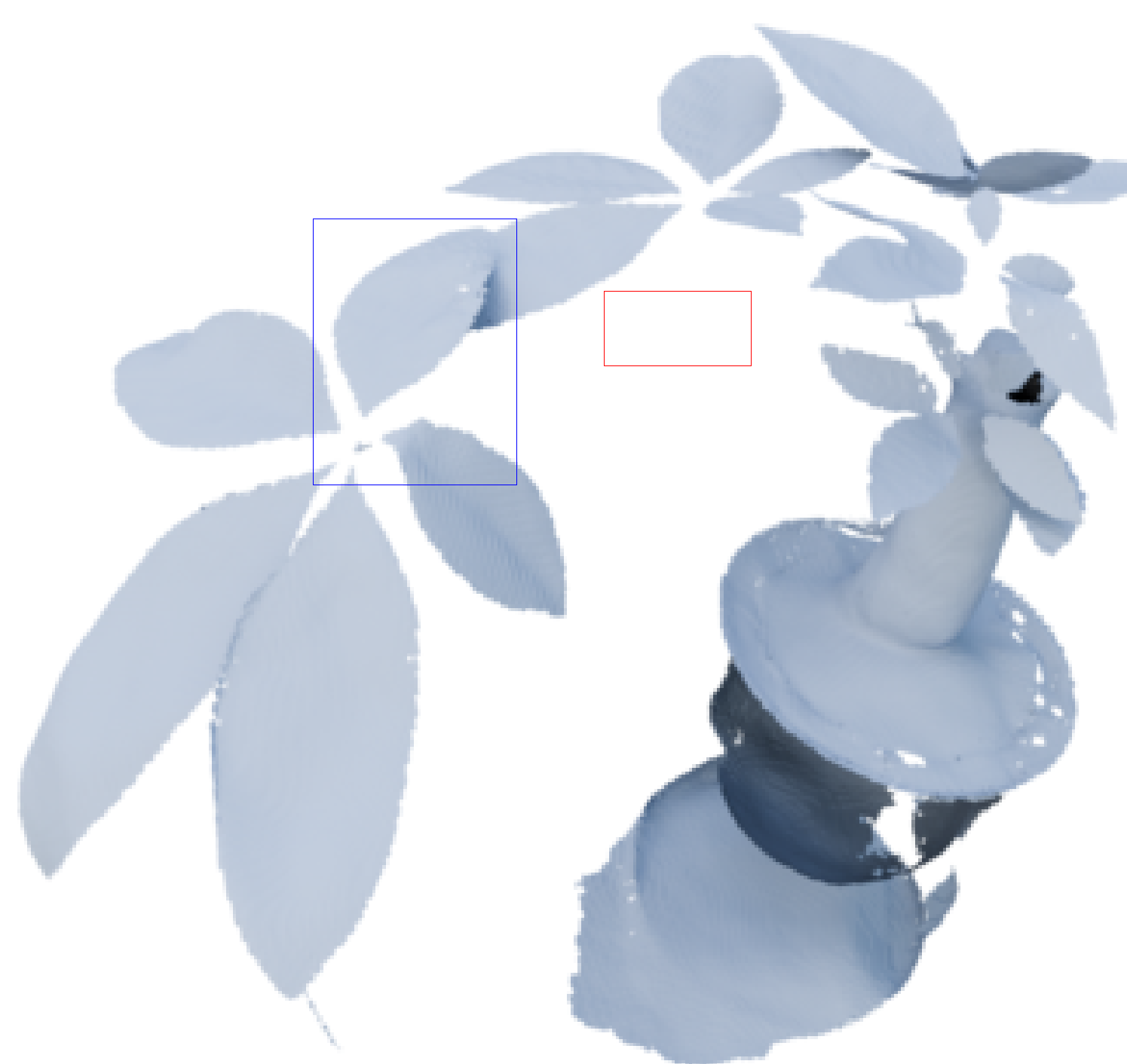} \\[5pt]
    \includegraphics[width=\linewidth, page=3]{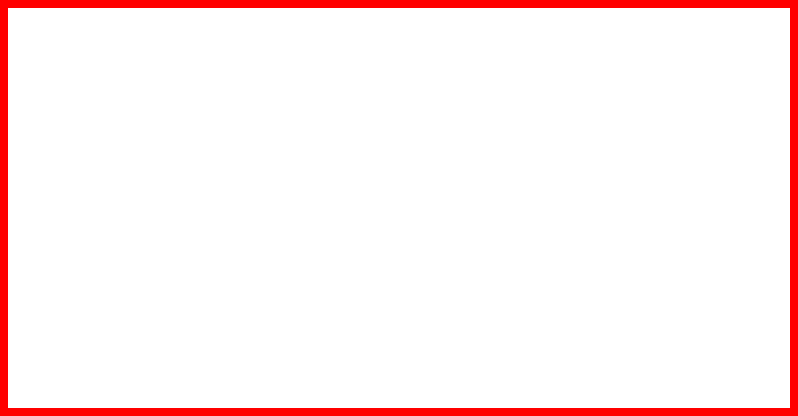}
\end{minipage}\hfill
\begin{minipage}[t]{.2372\linewidth}
    \centering
    \begin{minipage}[t][6pt]{\linewidth}
    \centering
        Basket
    \end{minipage}\\[5pt]
    \includegraphics[width=\linewidth, page=2]{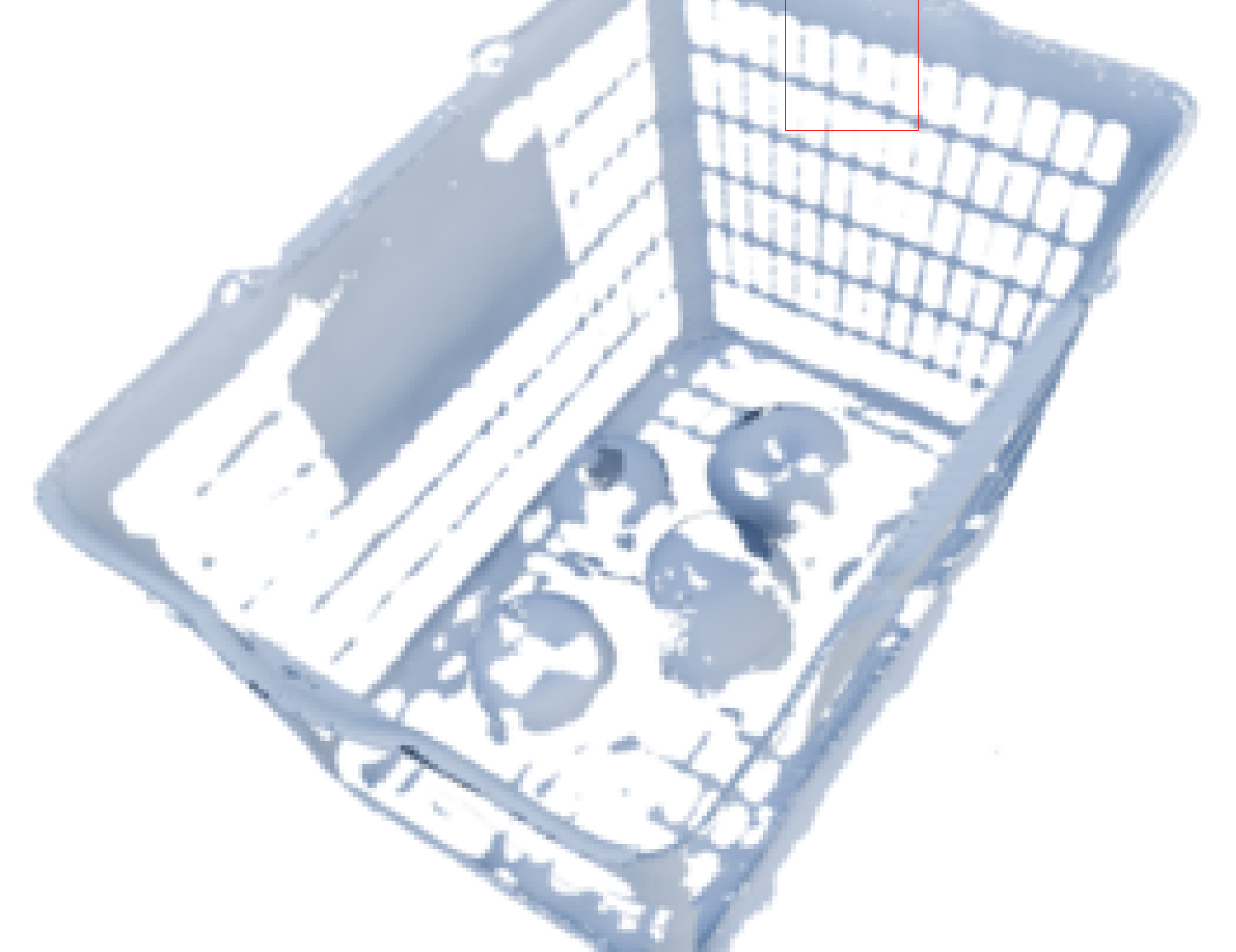} \\[5pt]
    \includegraphics[width=\linewidth, page=3]{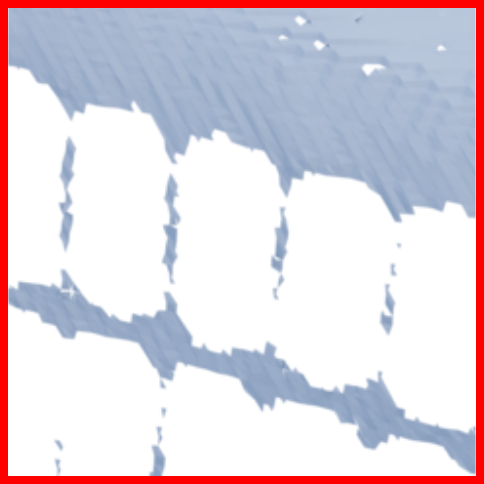}
\end{minipage}\hfill
\begin{minipage}[t]{.2497\linewidth}
    \centering
    \begin{minipage}[t][6pt]{\linewidth}
    \centering
        Cup
    \end{minipage}\\[5pt]
    \includegraphics[width=\linewidth, page=2]{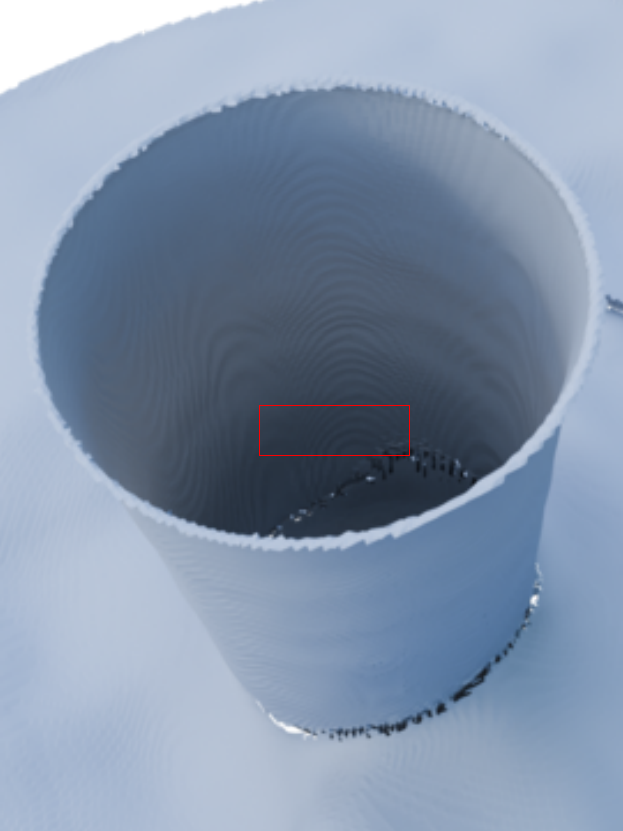} \\[5pt]
    \includegraphics[width=\linewidth, page=3]{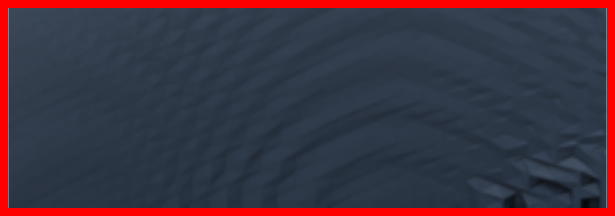}
\end{minipage}
\end{minipage}\vspace{5pt}
\rule{\linewidth}{1pt}\\
\vspace{5pt}
\begin{minipage}{\linewidth}
\begin{minipage}[t]{.2426\linewidth}
    \centering
    \begin{minipage}[t][6pt]{\linewidth}
    \centering
        LEGO
    \end{minipage}\\[5pt]
    \includegraphics[width=\linewidth, page=2]{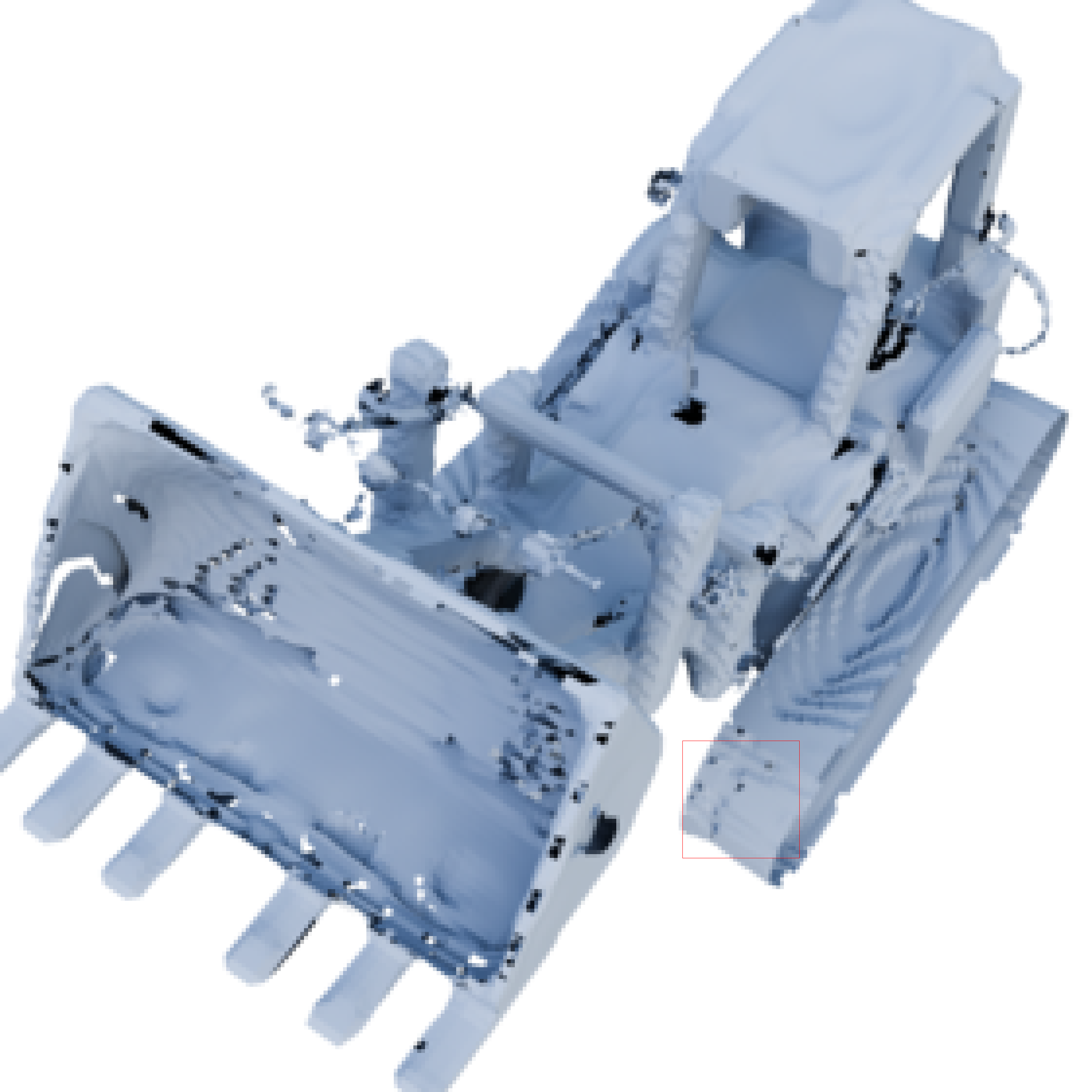} \\[5pt]
    \includegraphics[width=\linewidth, page=3]{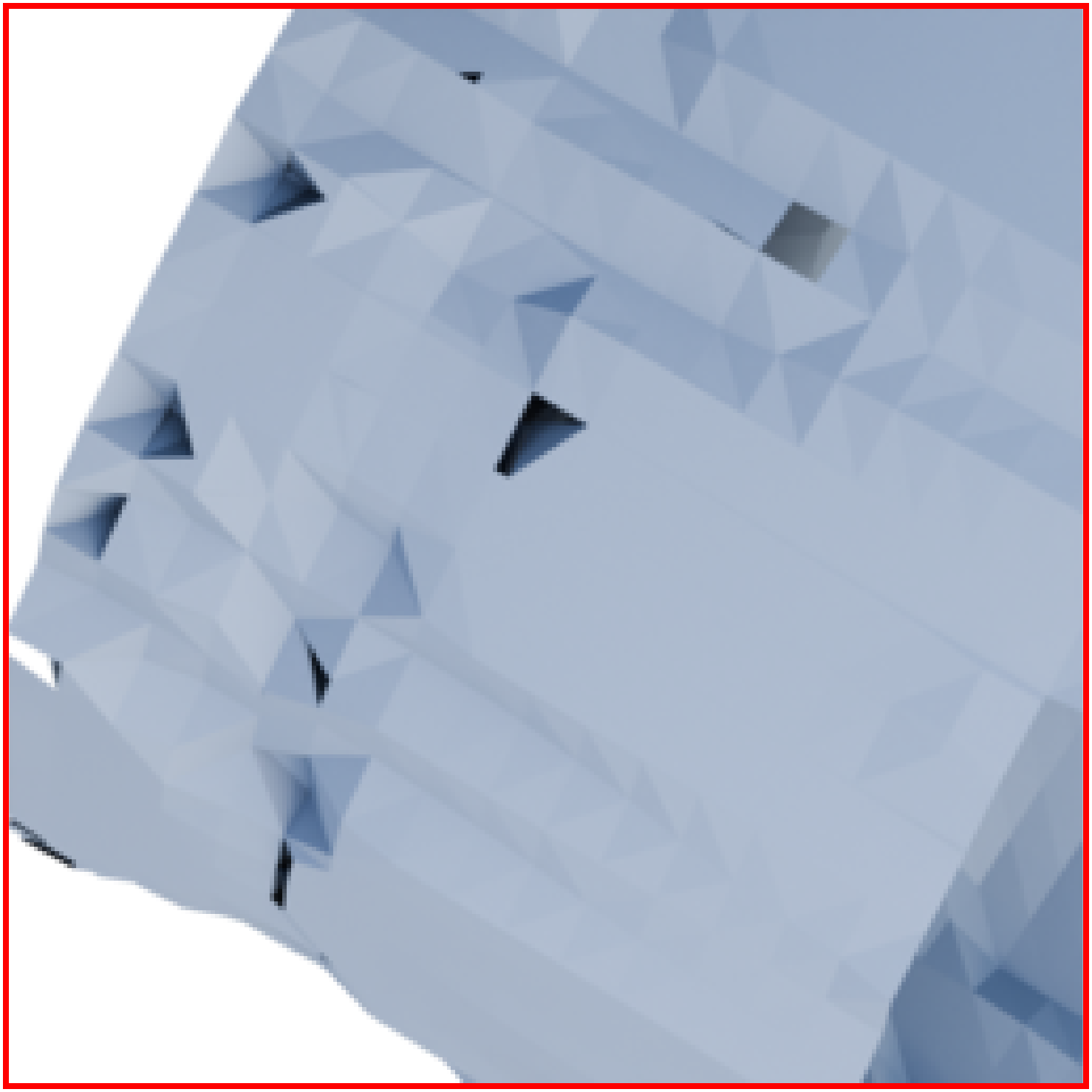}
\end{minipage}\hfill%
\begin{minipage}[t]{.2542\linewidth}
    \centering
    \begin{minipage}[t][6pt]{\linewidth}
    \centering
        Shoe
    \end{minipage}\\[5pt]
    \includegraphics[width=.67\linewidth, page=2]{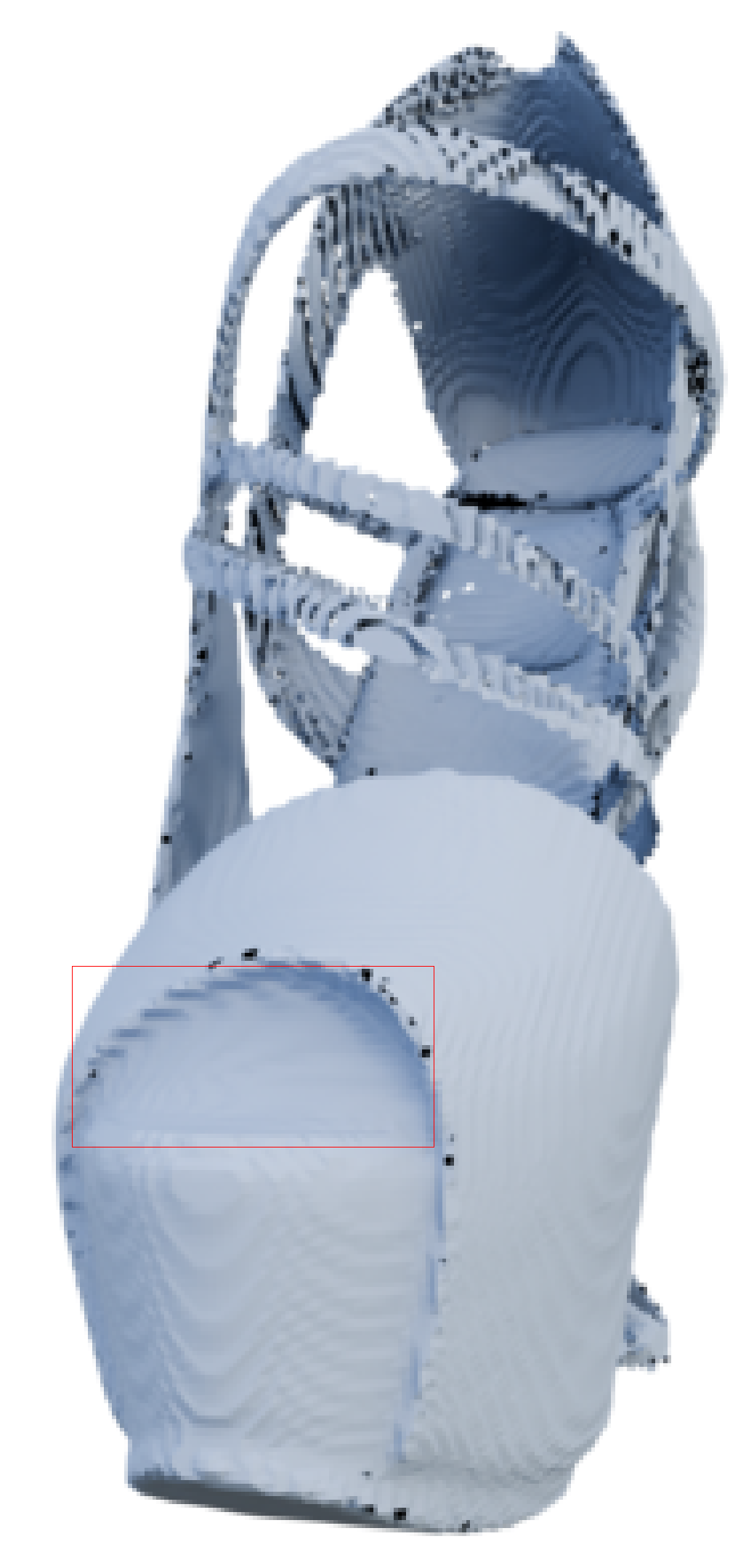} \\[5pt]
    \includegraphics[width=\linewidth, page=3]{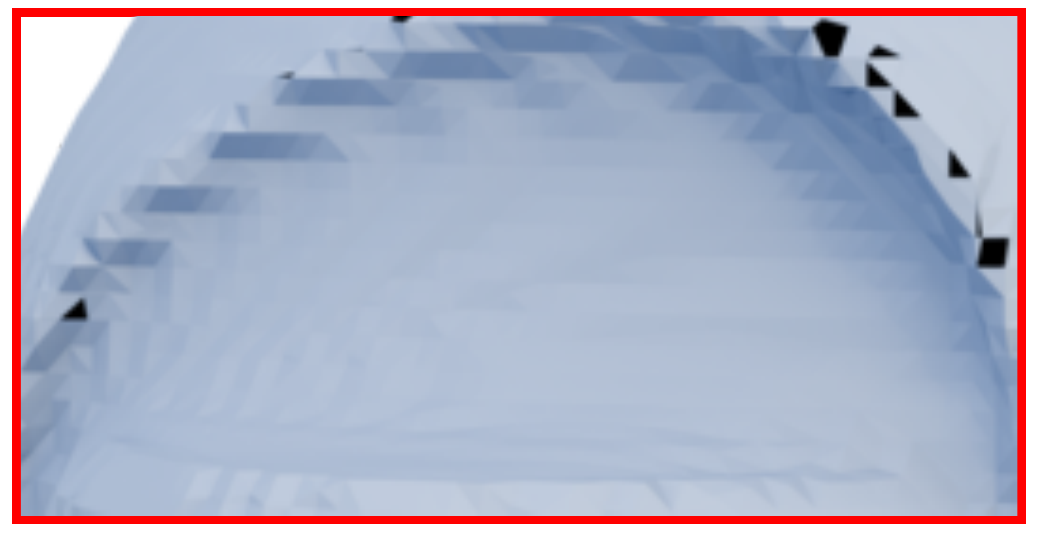}
\end{minipage}\hfill
\begin{minipage}[t]{.3153\linewidth}
    \centering
    \begin{minipage}[t][6pt]{\linewidth}
    \centering
        Tank
    \end{minipage}\\[5pt]
    \includegraphics[width=\linewidth, page=2]{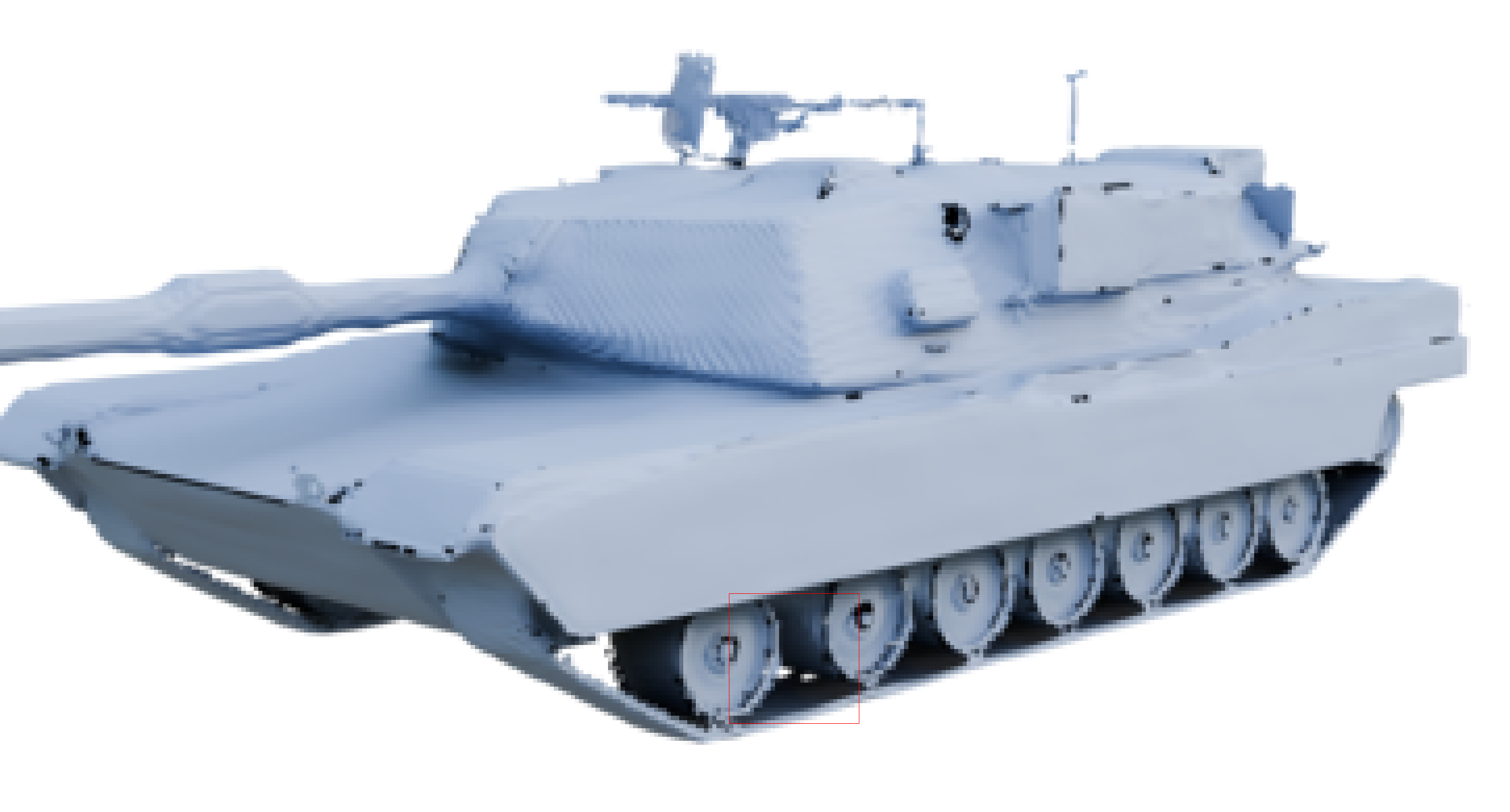} \\[5pt]
    \includegraphics[width=\linewidth, page=3]{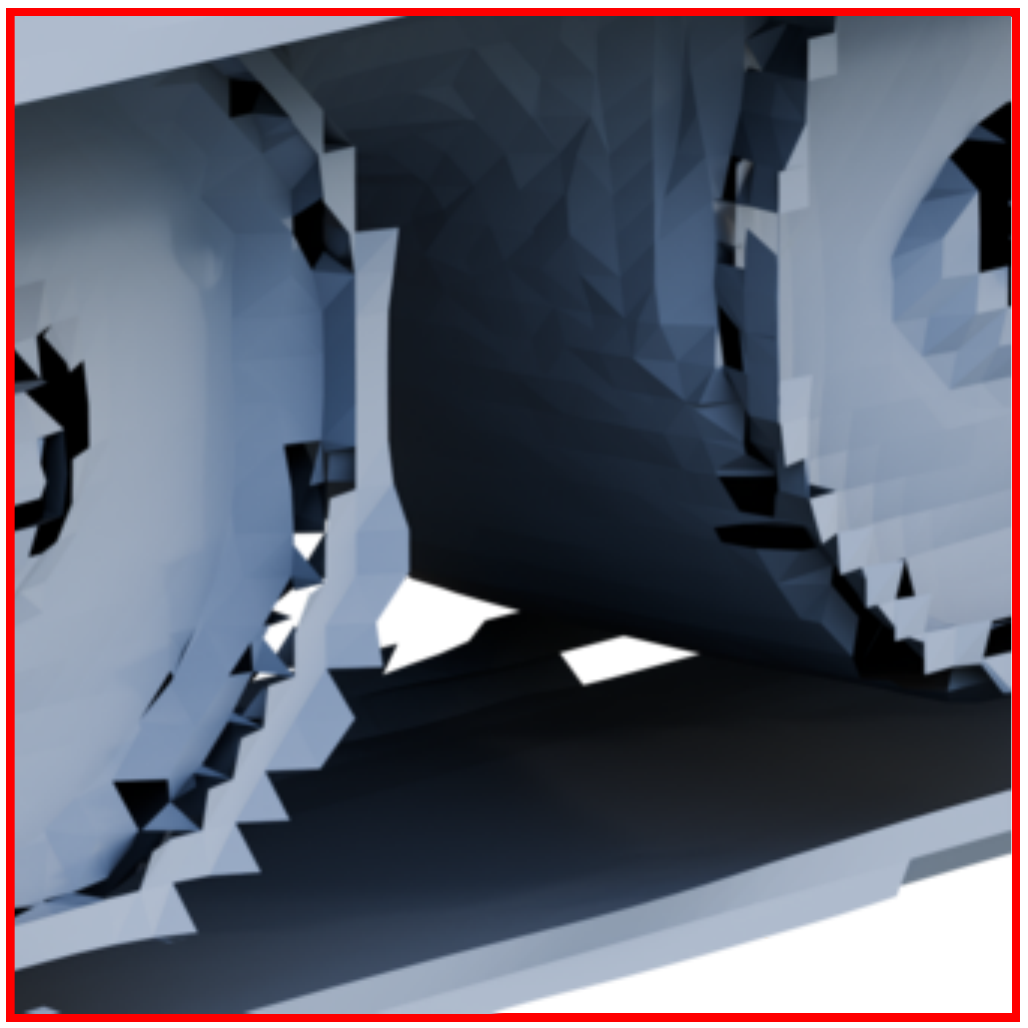}
\end{minipage}\hfill
\begin{minipage}[t]{.1679\linewidth}
    \centering
    \begin{minipage}[t][6pt]{\linewidth}
    \centering
        Figurine
    \end{minipage}\\[5pt]
    \includegraphics[width=\linewidth, page=2]{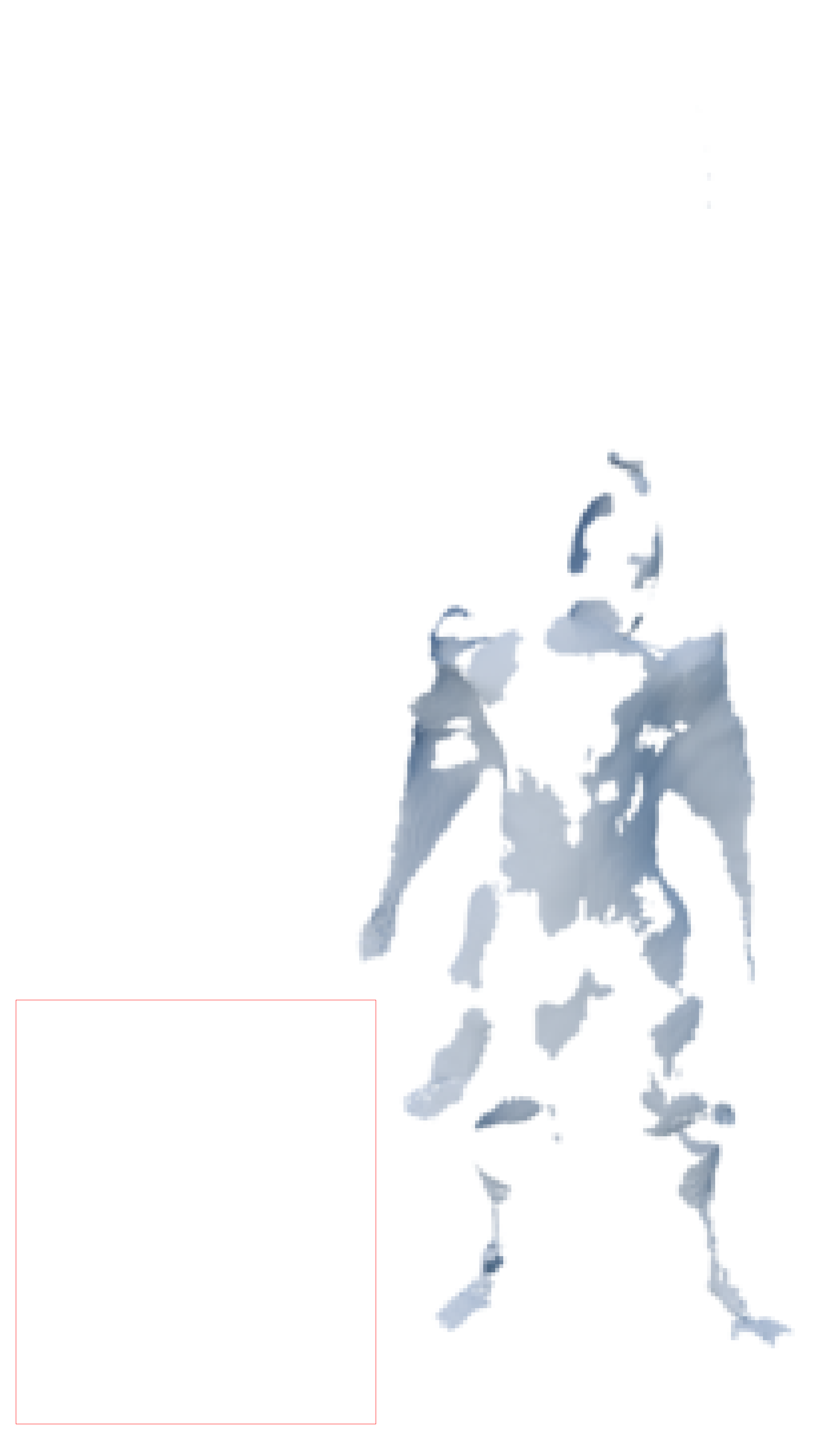} \\[5pt]
    \includegraphics[width=\linewidth, page=3]{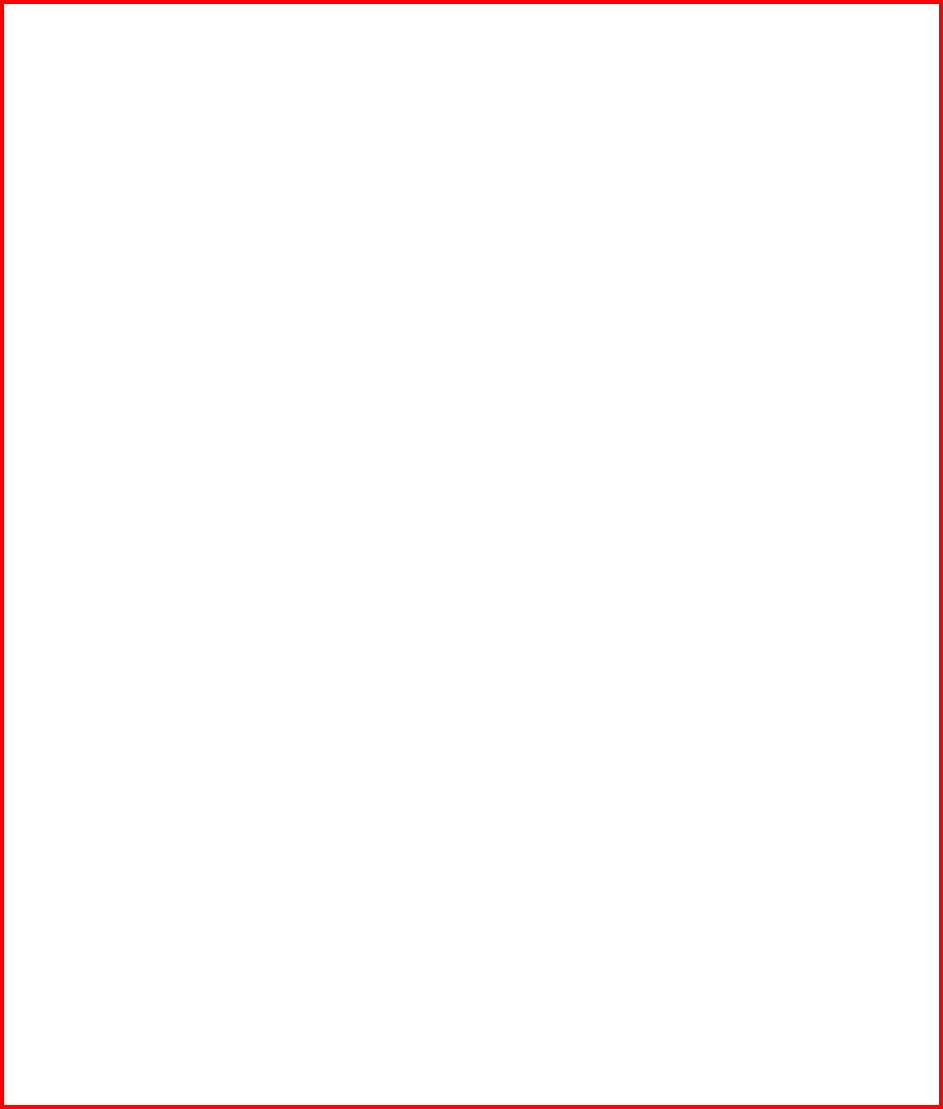}
\end{minipage}
\end{minipage}
\caption{Results of NeUDF on our dataset.}
\label{fig:neudf}
\end{figure}

We present training results of NeUDF on our dataset in Figure~\ref{fig:neudf}. In these cases, the meshes produced by NeUDF is of very poor quality, which can be attributed to both the unstable nature of UDF learning, and the lack of robust mesh extraction methods for UDFs.

\begin{figure}[!htb]
\centering
\setlength{\tabcolsep}{2pt}
\begin{tabular}{ccc}
    \includegraphics[width=.24\linewidth, page=2]{figs/exp/df3d/320/reference} & \includegraphics[width=.24\linewidth, page=2]{figs/exp/df3d/320/2sudf} & \includegraphics[width=.24\linewidth, page=2]{figs/exp/df3d/320/mpfs1} \\
    GT & 2S-UDF & Ours \\
    \includegraphics[width=.30\linewidth]{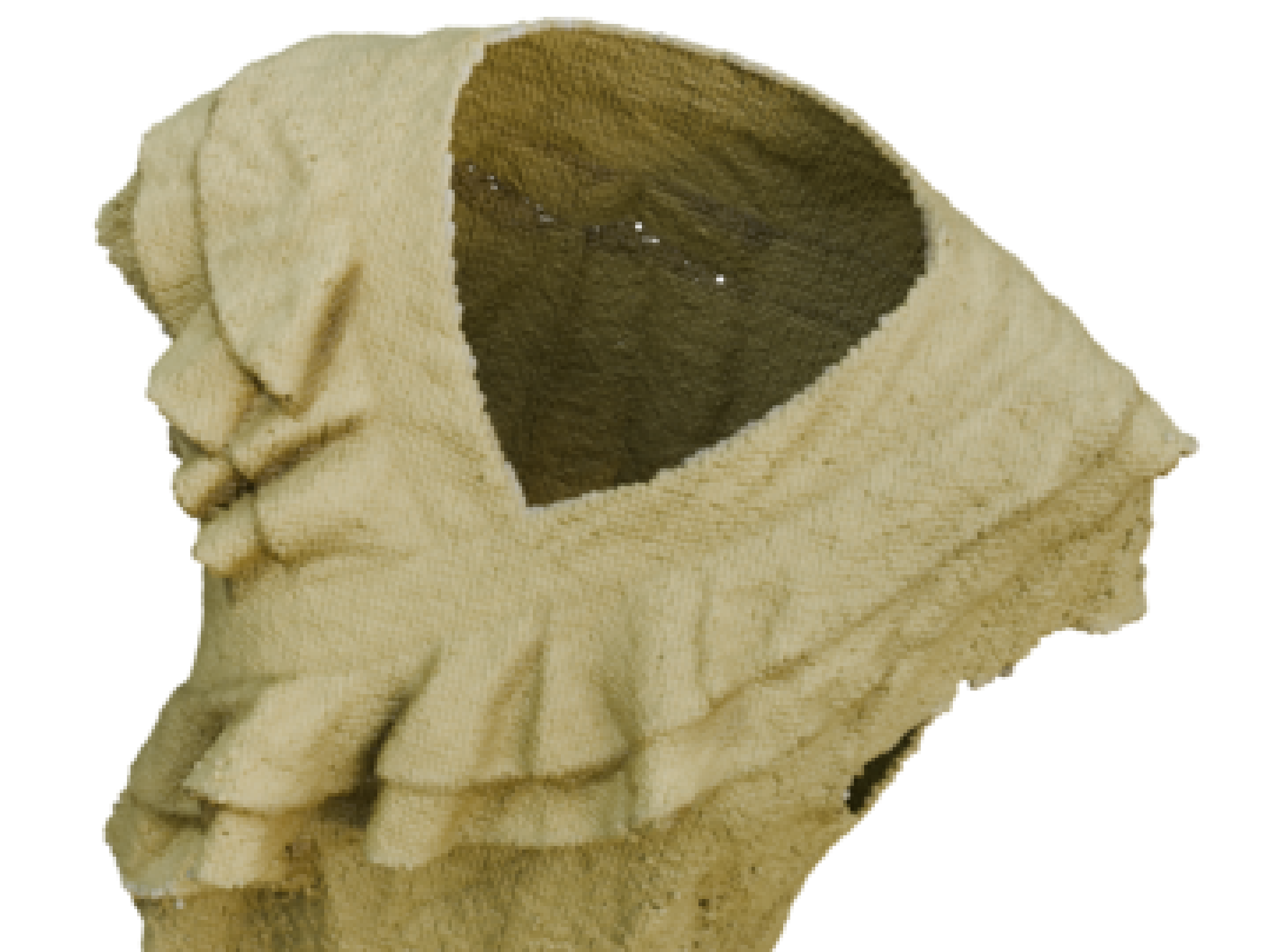} & \includegraphics[width=.30\linewidth]{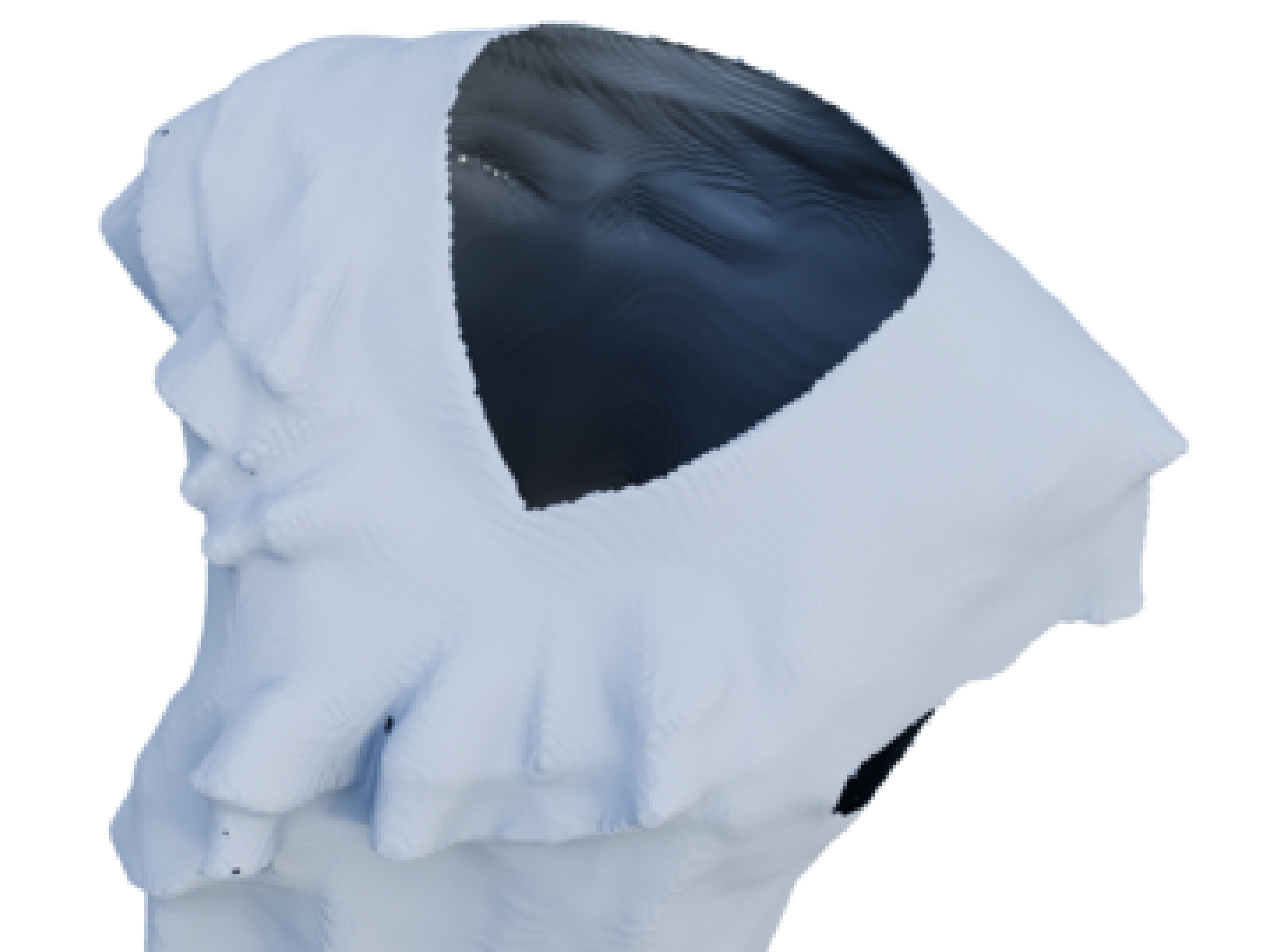} & \includegraphics[width=.30\linewidth]{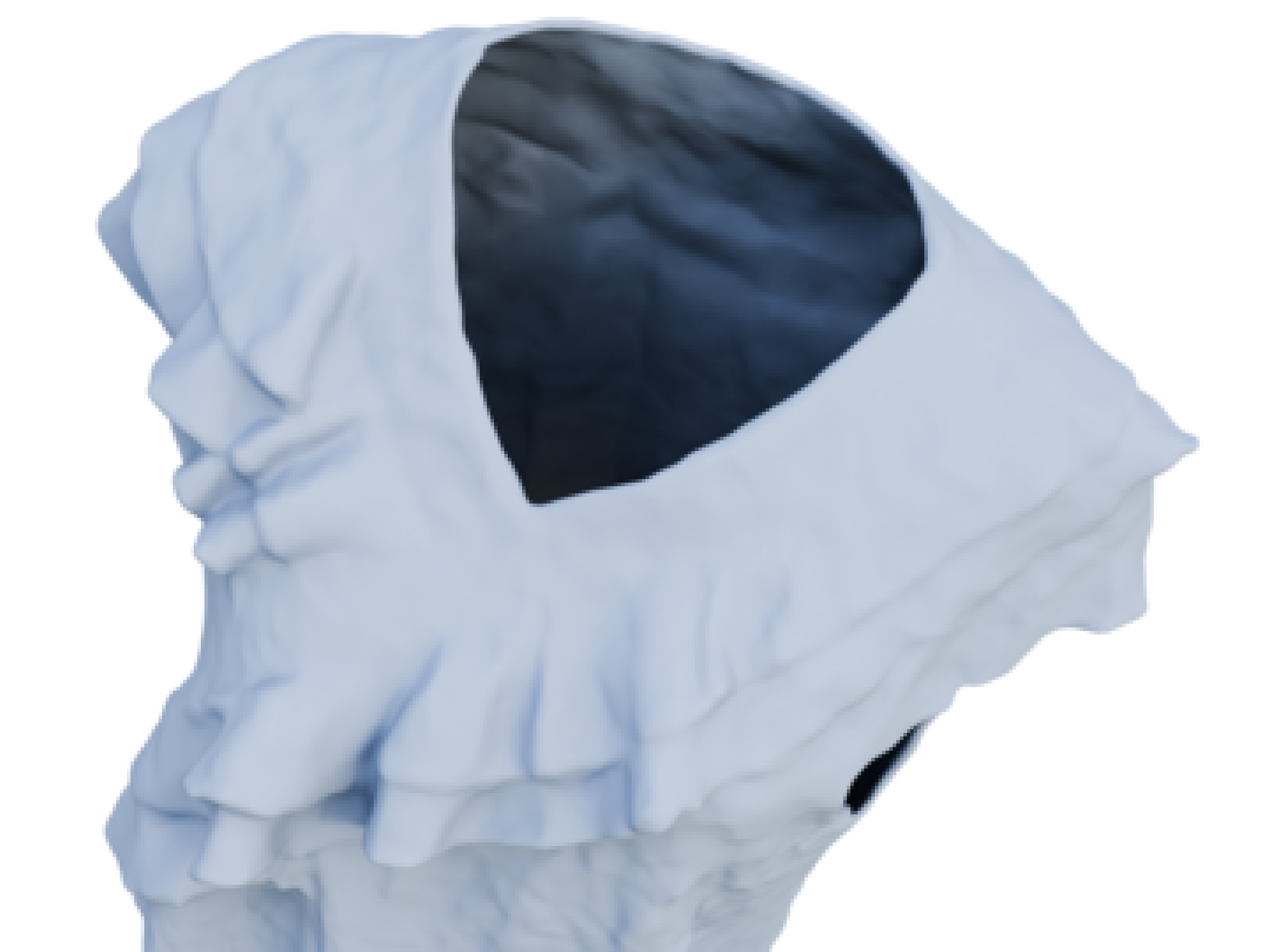}\\
    GT & NeUDF & Ours
\end{tabular}
\caption{Compared to UDF-centered reconstruction methods, our method produces two-layered watertight mesh that represents a layer of thin shell around the open surface. This sidesteps two major problems with UDF. Top: Implicit UDF methods have difficulties accurately learning the zero turning point on the surface, posing challenges to UDF mesh extraction and often produces fragmented meshes; our method constructs a more complete mesh. Bottom: Implicit UDF methods require heavy smooth terms for successful convergence, resulting in loss of details. Our method can better capture the decorative details of the surface.}
\label{fig:df3d}
\end{figure}

Deep Fashion3D~\cite{Zhu2020DF3D} is a garment dataset composed of various clothes of single layers and open surfaces, which is inherently suitable for UDF reconstruction. Since MPF is a signed field, reconstructing open surfaces—such as clothing—results in a two-layered watertight shell.
Our approach offers more stable training than UDF-based methods and benefits from more established mesh extraction techniques. Conversely, as shown in Figure~\ref{fig:df3d} at the top row, UDF methods often produce broken meshes such as the strap in the dress. In addition, UDF methods usually require heavy smoothing losses for UDF learning stability that sacrifice fine detail, as shown in the same figure in the bottom, our method preserves more decorative creases on the cloth.

We also demonstrate a test case in the Tanks \& Temples dataset~\cite{Knapitsch2017TnT}. In this case, we observe that our method and Neuralangelo can better resolve the arm in front of the chest. Other methods incorrectly connects the arm to the body. This result is consistent to our observation on the DTU dataset presented in Figure~\ref{fig:dtu}, where our method excels in better resolving ambiguous geometries.

\begin{figure}[!htb]
\centering
\setlength{\tabcolsep}{2pt}
\begin{tabular}{ccc}
\includegraphics[width=0.30\linewidth, page=2]{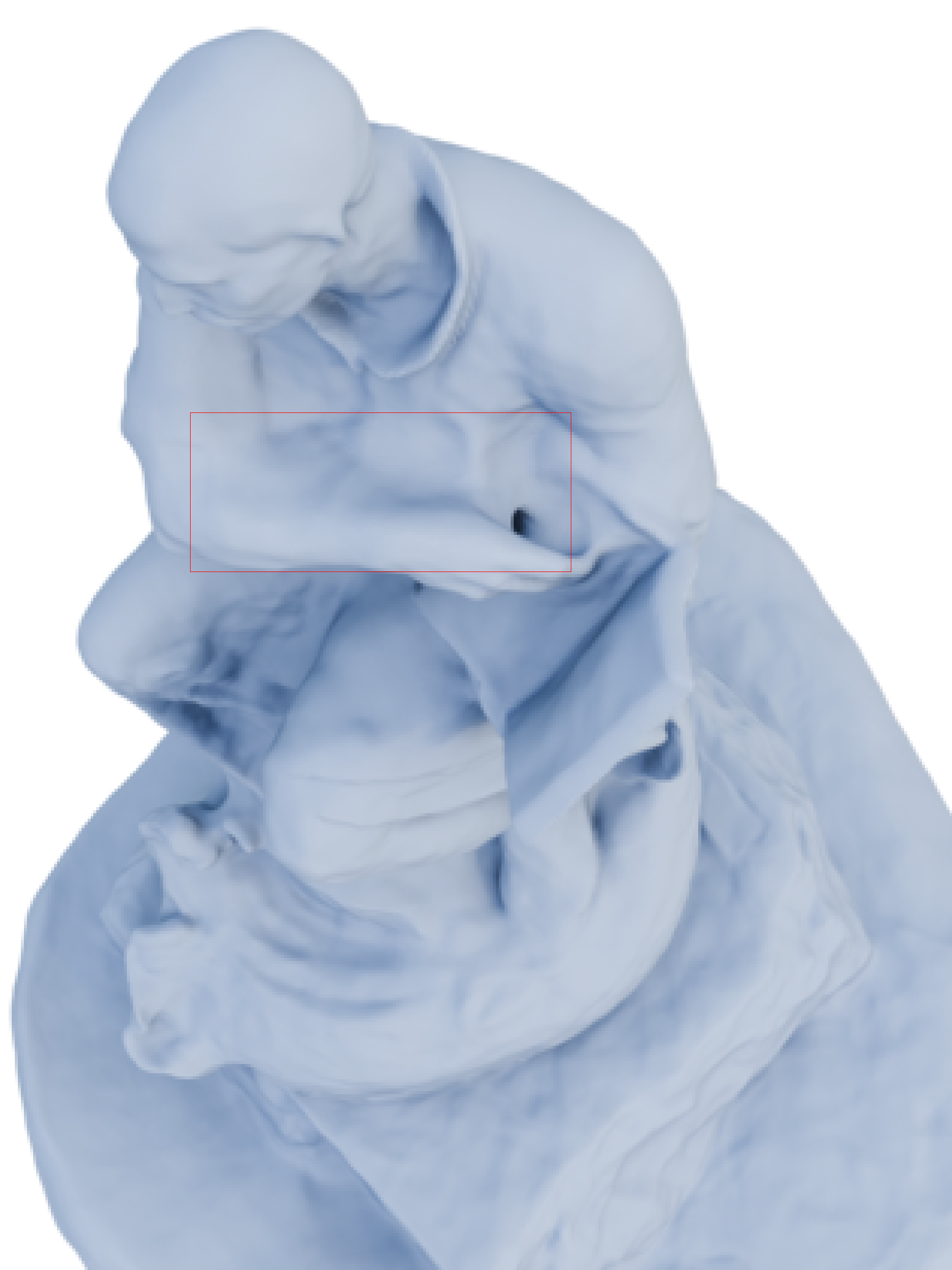} & 
\includegraphics[width=0.30\linewidth, page=2]{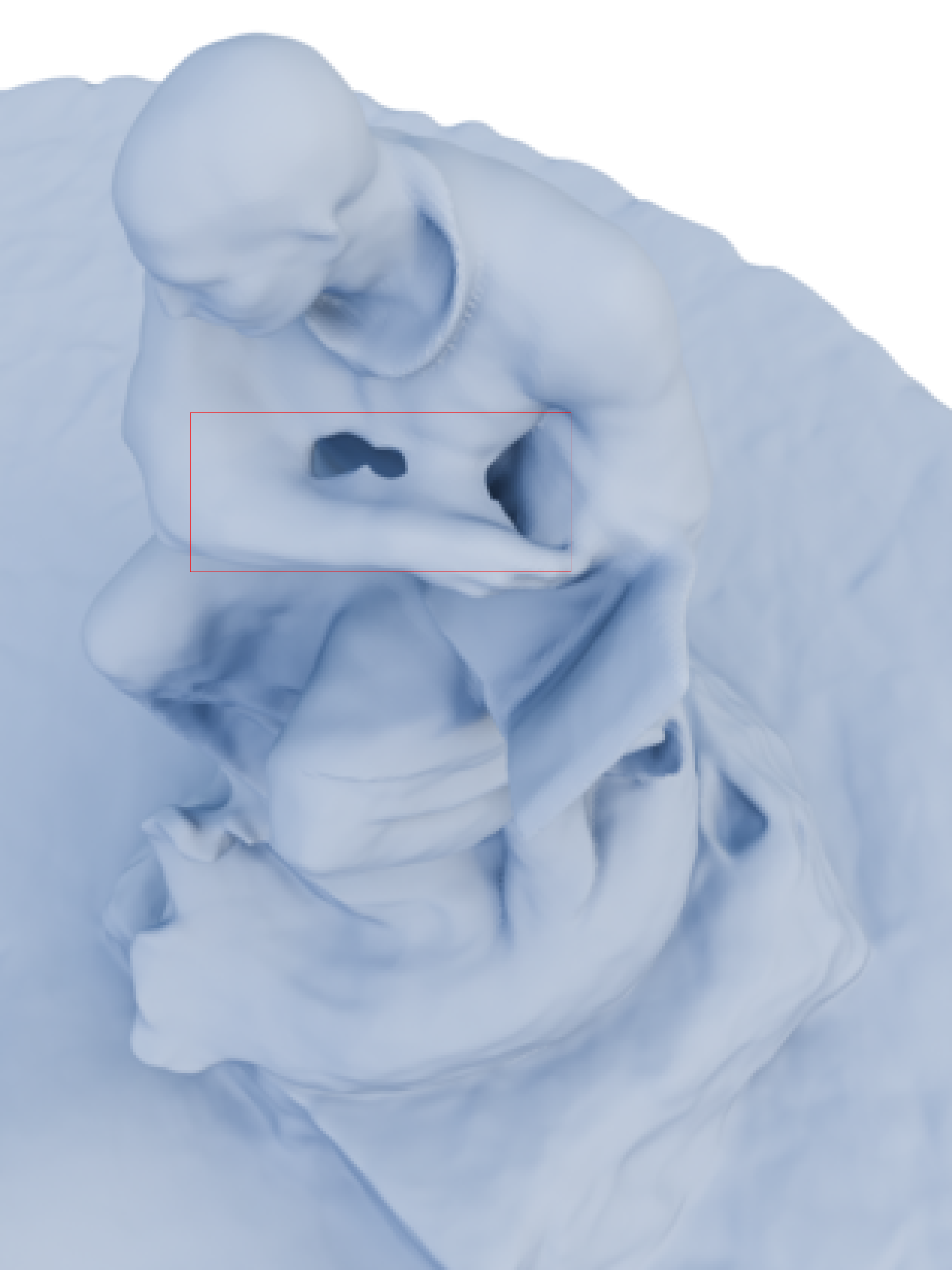} &
\includegraphics[width=0.30\linewidth, page=2]{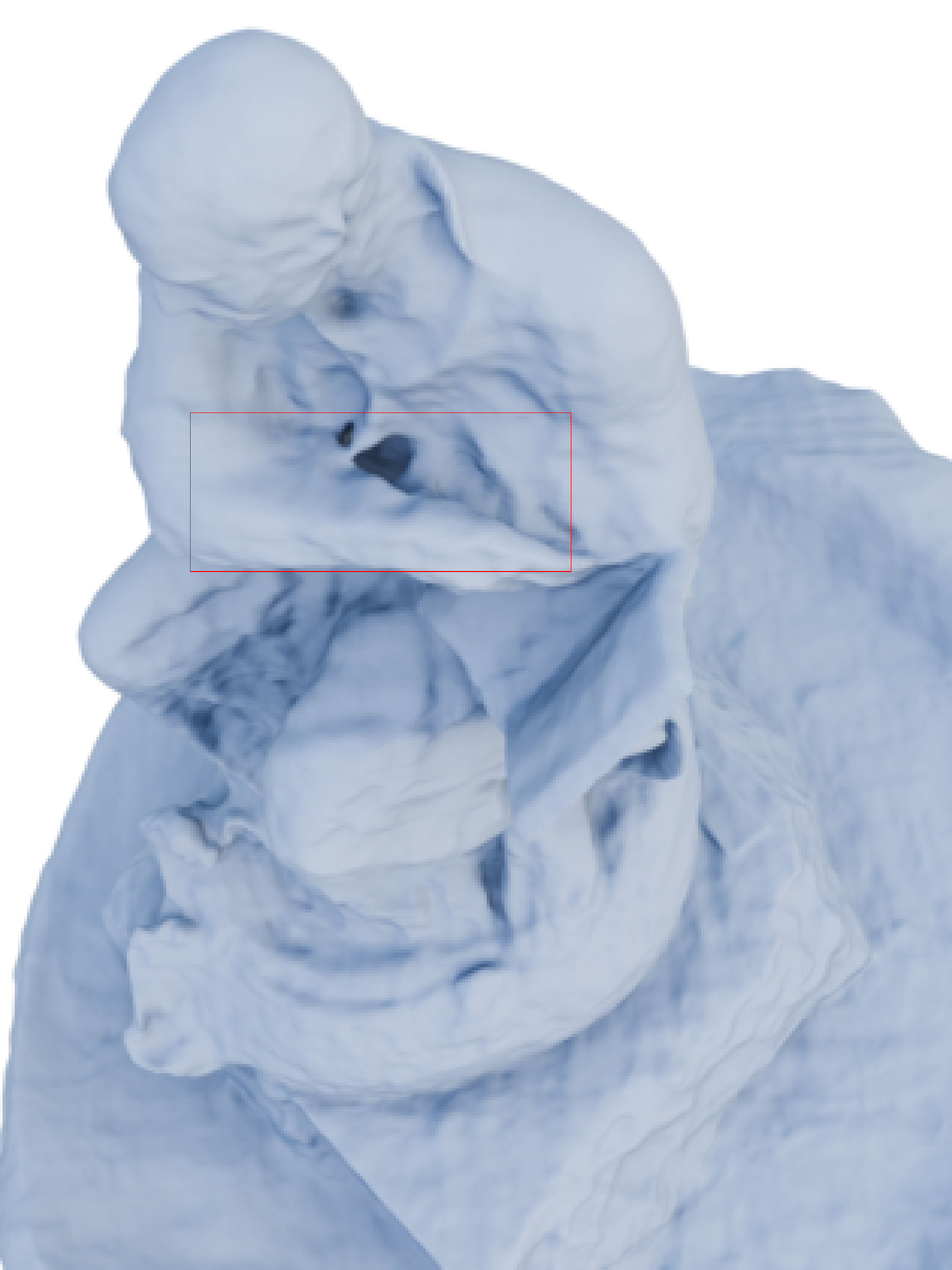} \\
\includegraphics[width=0.30\linewidth, page=3]{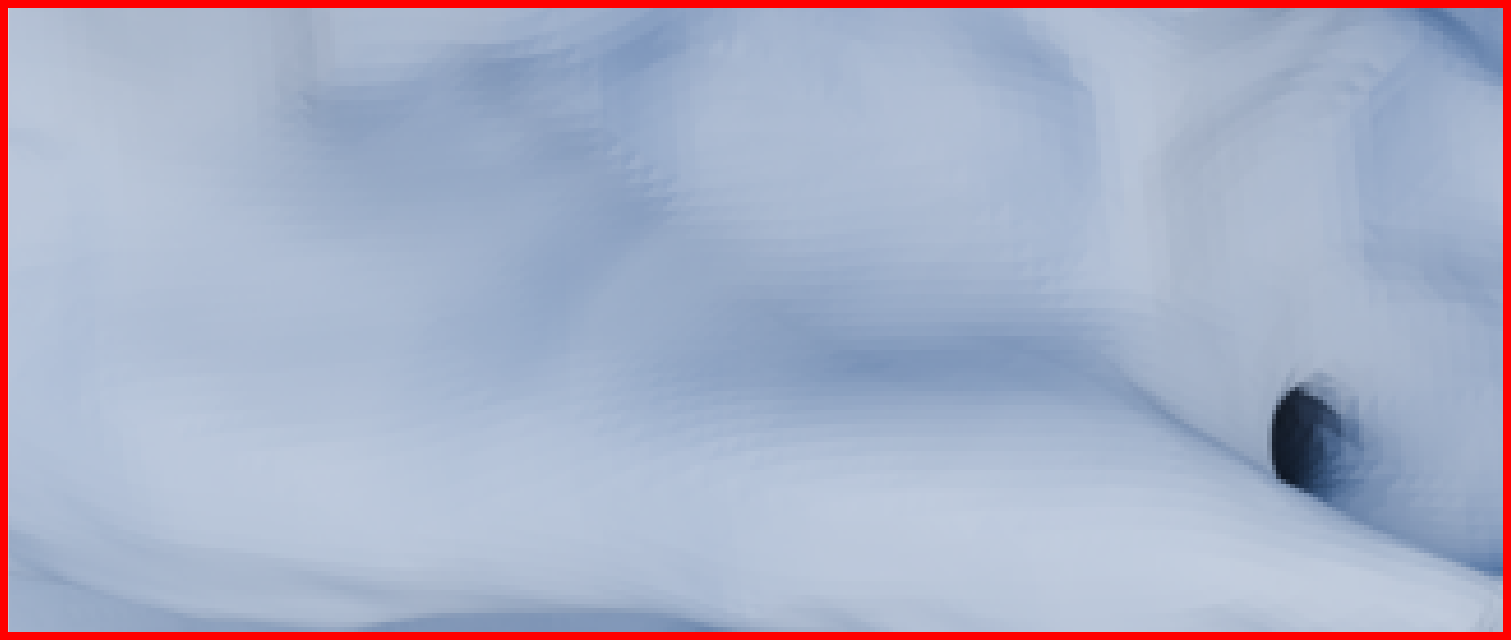} & 
\includegraphics[width=0.30\linewidth, page=3]{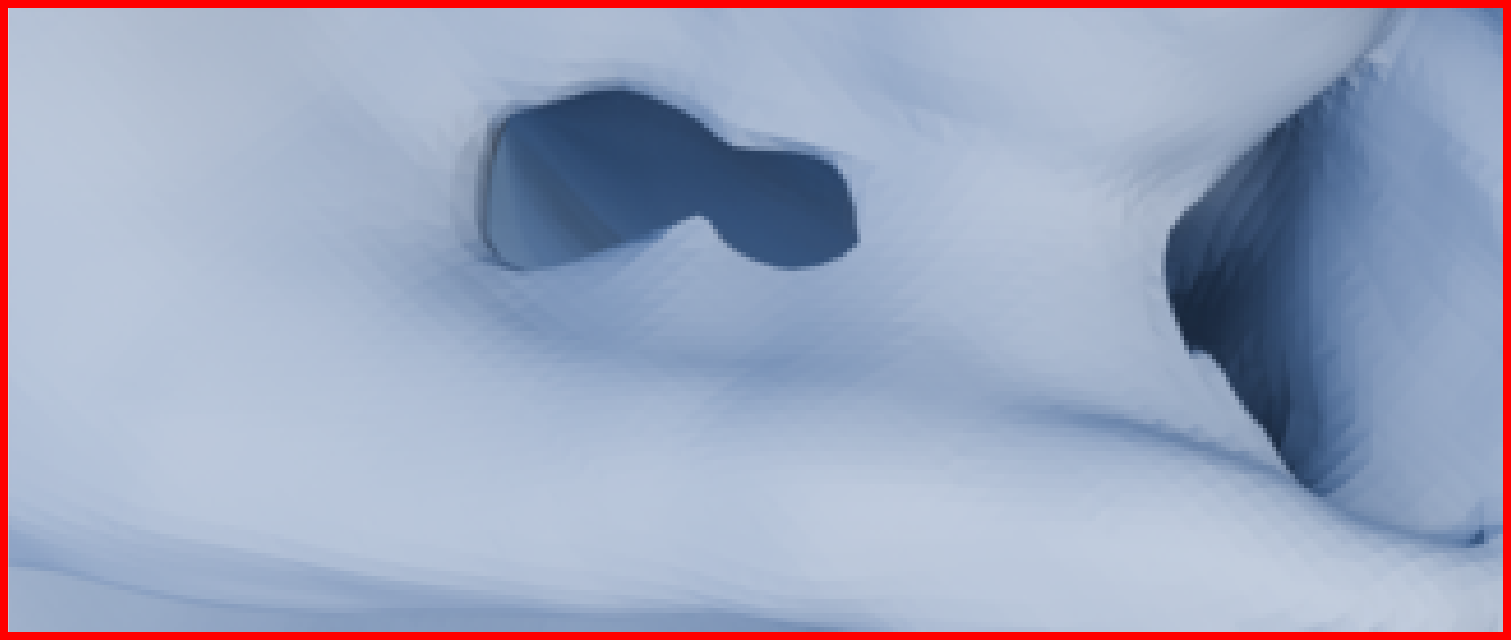} &
\includegraphics[width=0.30\linewidth, page=3]{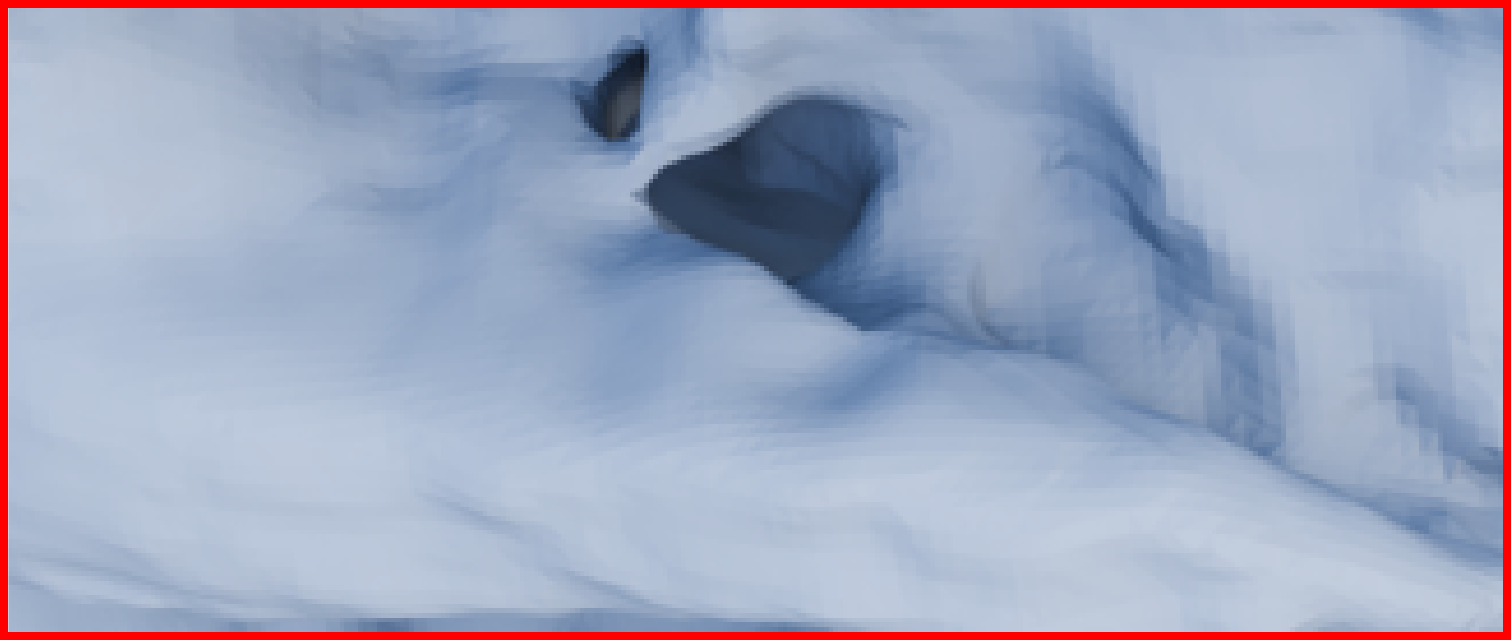} \\
NeuS & VolSDF & HF-NeuS \\[4pt]
\includegraphics[width=0.30\linewidth, page=2]{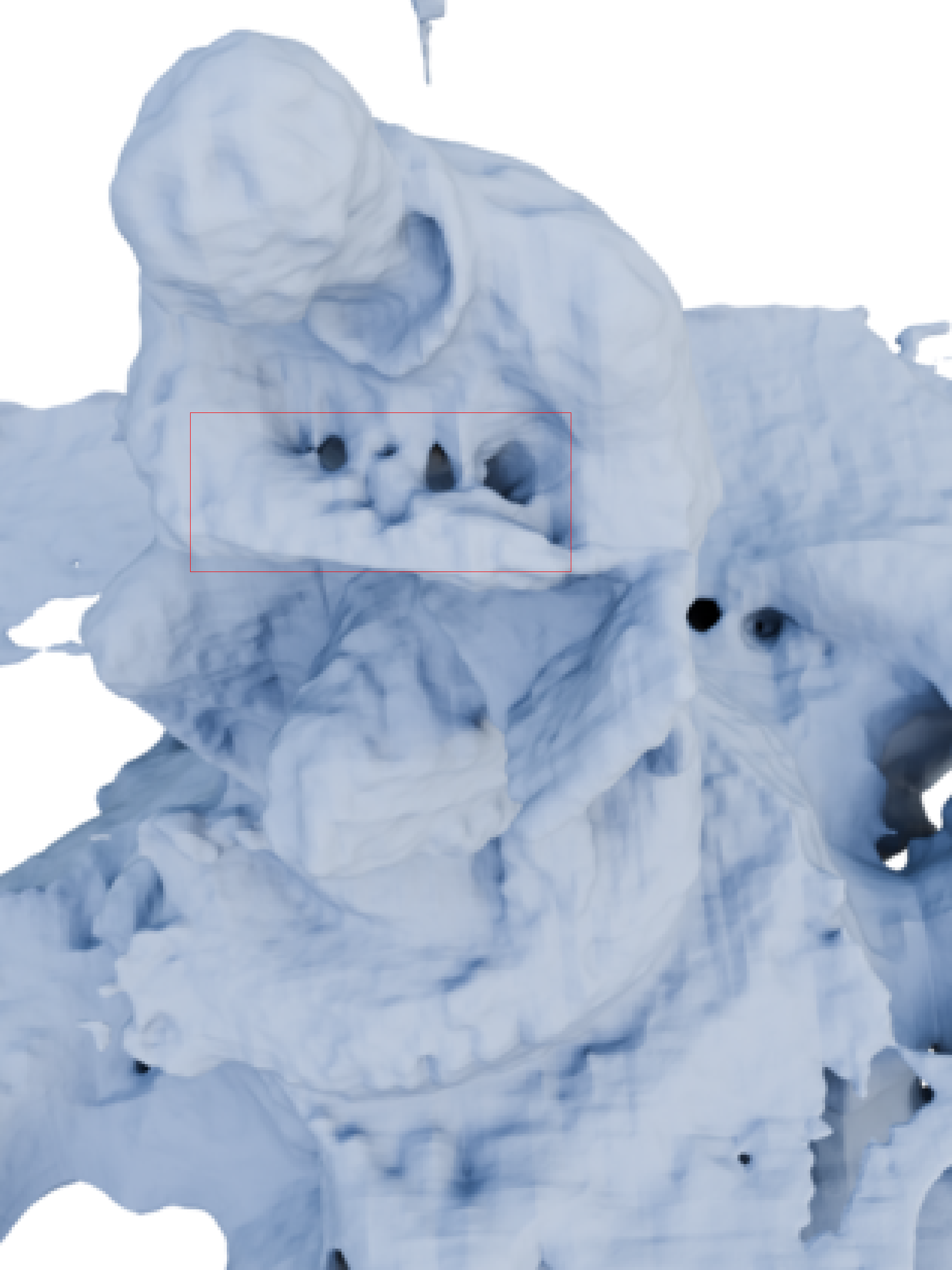} &
\includegraphics[width=0.30\linewidth, page=2]{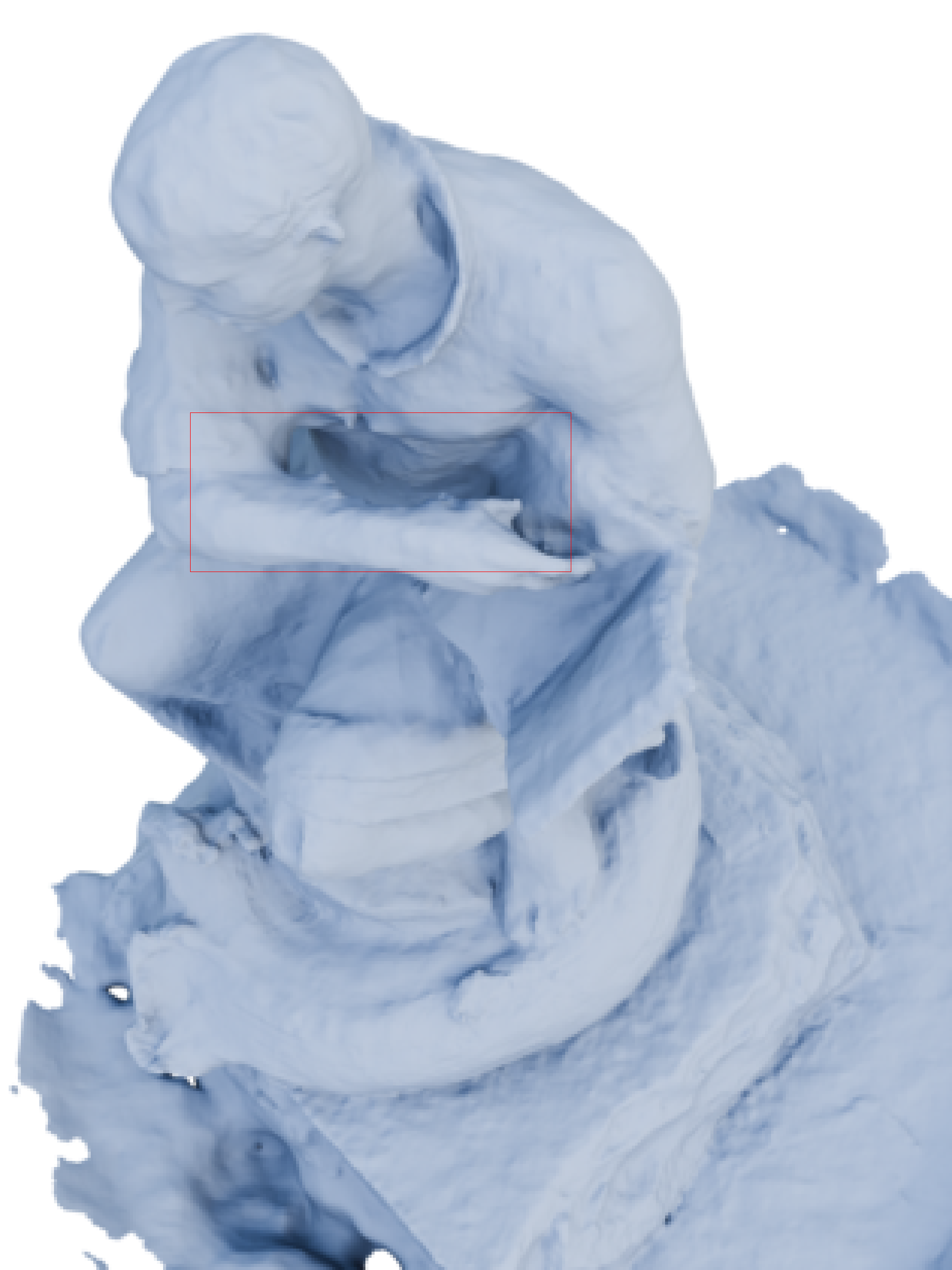} &
\includegraphics[width=0.30\linewidth, page=2]{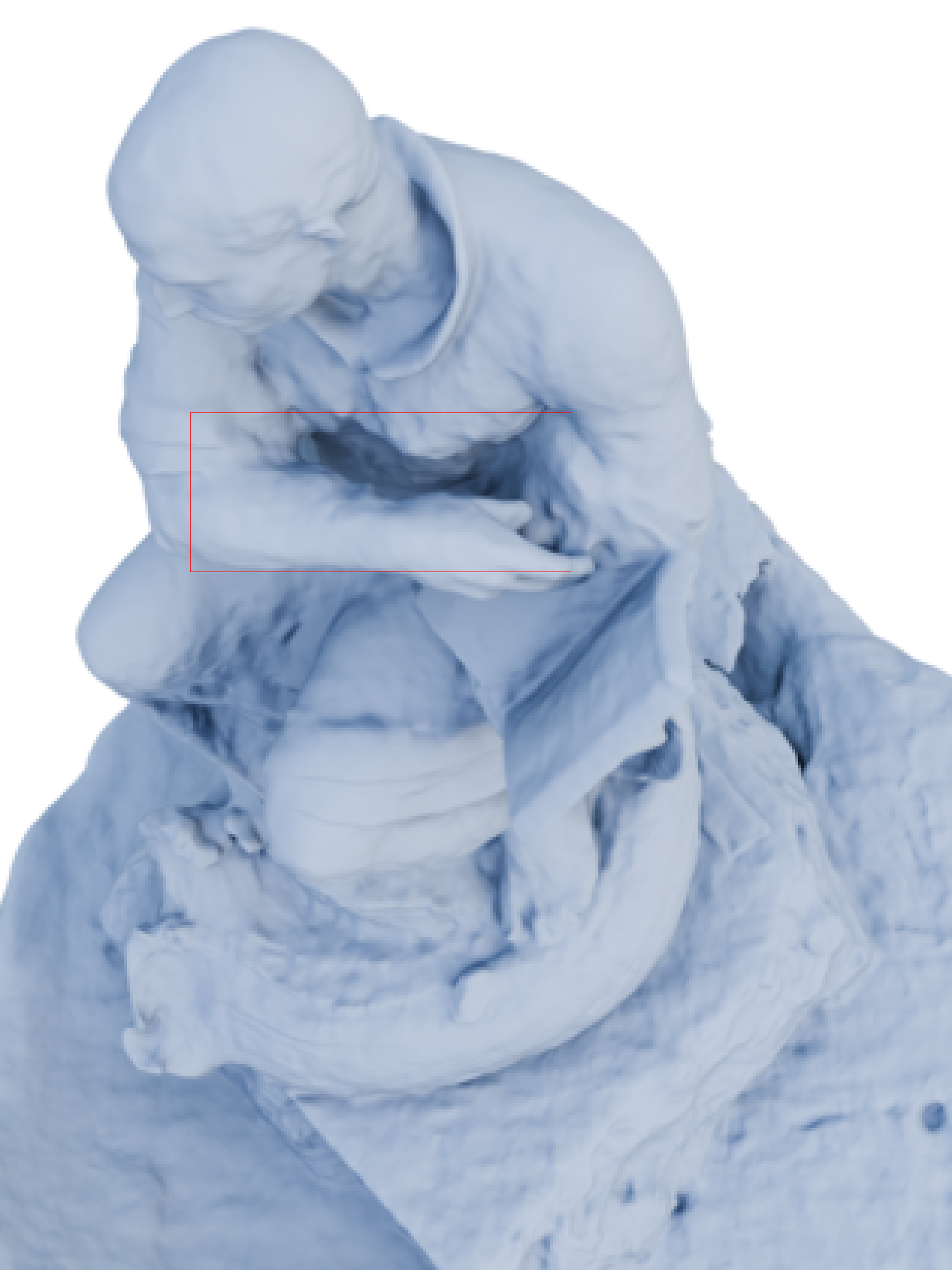} \\
\includegraphics[width=0.30\linewidth, page=3]{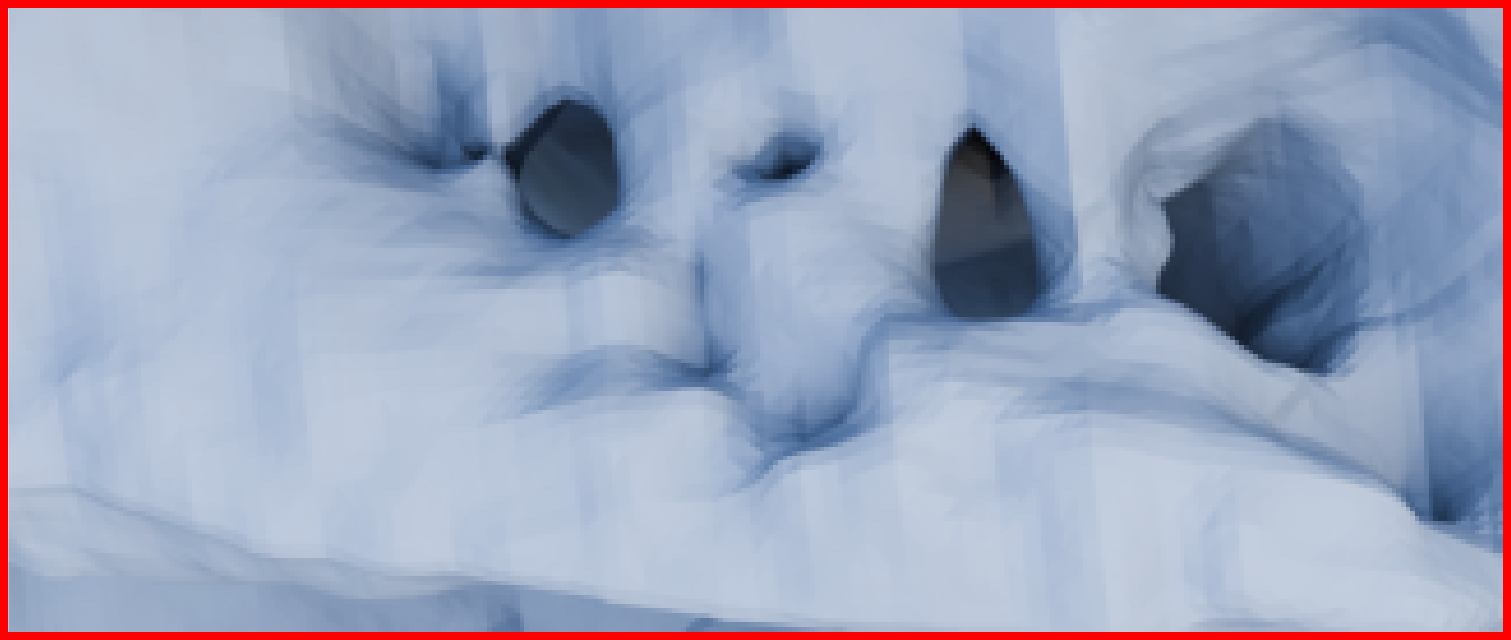} &
\includegraphics[width=0.30\linewidth, page=3]{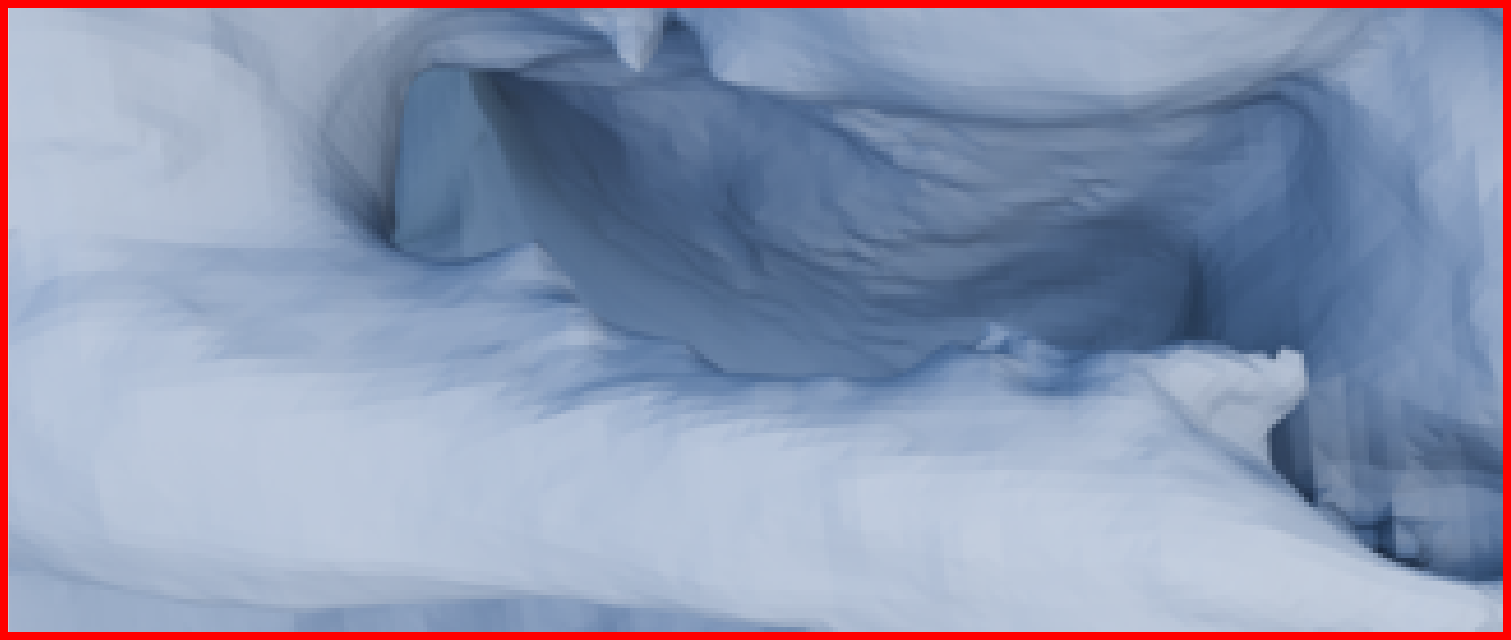} &
\includegraphics[width=0.30\linewidth, page=3]{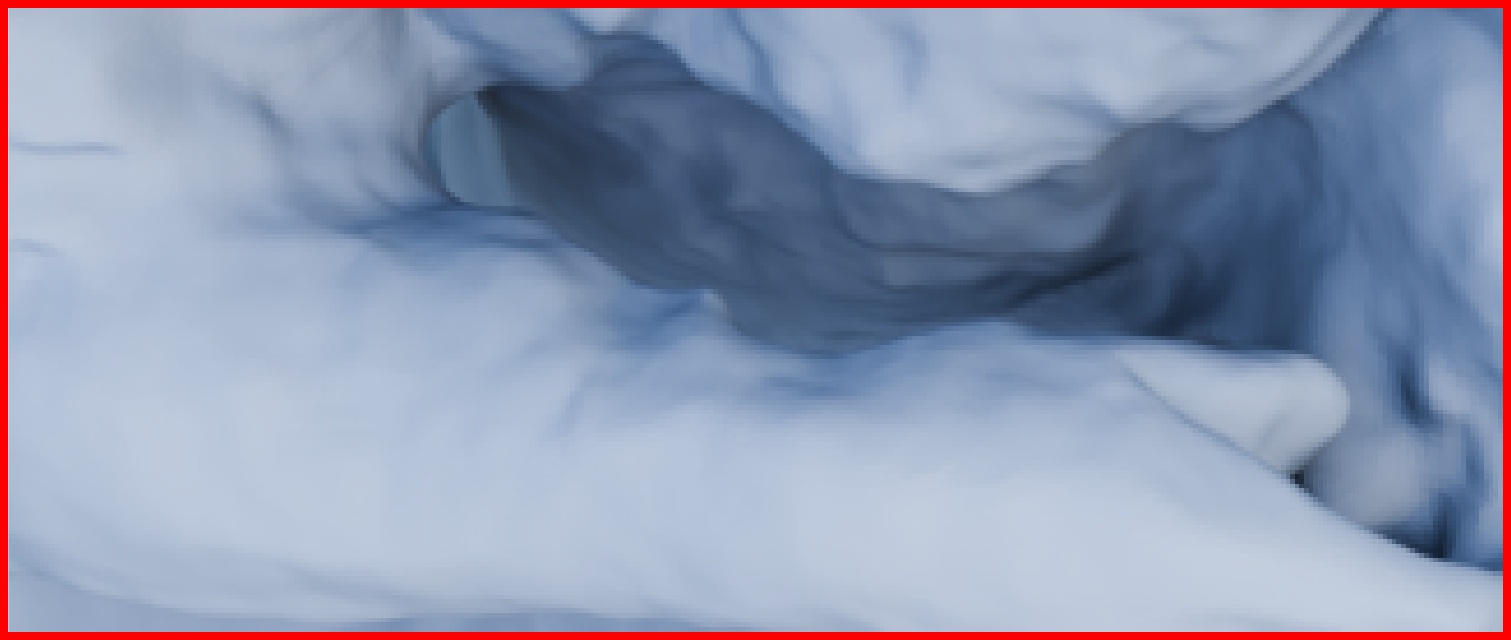} \\
    NeuS2 & Neuralangelo & Ours \\
\end{tabular}
\caption{Qualitative results on the Ignatius test case from the Tanks \& Temples dataset~\cite{Knapitsch2017TnT}. With the lack of the relevant view angles, only our method and Neuralangelo can correctly reconstruct the arm of the sculpture and separate the arm from the chest.} %
\label{fig:ignatius}
\end{figure}

\subsection{Ablation studies}
\begin{figure}[!htb]
\centering
\setlength{\tabcolsep}{3pt}
\begin{tabular}{ccc}
    \includegraphics[width=0.3\linewidth]{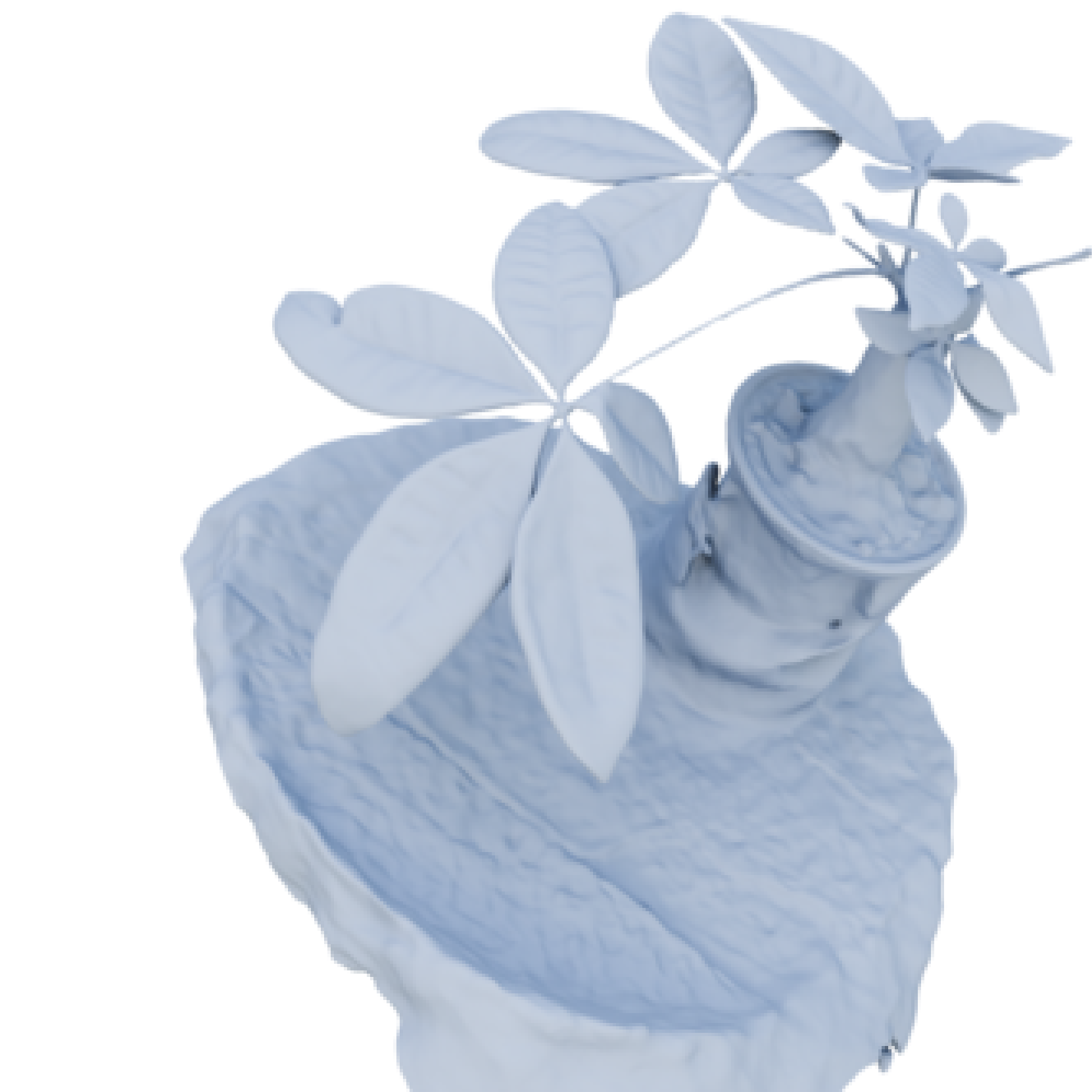} &
    \includegraphics[width=0.3\linewidth]{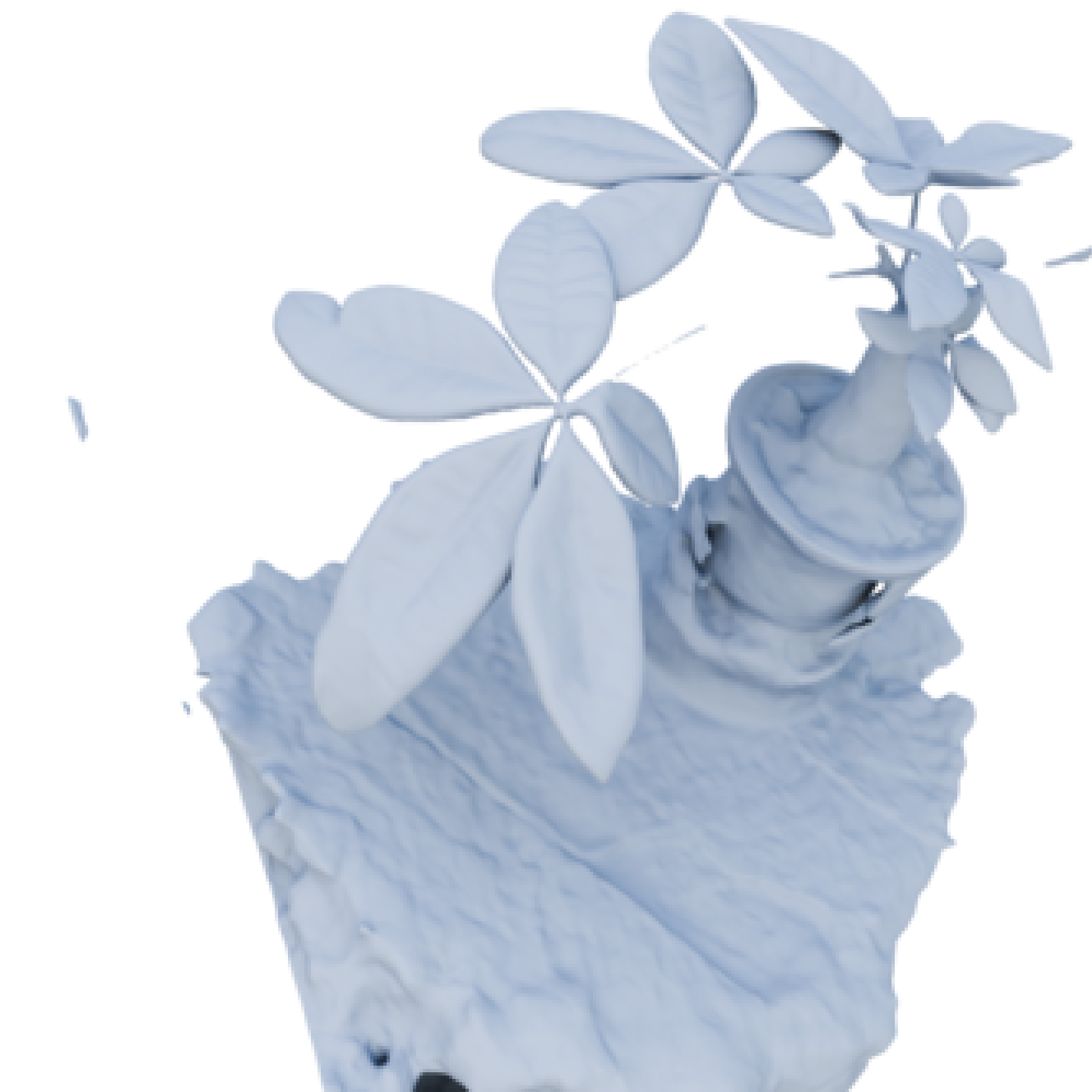} &
    \includegraphics[width=0.3\linewidth]{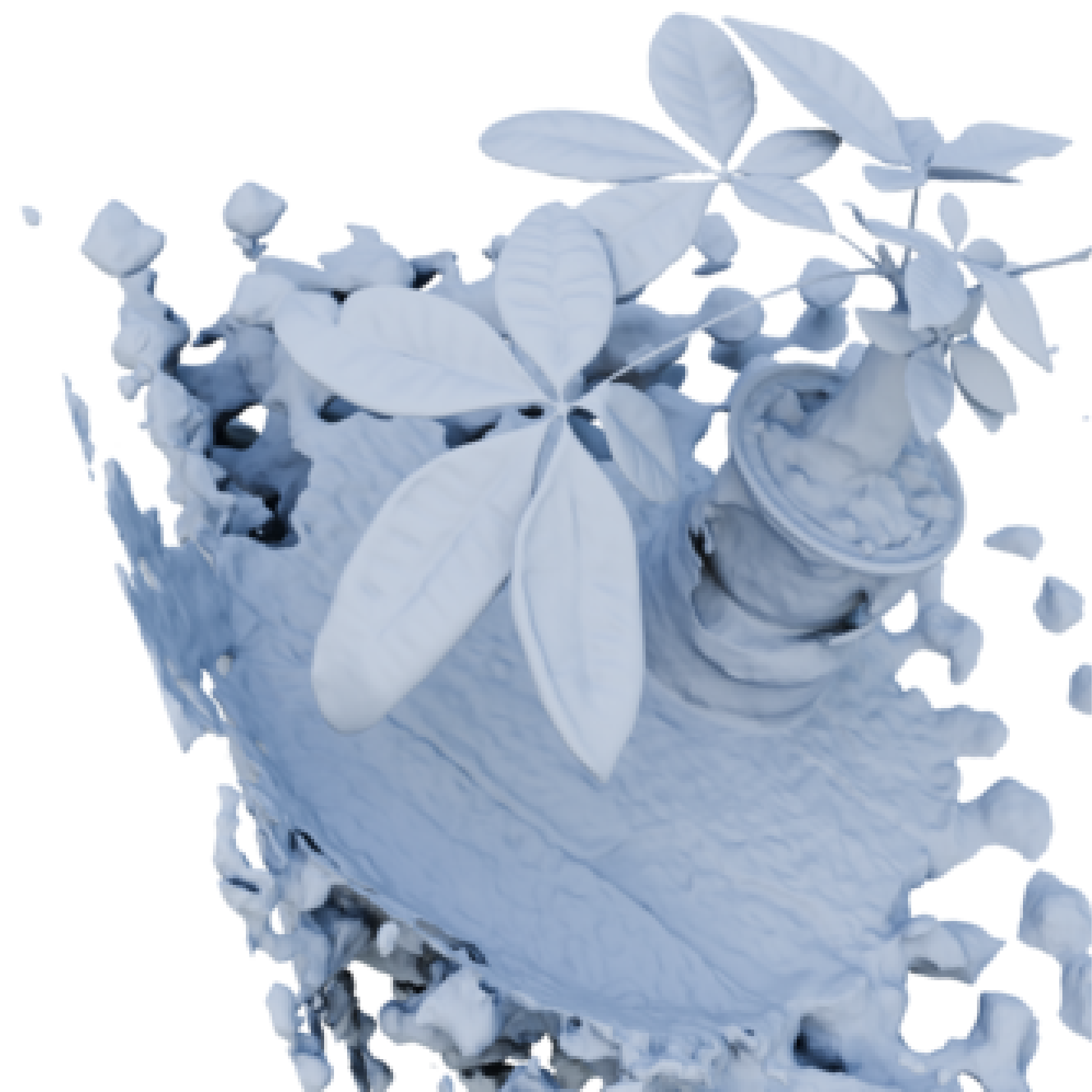} \\
    Full model & w/o near & w/o free
\end{tabular}
\caption{Qualitative comparison of our full model and our model without near surface regularization and free space regularization respectively. Without near surface regularization the model has difficulties learning thin structures, whereas without free space regularization the learned surface becomes noisy. }
\label{fig:ablation}
\end{figure}

In our method, the regularizers are categorized into two categories: near-surface regularization and free-space regularization. They reflect the interactions between the metric and phase branches. We provide ablation studies on these two types of regularizations. As shown in Figure~\ref{fig:ablation}, without near-surface regularization, the model has difficulties learning thin structures. Without free-surface regularization, the extracted mesh is noisy and contains floaters.

\begin{figure*}[!htb]
\centering
\setlength{\tabcolsep}{3pt}
\begin{tabular}{cccccc}
    \includegraphics[width=0.15\linewidth]{figs/ablation/mpfs1} &
    \includegraphics[width=0.15\linewidth]{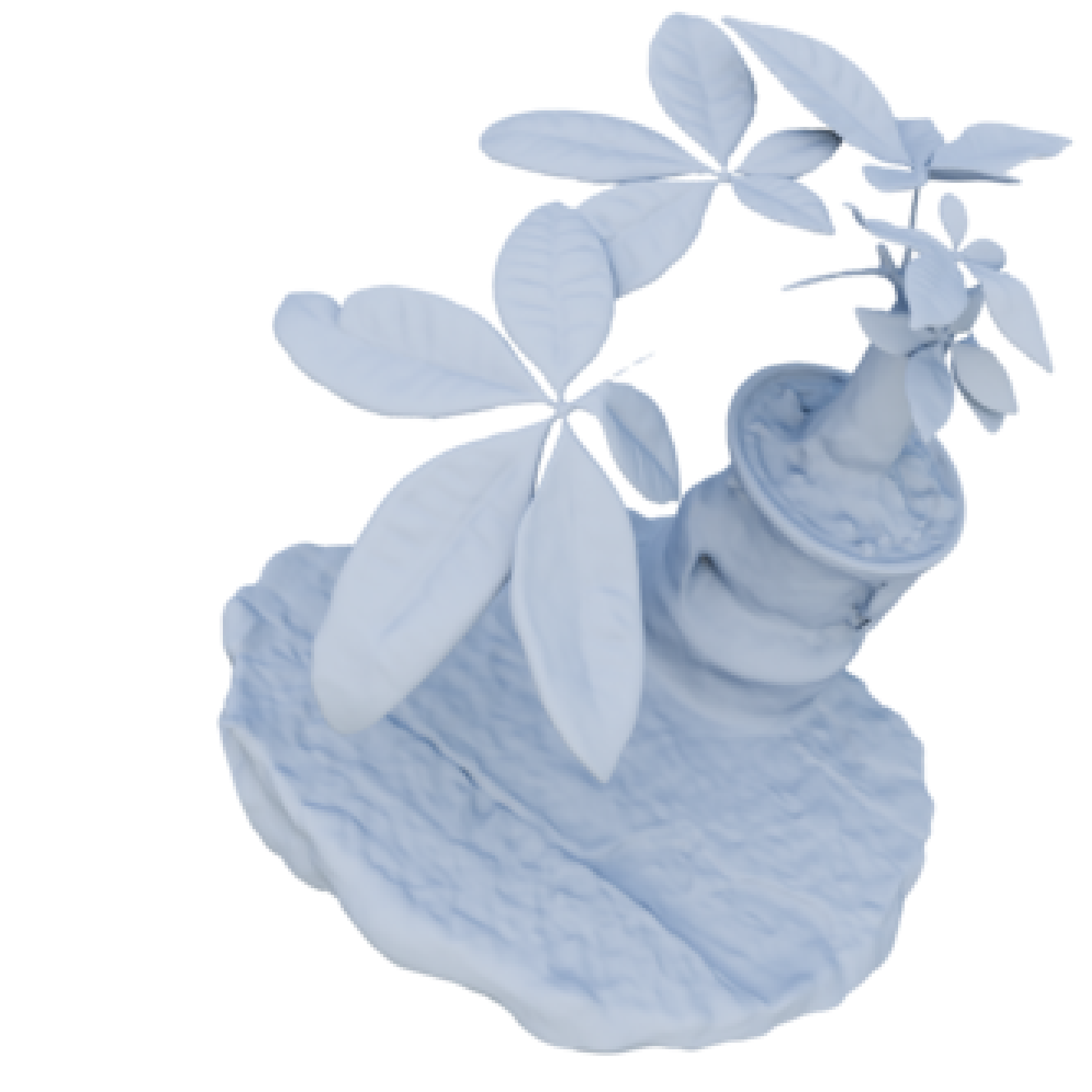} &
    \includegraphics[width=0.15\linewidth]{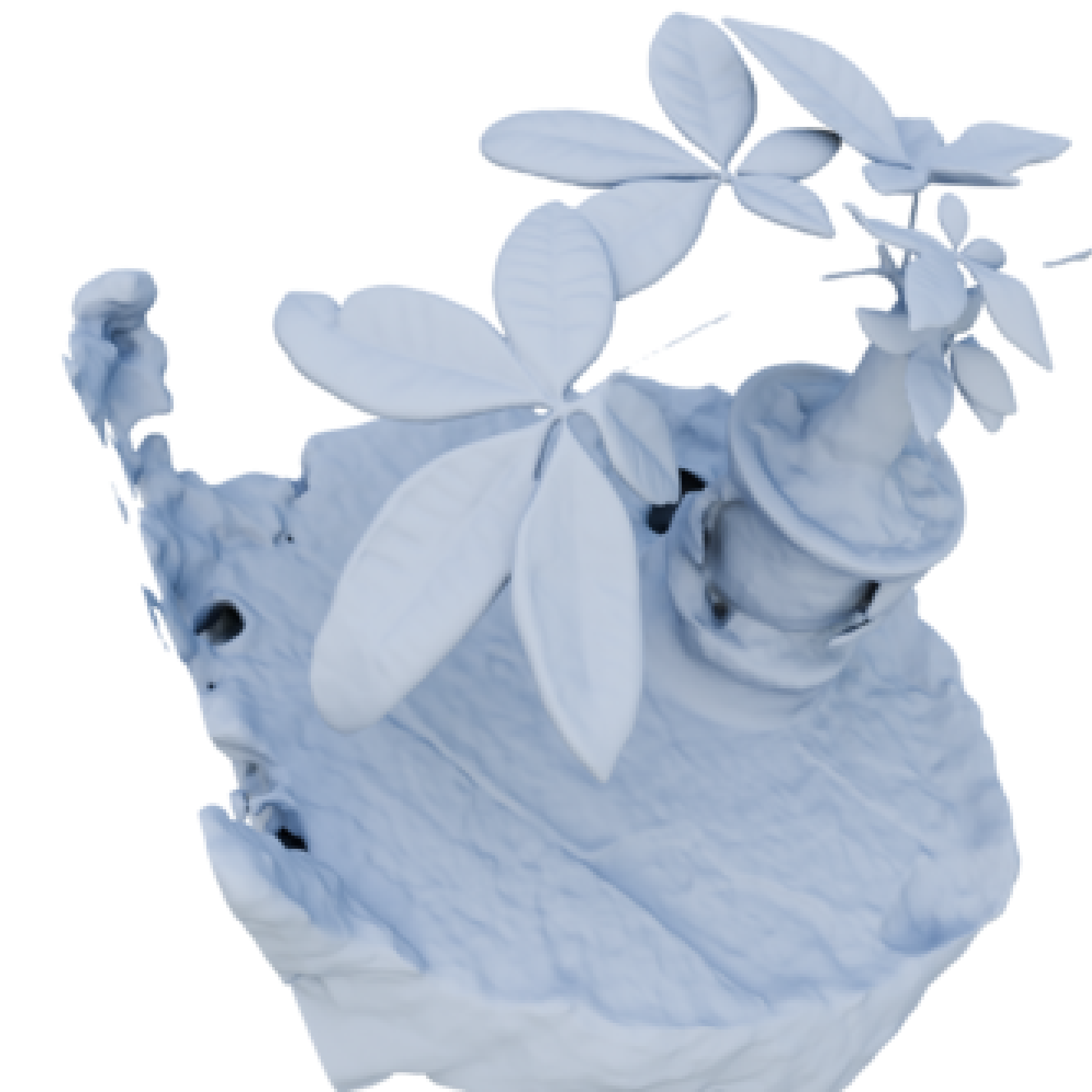} &
    \includegraphics[width=0.15\linewidth]{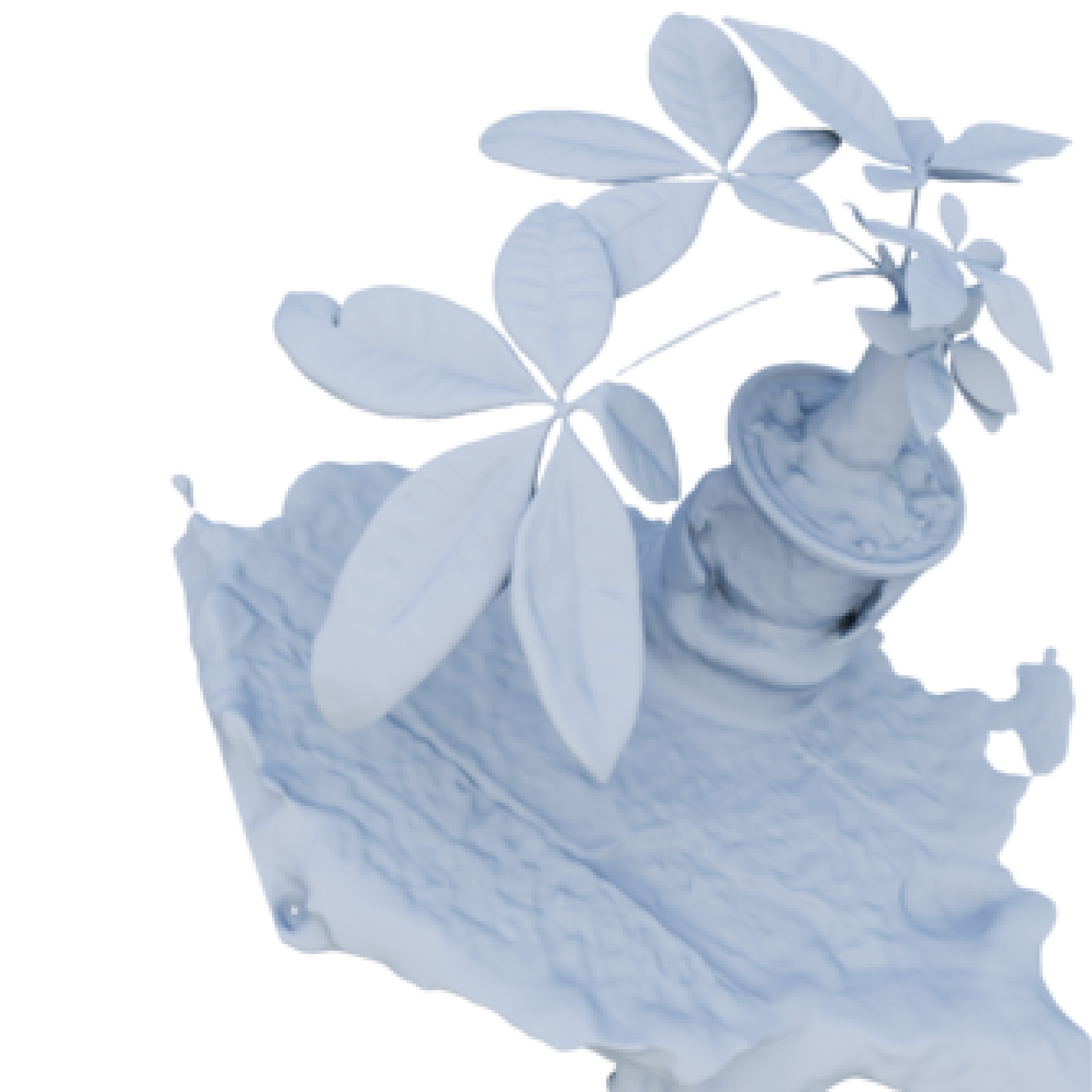} &
    \includegraphics[width=0.15\linewidth]{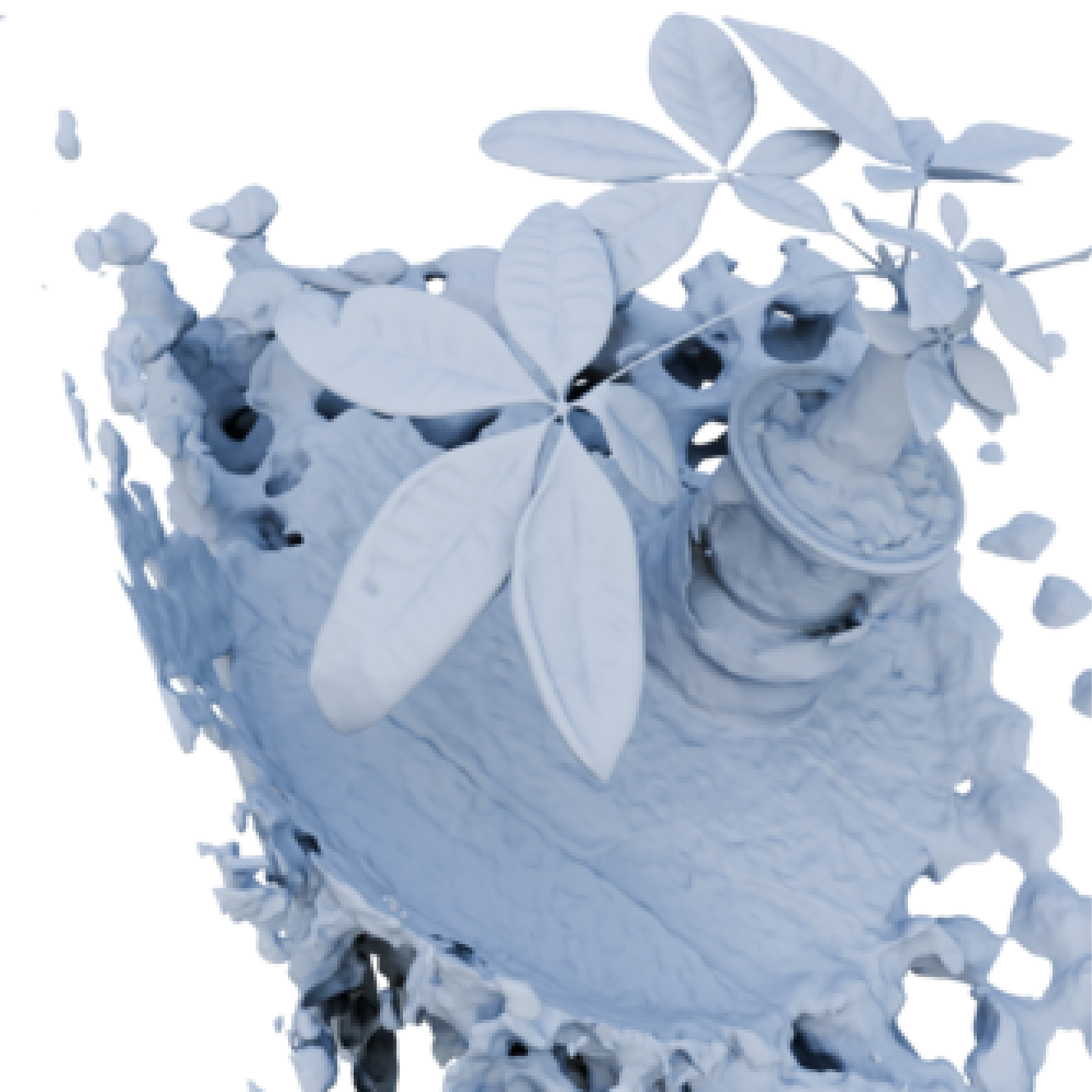} &
    \includegraphics[width=0.15\linewidth]{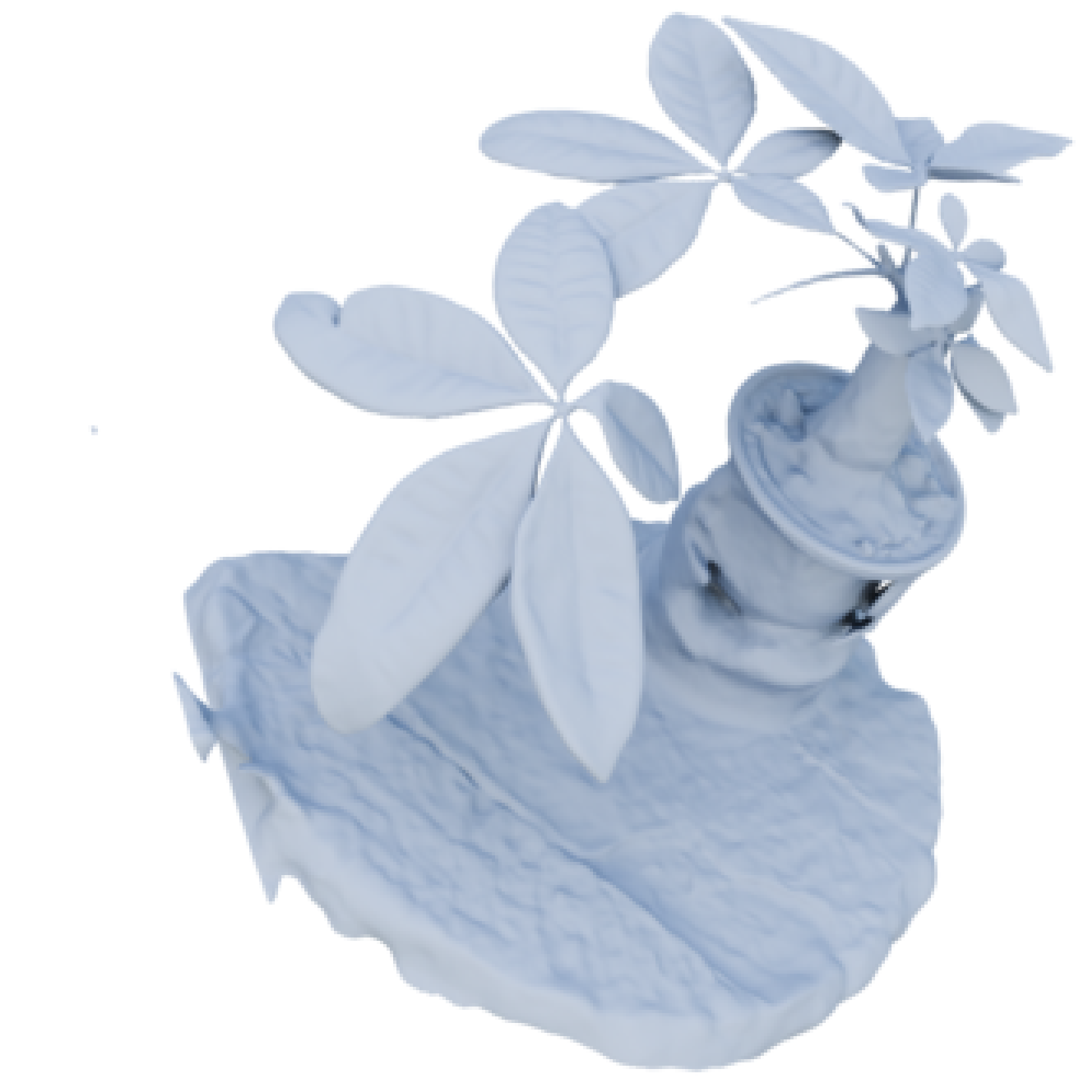} \\
    Full model & w/o $\mathcal{L}_{\text{anchor}}$ & w/o $\mathcal{L}_{\text{surf}}$ & w/o $\mathcal{L}_{\text{align}}$ & w/o $\mathcal{L}_{\text{eik}}$ & w/o $\mathcal{L}_{\text{far}}$
\end{tabular}
\caption{Qualitative comparison on the individual losses.}
\label{fig:ablation losses}
\end{figure*}

\subsubsection{The effect of near surface regularizers}
\begin{figure}[!htb]
    \centering
    \includegraphics[width=0.4\linewidth, page=2]{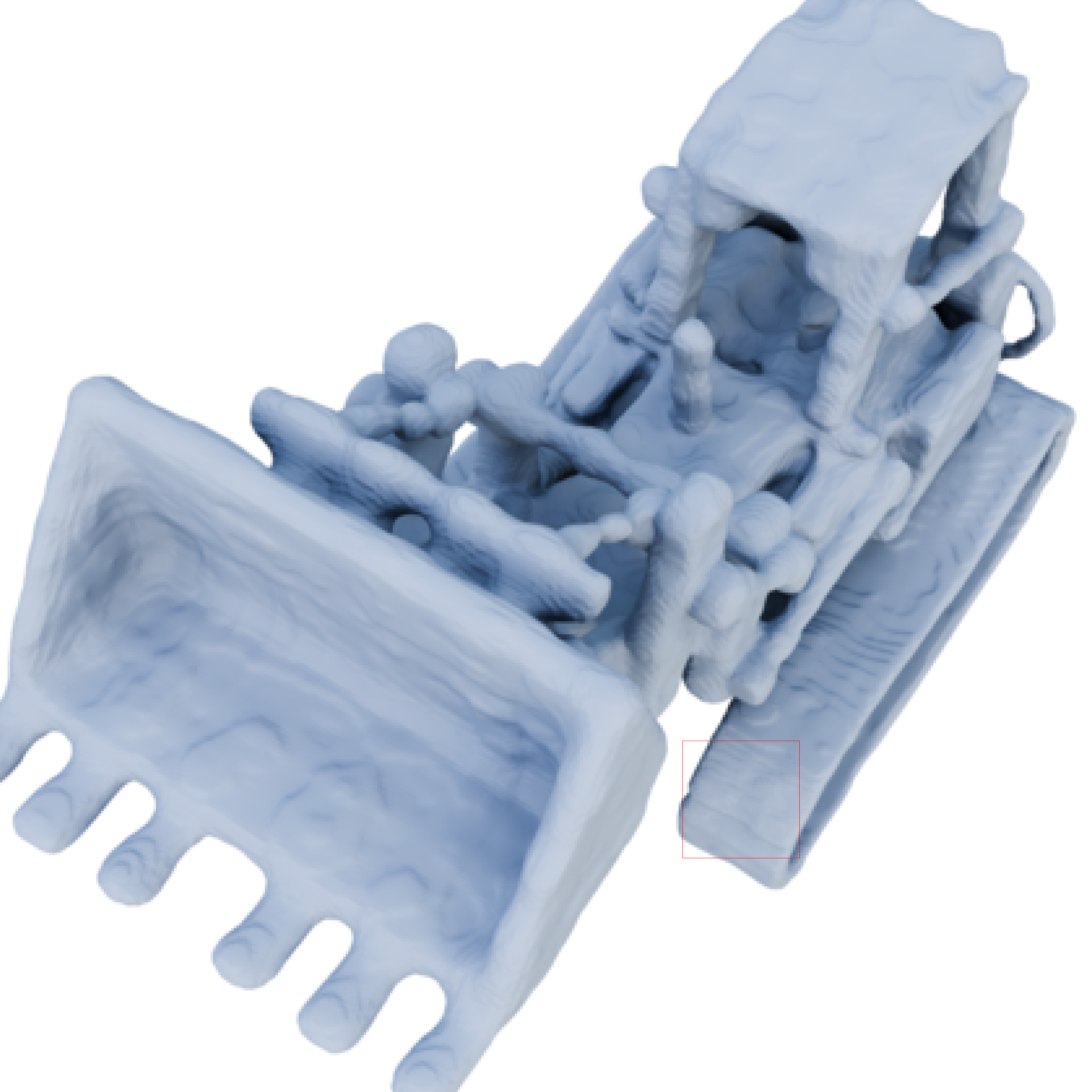}
    \includegraphics[width=0.4\linewidth, page=3]{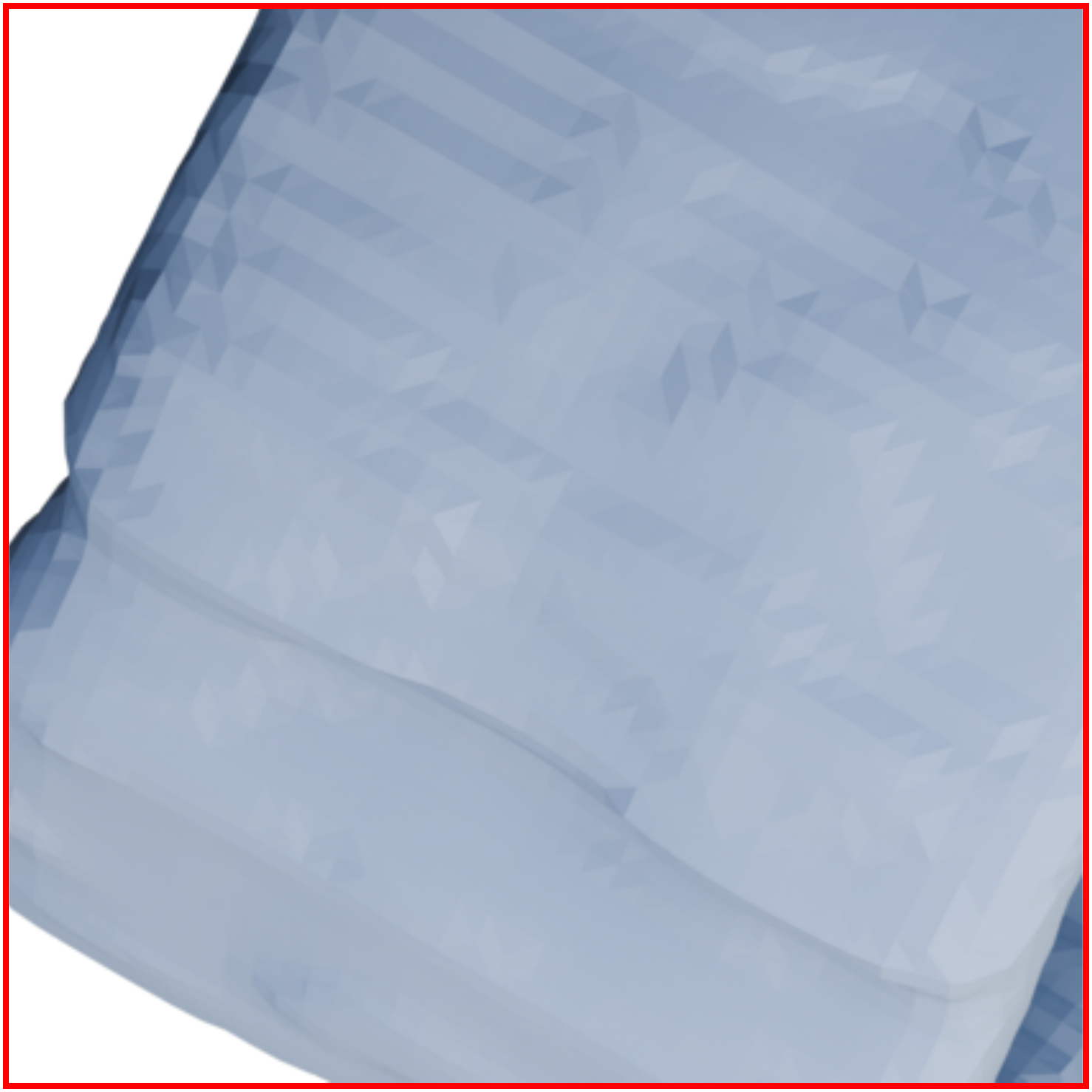} \\
    w/o $\mathcal{L}_{\text{surf}}$
    \caption{Omitting $\mathcal{L}_{\text{surf}}$ will make the training unstable, resulting in a completely degraded reconstruction quality compared to other methods.}
    \label{fig:loss surf}
\end{figure}

Near-surface regularization ensures SDF-like behavior near the surface of the object which is required for representing the intricate details of the object. Figure~\ref{fig:ablation losses} on the left shows the results of removing individual near-surface regularizers. Without the regularizations, the metric field and the phase field cannot be adequately separated and constrained, often leading to suboptimal learning results and consequently cannot accurately reconstruct the intricate details of the object. We specially observed that $\mathcal{L}_{\mathrm{surf}}$, which enforces the Eikonal constraint of the composite field $\phi$ and the phase field $\theta$, is critical for the stability of the surface. As shown in Figure~\ref{fig:loss surf}, removing $\mathcal{L}_{\mathrm{surf}}$ completely deteriorates the reconstruction quality.

\subsubsection{The effect of free space regularizers}
The Free-space regularizers aims to stabilize the neural fields and prevent floaters. Figure~\ref{fig:ablation losses} on the right demonstrates the effect of removing the free-space regularizers $\mathcal{L}_{\text{eik}}$ and $\mathcal{L}_{\text{far}}$.
We observe that removing $\mathcal{L}_{\mathrm{eik}}$, which encourages the metric field $r$ to exhibit UDF behavior and abide to the Eikonal constraint away from the surface, will make the implicit geometry unstable, introducing floating geometry into the extracted mesh in Figure~\ref{fig:ablation losses}, even making training completely collapse in the Lego and Shoe test case.
$\mathcal{L}_{\text{far}}$ suppresses large regions of near-zero values. This is important for the implicit geometry field to accurately represent the geometry as well as for the mesh extraction method to successfully extract the mesh surface. We observe that removing $\mathcal{L}_{\text{far}}$ for the metric and phase fields will make our method fail to reconstruct the thin stem.

\subsection{Remarks on Mesh Extraction Methods}

The choice of mesh extraction technique is an important part of the reconstruction pipeline. For 3DGS-based reconstruction methods, a majority of them rely on TSDF fusion technique, and for implicit field-based reconstruction methods, standard marching cubes can be used. Marching Tetrahedra is another iso-surface extraction method that works on tetrahedral grids. Compared to the classical Marching Cubes algorithm, it avoids certain ambiguity cases and is often used in pipelines where surfaces are reconstructed from volumetric discretizations or tetrahedral meshes. 

TSDF fusion, as used by many 3DGS-based reconstruction methods, works by first rendering depth maps and then integrating depth maps into a volumetric truncated signed distance field (TSDF) and extracting the zero level set using volumetric fusion techniques~\cite{Brian1996Volumetric,Newcombe2011KinectFusion}. 

For PGSR and GeoSVR, meshes are extracted using TSDF fusion since this is the only extraction method supported by these pipelines. 

GGGS supports both TSDF fusion and Marching Tetrahedra. 
In our experiments, we adopt the latter for its higher efficiency and lower memory usage. 
GaussianUDF represents geometry using an implicit unsigned distance field and therefore requires the MeshUDF algorithm~\cite{guillard2022udf} for surface extraction.

In contrast, NeuS and our method are based on implicit signed fields, allowing direct iso-surface extraction using the standard Marching Cubes algorithm. 
This approach is both efficient and robust and avoids the additional post-processing steps often required by TSDF-based pipelines.

Overall, a majority of 3DGS-based surface reconstruction methods still rely on TSDF Fusion pipeline. This pipeline brings two problems.

The first problem is the resource consumption of such method. The voxel grid produced by integrating depth maps is usually very dense and requires a very large amount of memory space, as a result, they often trigger \texttt{Segmentation Fault} errors. For example, on the TAAS-synthetic dataset, GeoSVR can learn the radiance field but fails to generate triangle meshes from the learned Gaussian primitives. The TSDF fusion step repeatedly reports \texttt{Segmentation Fault} errors during mesh extraction. Even if the error is not triggered, the final mesh file after extraction is also very large, often reaching hundreds of megabytes even into gigabytes. Such resource-intensive extraction procedures limit their practical use in large-scale experiments. Table~\ref{tab:meshextract} presents an overview of mesh extraction methods used by different reconstruction methods.

\begin{table}[!htb]
    \centering
    \caption{
    Comparison of mesh extraction strategies used by different reconstruction methods. 
    }
    \label{tab:meshextract}
    \setlength\tabcolsep{5pt}
    \begin{tabular}{c|c|c|c|c}
        \toprule
        Method & \makecell{Marching\\Cubes} & \makecell{Marching\\Tetrahedra} & TSDF Fusion & MeshUDF \\
        \midrule
        PGSR & & & $\checkmark$ & \\
        GeoSVR & & & $\checkmark$ & \\
        GaussianUDF & & & & $\checkmark$ \\
        GGGS & & $\checkmark$ & $\checkmark$ & \\
        NeuS & $\checkmark$ & & & \\
        HF-NeuS & $\checkmark$ & & & \\
        Neuralangelo & $\checkmark$ & & & \\
        Ours & $\checkmark$ & & & \\
        \bottomrule
    \end{tabular}
\end{table}

\begin{figure*}[!htb]
    \setlength{\tabcolsep}{0.005\linewidth}
    \centering
    \begin{tabular}{cc|cc|cc|c}
\includegraphics[width=0.125\linewidth]{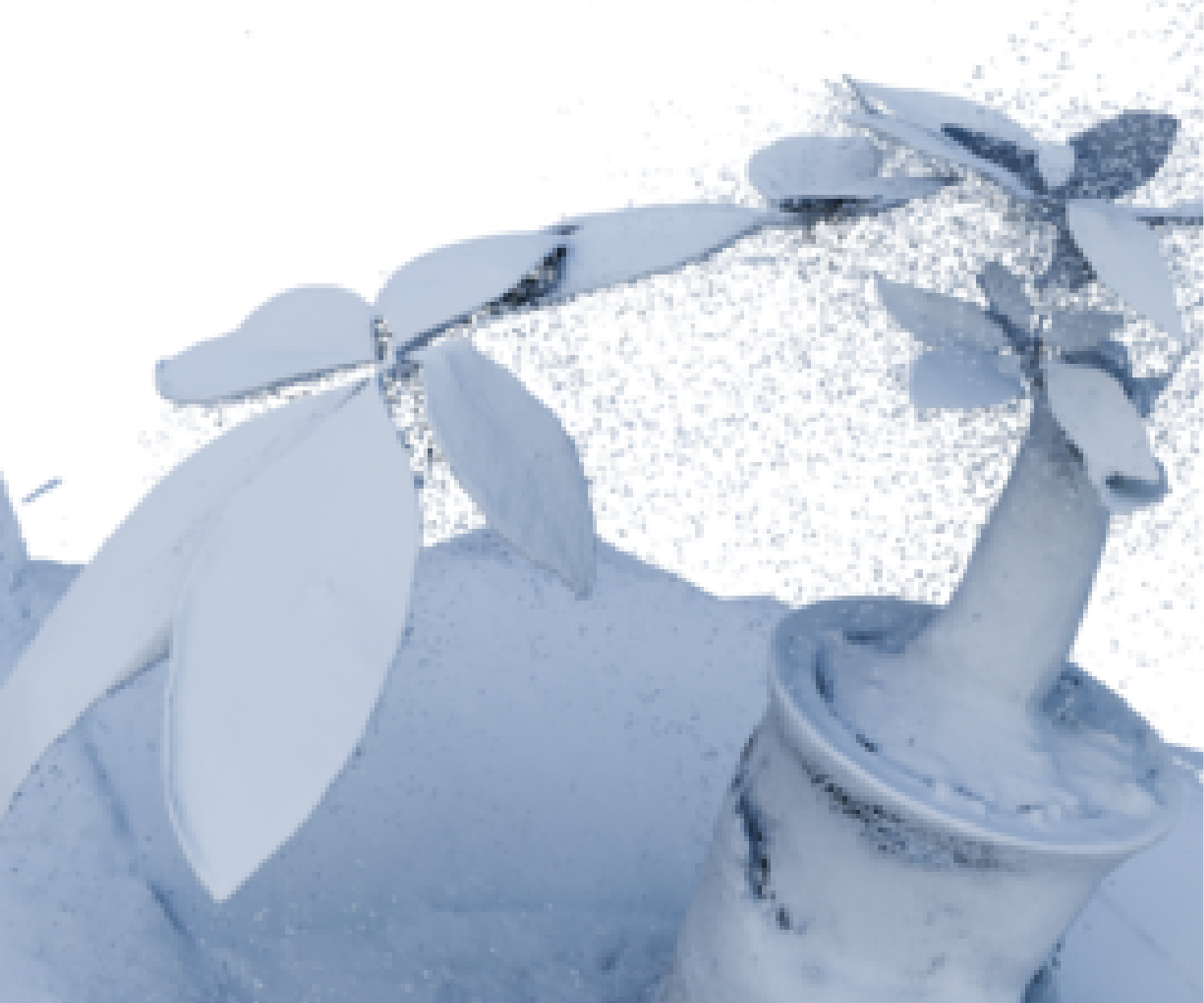} &
\includegraphics[width=0.125\linewidth,]{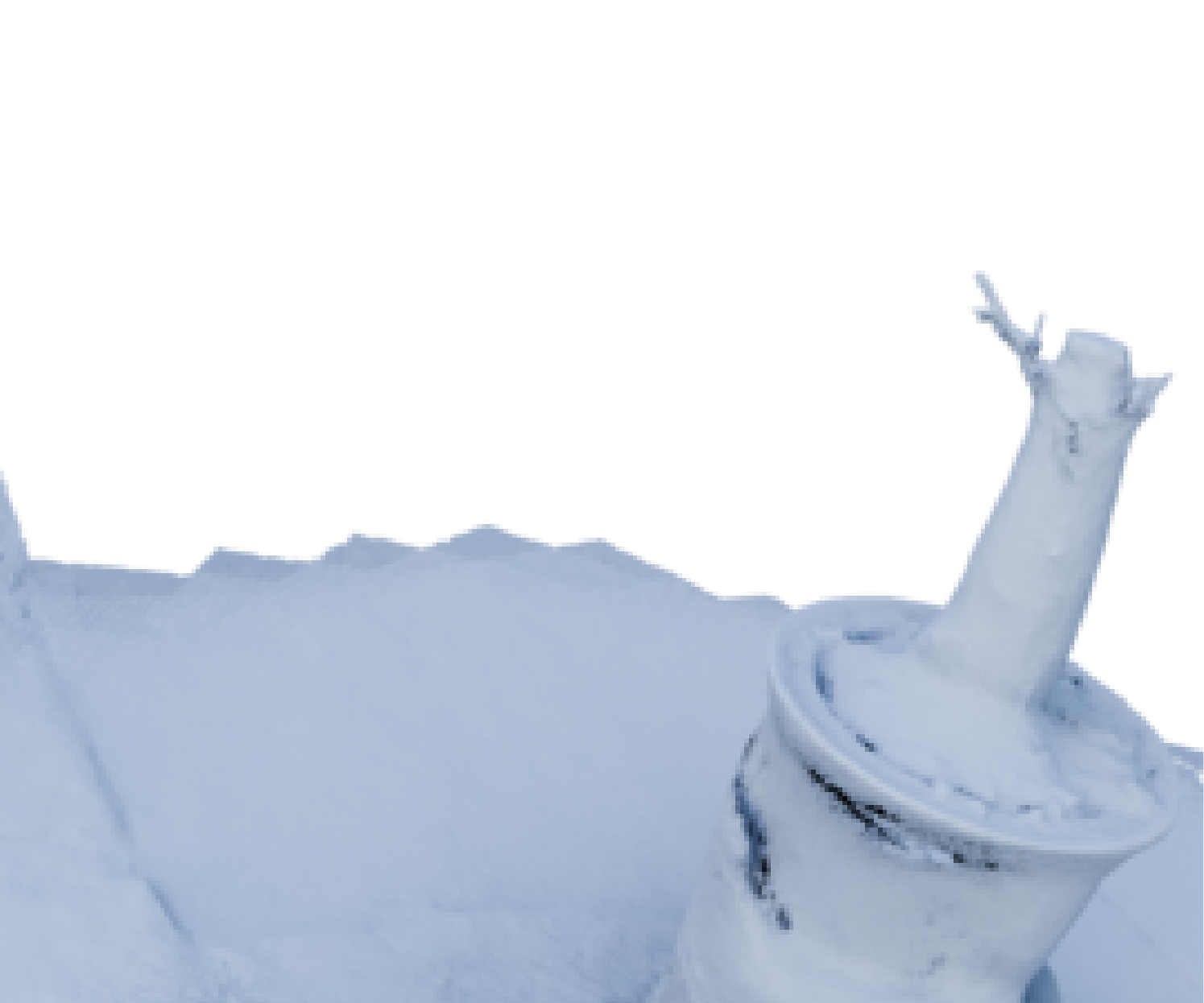}\hspace{1pt} &
\hspace{1pt}%
\includegraphics[width=0.125\linewidth]{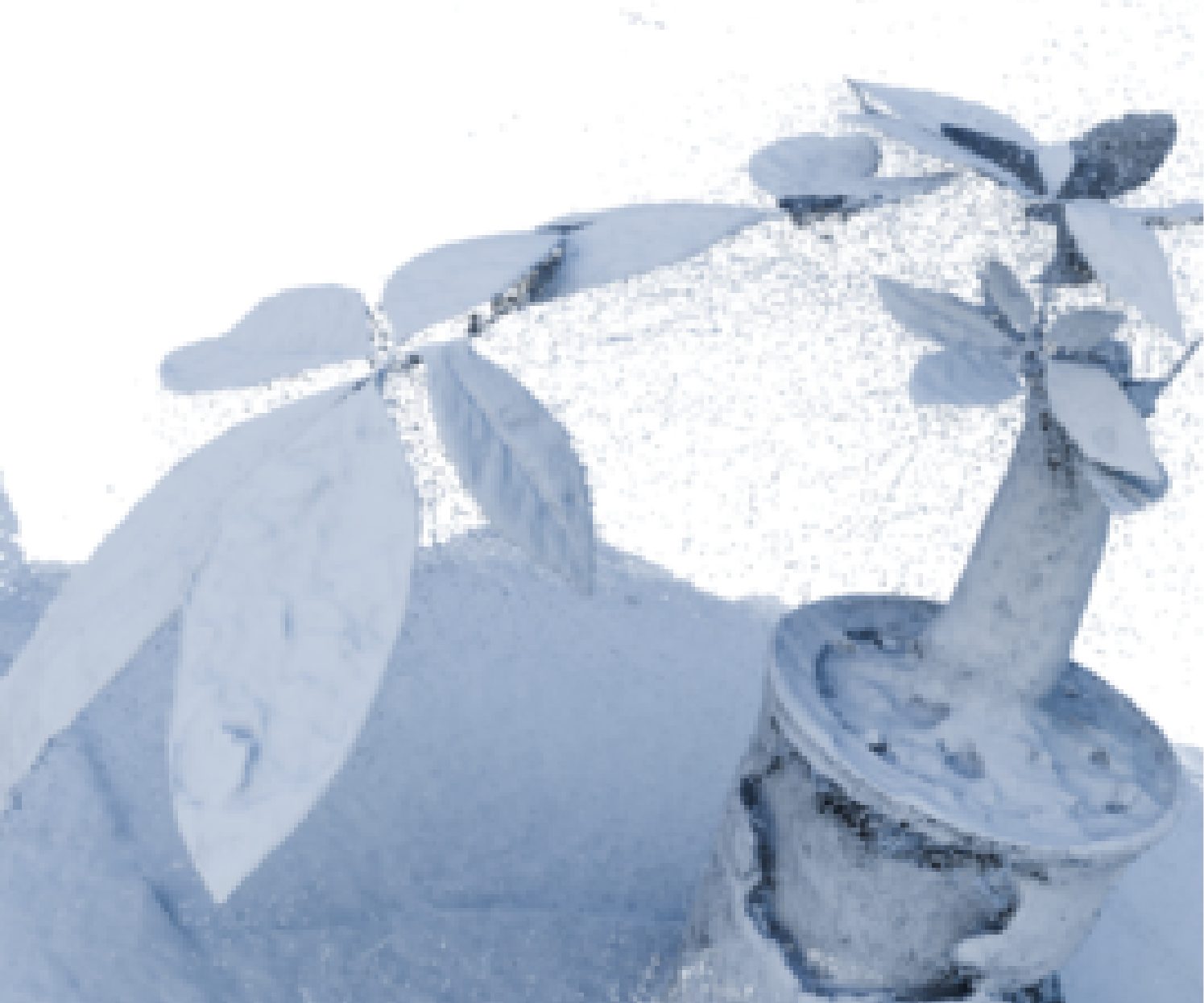} &
\includegraphics[width=0.125\linewidth]{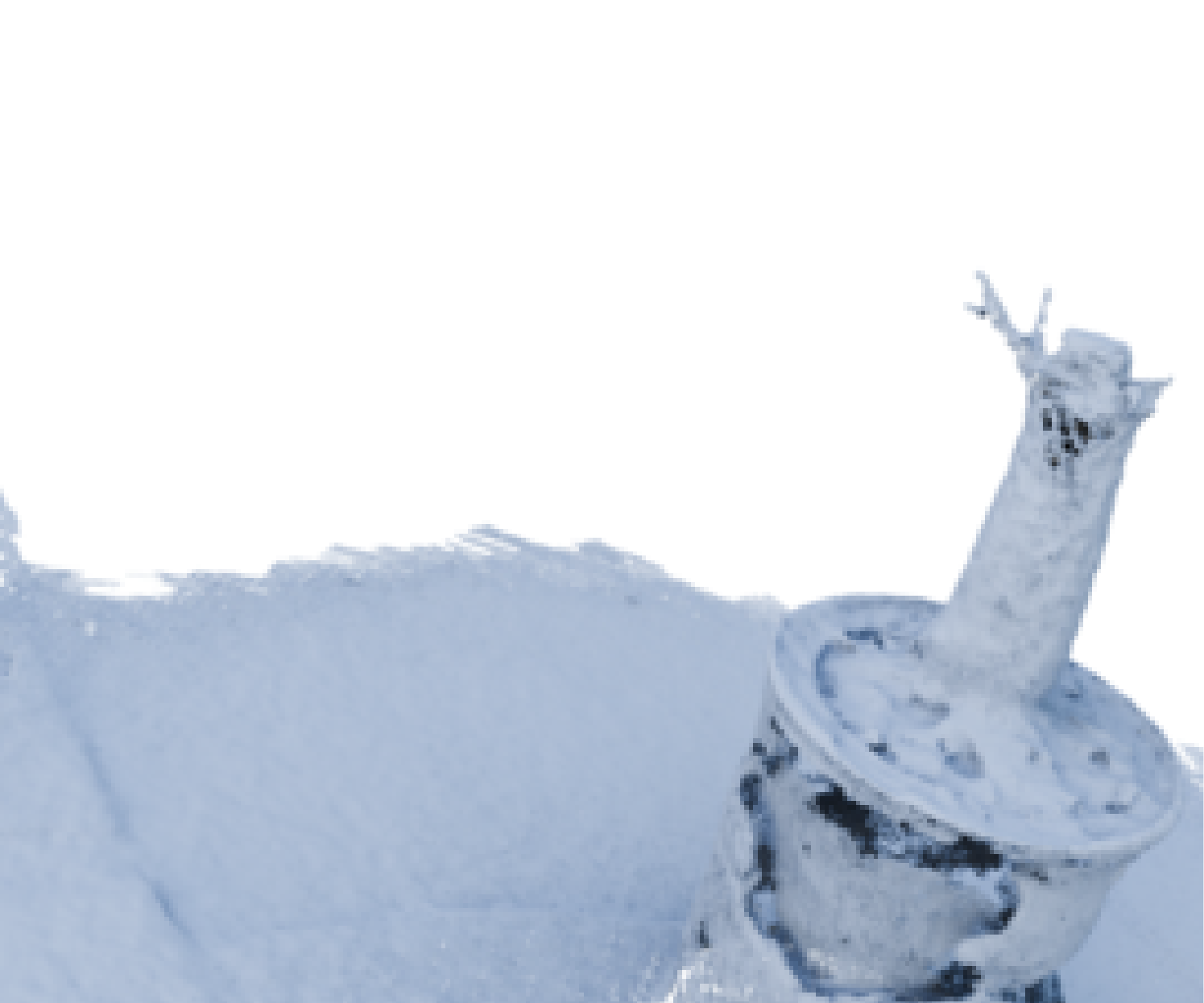}\hspace{1pt} &
\hspace{1pt}%
\includegraphics[width=0.125\linewidth]{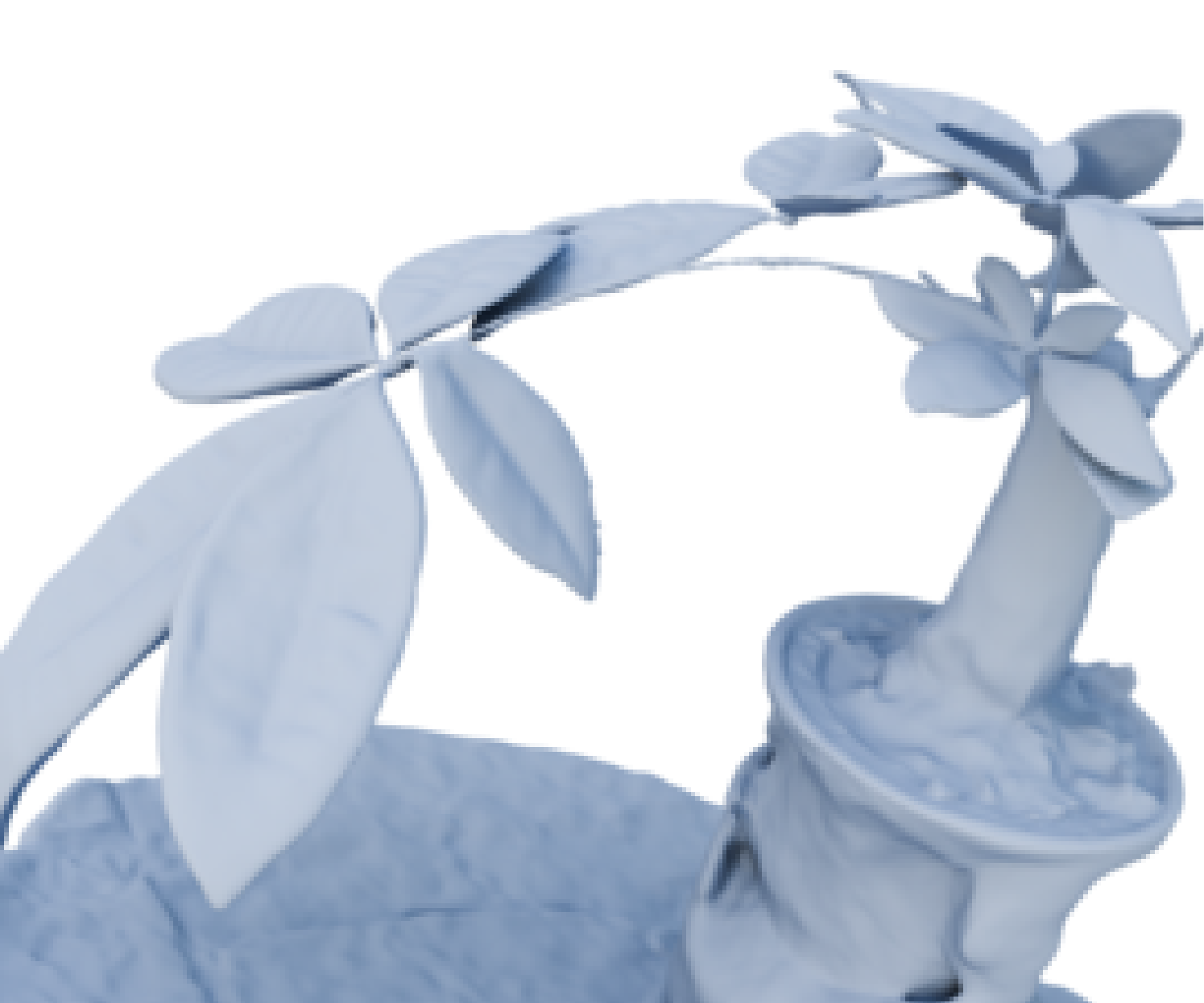} &
\includegraphics[width=0.125\linewidth]{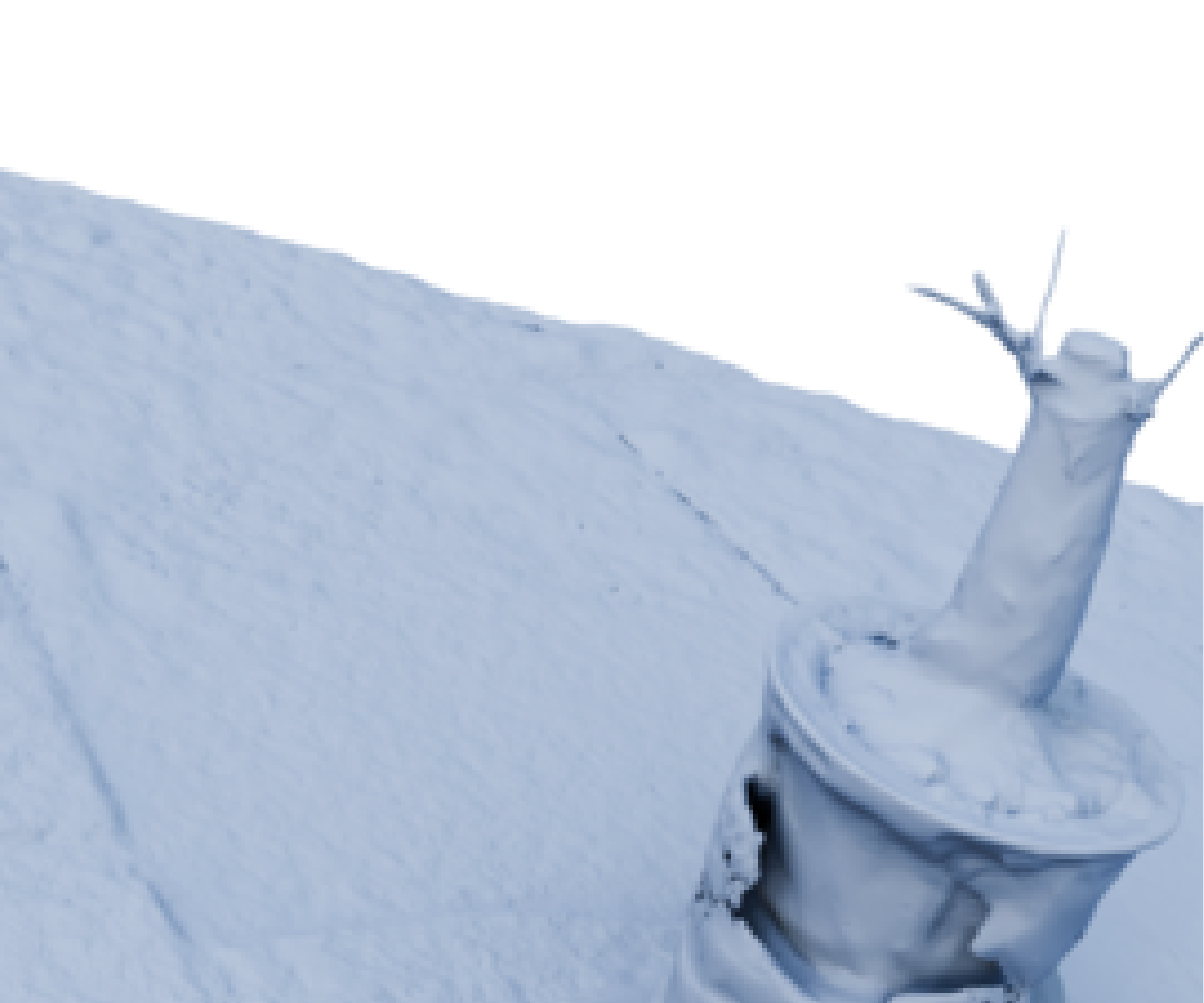}\hspace{1pt} & 
\hspace{1pt}\includegraphics[width=0.125\linewidth]{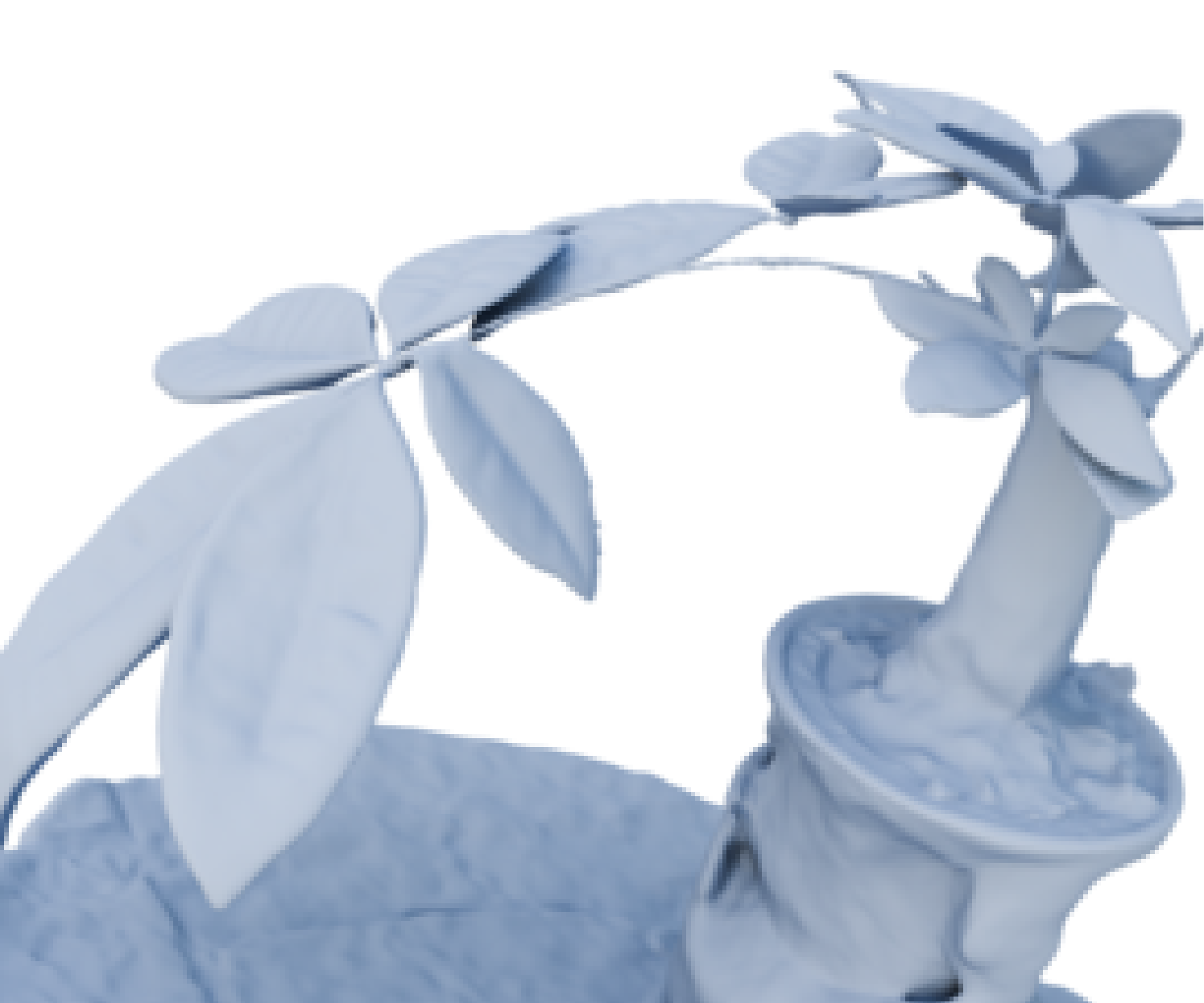} \\
\multicolumn{2}{c}{PGSR~\cite{chen2024pgsr}}
 & \multicolumn{2}{c}{GeoSVR~\cite{li2025geosvr}} &
 \multicolumn{2}{c}{GGGS~\cite{zhang2026geometrygroundedgaussiansplatting}} & Ours\\
    \end{tabular}
    \caption{3DGS-based reconstruction pipelines often rely on a post-processing step to clean the extracted meshes. Due to the lack of strong global geometric constraints, reconstructed components may become noisy or disconnected. Here we show the raw meshes extracted from PGSR, GeoSVR, and GGGS, where the first two contain noticeable noise and floaters, together with the corresponding meshes after the cleaning step. Although post-processing removes these artifacts, it also eliminates thin structures such as the leaves, resulting in missing geometry. In contrast, our method preserves global geometric integrity and does not require filtering or denoising post-processing.}
    \label{fig:tsdf}
\end{figure*}

Furthermore, this pipeline is sensitive to noise in the Gaussian primitives and often produces a large number of floating artifacts, as illustrated in Figure~\ref{fig:tsdf}. To mitigate these artifacts, these methods typically apply a post-processing step based on connected-component analysis to clean the reconstructed mesh. However, for scenes containing thin structures and sign-ambiguous geometry, the raw meshes generated by 3DGS pipelines are often fragmented and exhibit low geometric integrity due to the lack of strong global geometric constraints. Consequently, the cleaning process frequently removes not only noise but also legitimate structures, leading to incomplete reconstructions. In contrast, implicit-field-based approaches, such as NeuS and our method, produce geometries with stronger global consistency and therefore do not require such post-processing steps. Nevertheless, NeuS still struggles to faithfully reconstruct thin structures, whereas our method preserves these structures more accurately.

\section{Conclusion and Discussion}
\label{sec:conclusion}

We presented \method{}, a neural volume rendering framework based on Metric--Phase Fields for stable multi-view reconstruction of sign-ambiguous geometry. By decoupling geometry into an unsigned metric field and a continuous phase field, MPFs maintain the robustness of UDF-like representations for open and thin structures while providing a stable signed signal for occlusion-aware rendering and well-behaved gradients near the zero level set, enabling iso-surface extraction using standard Marching Cubes. Integrating MPF into a volume rendering pipeline requires more than just a direct replacement of SDFs; our framework addresses key challenges in rendering weight design, optimization stability, and supervision ambiguity through residual phase injection and MPF-specific geometric regularization.

Extensive experiments on standard benchmarks and datasets with open surfaces, thin sheets, and non-manifold configurations demonstrate that our method improves reconstruction fidelity and training stability compared to representative SDF, UDF, and 3DGS-based approaches, especially in sign-ambiguous cases where existing methods struggle. These results suggest that MPFs provide a new geometry backbone for extending neural volume rendering beyond the traditional watertight SDF assumption.

\paragraph{Limitations and future directions.}
A current limitation of \method{} is its training efficiency.
In its present form, training a single scene typically takes on the order of 10+ hours, which is comparable to NeuS-style SDF-based volume rendering methods.
This cost mainly stems from dense geometry queries and gradient evaluations along camera rays using an implicit MLP representation.
Improving training efficiency is therefore an important direction for future work.
One promising approach is to incorporate explicit spatial data structures, such as voxel grids or octrees, following frameworks like Voxurf~\cite{wu2022voxurf}, to accelerate geometry queries and reduce unnecessary evaluations in empty space. Because a majority of time is used to calculate the gradients of the implicit field and calculate the computational graph to the gradient to enable optimization with respect to the losses, another promising approach is to adopt numerical gradients instead of analytical gradients as pioneered by Neuralangelo~\cite{Li2023Neuralangelo}. We observe that using numerical gradients can reduce the training time by half without significantly impacting the results.
Another direction is to adopt multi-resolution hash encodings, as in Instant-NGP~\cite{Muller2022InstantNGP}, which have been shown to significantly improve convergence speed and scalability for neural fields.

 As an implicit-field based reconstruction method, our method performs comparable to NeuS but inferior than Gaussian Splatting methods in novel view synthesis. Table~\ref{tab:rendering} reports the quantitative comparison on novel view synthesis using PSNR and SSIM. This result is expected: our method is comparable performance to NeuS because both approaches adopt a similar differentiable volume rendering architecture; while 3DGS methods explicitly optimize radiance fields by fitting dense Gaussian primitives to the observed images, which naturally favors photometric fidelity.  In contrast, our method focuses on learning a well-conditioned geometric field. Although this design may sacrifice a small amount of rendering accuracy, it leads to significantly more reliable surface reconstruction.
Finally, hybrid representations that combine the efficient rendering and optimization properties of 3D Gaussian Splatting~\cite{Kerbl20233D} with the robust geometric modeling capabilities of MPFs represent an exciting avenue for future research, potentially combining the strengths of both paradigms.  

\begin{table}[ht]
\centering
\caption{
Quantitative rendering results on the TAAS-Real (top) and TAAS-Synthetic (bottom) datasets.
Each entry reports PSNR $(\uparrow)$ / SSIM $(\uparrow)$.
Best results are shown in \textbf{bold}, and second-best results are \underline{underlined}.
}
\label{tab:rendering}

\setlength{\tabcolsep}{3pt}
\renewcommand{\arraystretch}{1.15}

\resizebox{\linewidth}{!}{\begin{tabular}{lcccc}
\toprule
Method & Basket & Umbrella & Cup & Flower \\
\midrule
GaussianUDF & \textbf{28.247} / \textbf{0.909} & 32.023 / \underline{0.916} & 26.244 / \underline{0.918} & \textbf{27.308} / \underline{0.862} \\
GeoSVR & 26.492 / \underline{0.883} & 28.429 / \textbf{0.930} & 23.440 / 0.890 & \underline{26.529} / \textbf{0.869} \\
NeuS & 27.907 / 0.857 & \textbf{32.137} / 0.857 & \textbf{29.145} / \textbf{0.919} & 25.425 / 0.728 \\
PGSR & 26.050 / 0.880 & 29.053 / 0.900 & 23.993 / 0.887 & 25.659 / 0.843 \\
Ours & \underline{28.033} / 0.861 & \underline{32.128} / 0.855 & \underline{28.991} / 0.917 & 25.734 / 0.730 \\
\midrule
Method & High-Heeled Shoe & Tank & LEGO & Figurine \\
\midrule
GaussianUDF & \underline{24.847} / \underline{0.921} & 31.791 / \underline{0.933} & \textbf{25.413} / \textbf{0.931} & 27.818 / \underline{0.874} \\
GeoSVR & 21.002 / 0.875 & 28.796 / 0.913 & 23.546 / 0.899 & 27.239 / 0.870 \\
NeuS & 23.103 / 0.851 & \textbf{33.288} / 0.931 & 24.854 / 0.908 & 28.764 / 0.858 \\
PGSR & \textbf{25.763} / \textbf{0.923} & \underline{32.952} / \textbf{0.945} & \underline{25.065} / \underline{0.925} & \textbf{29.924} / \textbf{0.898} \\
Ours & 23.465 / 0.853 & 32.646 / 0.929 & 24.966 / 0.914 & \underline{29.222} / 0.862 \\
\bottomrule
\end{tabular}}

\end{table}

\bibliographystyle{IEEEtran}
\bibliography{main}

\end{document}